\documentclass[letterpaper, 10 pt, conference]{ieeeconf}  % Comment this line out if you need a4paper
\IEEEoverridecommandlockouts                              % This command is only needed if 
\usepackage{amsmath,amsfonts,bm}

\def\eqref#1{equation~\ref{#1}}
\def\1{\bm{1}}

\DeclareMathAlphabet{\mathsfit}{\encodingdefault}{\sfdefault}{m}{sl}
\SetMathAlphabet{\mathsfit}{bold}{\encodingdefault}{\sfdefault}{bx}{n}

\DeclareMathOperator*{\argmax}{arg\,max}

\usepackage{hyperref}
\usepackage{url}
\usepackage{amsfonts} 
\usepackage{graphics} % for pdf, bitmapped graphics files
\usepackage{epstopdf} % for postscript graphics files
\usepackage{times} % assumes a new font selection scheme installed
\usepackage{amsmath} % assumes amsmath package installed
\usepackage{amssymb}  % assumes amsmath package installed
\usepackage[ruled,vlined]{algorithm2e}
\usepackage{subfig}
\usepackage{hyperref}
\usepackage{xcolor}

\usepackage{graphicx} % omit the 'demo' option in a real document
\usepackage{float}
\usepackage{cite}
\usepackage{soul}
\usepackage{breqn}

\title{\bf RE-MOVE: An Adaptive Policy Design for Robotic Navigation Tasks \\ in Dynamic Environments via Language-Based Feedback
}

\author{Souradip Chakraborty$^{1,*}$, Kasun Weerakoon$^{1,*}$, Prithvi Poddar$^{2,*}$, Mohamed Elnoor$^{1}$, Priya Narayanan$^{3}$, \\  Carl Busart$^{3}$, Pratap Tokekar$^{1}$, Amrit Singh Bedi$^{1}$, and Dinesh Manocha$^{1}$
\\
\url{http://gamma.umd.edu/remove/}
\thanks{*Denotes equal contribution.}% <-this % stops a space
\thanks{$^{1}$Department of Computer Science, University of Maryland, College Park, USA. 
        {Email:\tt schakra3@umd.edu}}%
\thanks{$^{2}$Department of Mechanical and Aerospace Engineering, University at Buffalo, Buffalo, NY, USA.}
        \thanks{$^{3}$DEVCOM US Army Research Laboratory, Adelphi, MD, USA.}
        \thanks{This work was supported by Army
Cooperative Agreement W911NF2120076. We acknowledge the support of the Maryland Robotics Center.}% \
}

\begin{document}

\maketitle

\begin{abstract}
Reinforcement learning-based policies for continuous control robotic navigation tasks often fail to adapt to changes in the environment during real-time deployment, which may result in catastrophic failures.
To address this limitation, we propose a novel approach called RE-MOVE (\textbf{RE}quest help and \textbf{MOVE} on) to {adapt already trained policy to real-time changes in the environment without re-training} via utilizing a language-based feedback. The proposed approach essentially boils down to addressing two main challenges of \emph{(1) when to ask for feedback} and, if received, \emph{(2) how to incorporate feedback into trained policies}. RE-MOVE incorporates an {epistemic uncertainty-based framework} to determine the optimal time to request instructions-based feedback. For the second challenge, we employ a zero-shot learning natural language processing (NLP) paradigm with efficient, prompt design and leverage state-of-the-art {GPT-3.5, Llama-2} language models. To show the efficacy of the proposed approach, we performed extensive synthetic and real-world evaluations in several test-time dynamic navigation scenarios. Utilizing RE-MOVE result in up to 80\% enhancement in the attainment of successful goals, coupled with a reduction of 13.50\% in the normalized trajectory length, as compared to alternative approaches, particularly in demanding real-world environments with perceptual challenges.
\end{abstract}

\section{Introduction}
Reinforcement learning (RL) has gained popularity for navigating complex, dynamic environments \cite{weerakoon2022terp}.
Real-world scenarios often involve unpredictable obstacles or perceptually deceptive objects absent during training, limiting the generalizability of learned policies \cite{perception_challenges}. To address these challenges, there is a need for novel solutions to enhance the adaptability of RL policies for robot navigation in complex and dynamic environments. 
\begin{figure}
    \centering
\includegraphics[width=0.7\columnwidth]{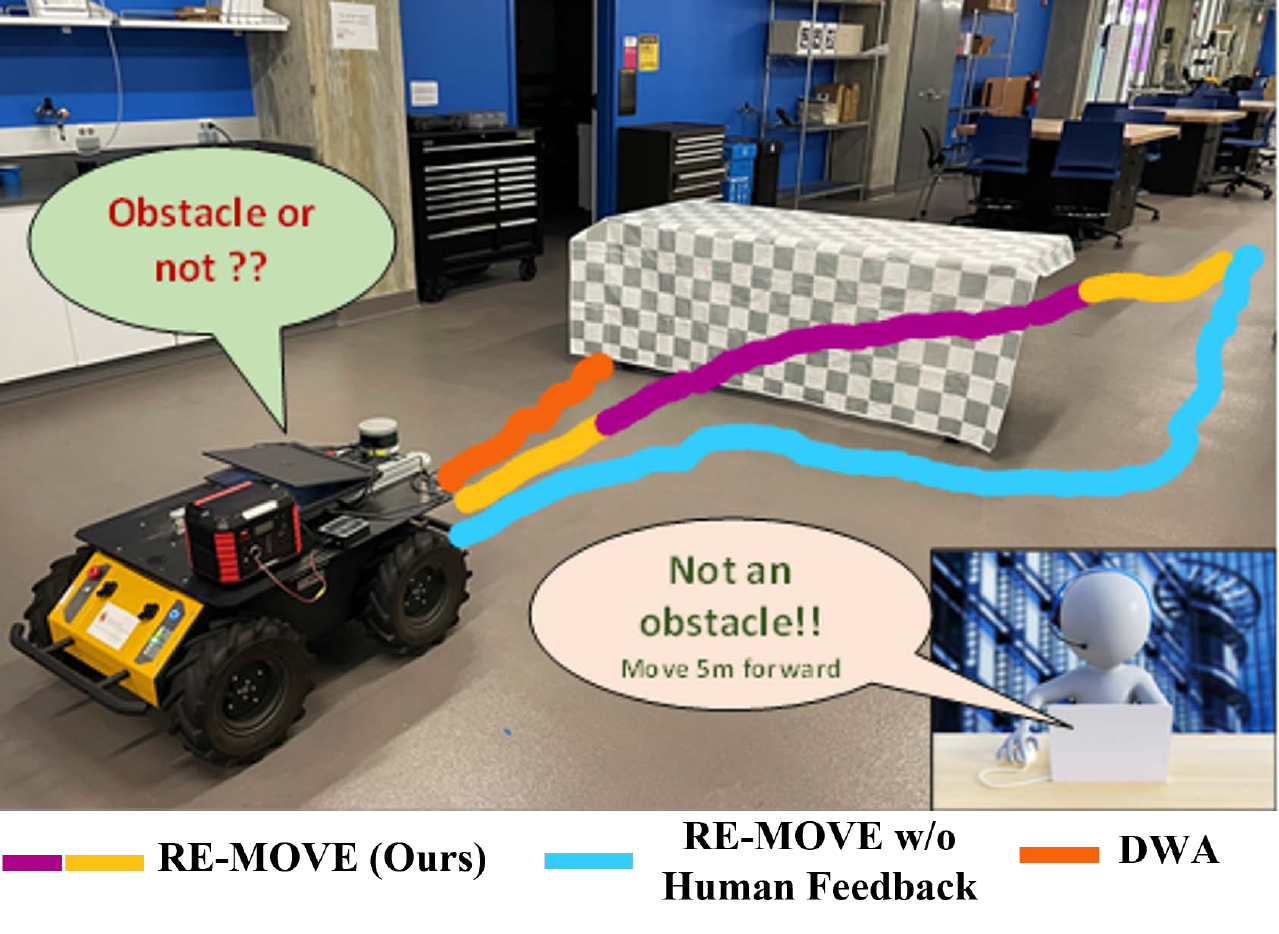}
    \caption{\small {This figure shows robot navigation using our RE-MOVE approach with a language-based feedback scenario.
    During deployment in dynamic scenes, RE-MOVE identifies the uncertainties that appear in the observation space (i.e., a LiDAR laser scan-based 2D cost map in our context) and requests assistance from a human. Such assistance is essential in scenarios where the laser scan misleadingly detects pliable regions (i.e., \textit{\textbf{perceptually deceptive yet navigable objects}} such as hanging clothes, curtains, thin tall grass, etc.) as solid obstacles due to the sensing limitations of the LiDAR. %
    }}
    \label{fig:cover}
  %  \vspace{-20pt}
 \end{figure}
In this work, we propose utilizing language-based human feedback to handle unexpected environmental changes and objects during test time. To achieve this, the first challenge to address is determining the optimal timing for a robot to request assistance in the form of feedback. The subsequent challenge involves identifying the most effective form for the robot to request feedback and implementing the necessary adaptations to the policy based on the received feedback. The first challenge underscores the importance of managing sample complexity in the strategy, as we cannot continuously request feedback during deployment. To tackle this, we quantify epistemic uncertainty precisely, considering specific design considerations within the robot's input space. The prior research on uncertainty-driven learning approaches for navigation is mainly limited to training time \cite{du2022bayesian, bayesian2}. Further, there are no detailed studies on the impact of the observation or input-space design and its entanglement with downstream uncertainty quantification during test time, which becomes a critical task for the navigation setting.

A key novelty of our approach in robotic navigation is the efficient quantification of epistemic uncertainty in test-time scenarios. This provides a measure of the robot's confidence in its action predictions given the observation and is crucial in real-world navigation settings where learning from mistakes is not an option \cite{dulac2019challenges}. Similarly, providing a large diverse batch of simulations for training that can cover all possibilities of the real world is also infeasible. Hence, the idea is to monitor the epistemic uncertainty over the episode during the test time. Then, based on a predefined robust change point detection algorithm, the robot can determine when it is appropriate to ask for help, thus improving upon its test-time adaptability to unseen scenarios.  
The second question is regarding the design of an efficient \textit{interpretation and adaptation} module for language feedback. To tackle this challenge, we employ a zero-shot learning Natural Language Processing (NLP) paradigm.

\noindent \textbf{Main contributions:} Our \textit{contributions} are as follows. 
\begin{itemize}
\item In this work, we develop a novel approach called RE-MOVE (\textbf{RE}quest and \textbf{MOVE} on) for robot navigation tasks in order to improve the test time adaptability of trained RL policies {without re-training}.  

\item To decide when to ask for feedback, we develop a novel characterization of epistemic uncertainty with ensemble-based approximate Bayesian inference for navigation scenarios and are one of the first to conduct a design study aimed at understanding the relationship between the agent's input space and downstream uncertainty estimation \& calibration.

\item As another key contribution, we focus on developing an efficient \textit{interpretation and adaptation} module for handling language feedback. We adopt a zero-shot learning approach in Natural Language Processing (NLP) to address this challenge with a streamlined, prompt-driven design. We harness the capabilities of state-of-the-art language models such as {GPT-3.5, Llama-2} for our interpretation module, as detailed in \cite{GPT3}.

\item We demonstrate the effectiveness of our proposed framework RE-MOVE via various synthetic (tabular, continuous, and visual) and real-world experiments. We show that RE-MOVE can interpret grounded language and adapt to it efficiently with minimal supervision and query efficiency. Hence, RE-MOVE results in an increase upto 80\% in successful goal-reaching and a decrease of 13.50\% in the normalized trajectory length compared to other methods in perceptually challenging real-world environments. 
    % \textcolor{red}{may be some numbers}
\end{itemize}

\section{Related Works}

\noindent \textbf{Robot Navigation with Feedback:} Development of fully autonomous navigation algorithms still remains a challenging task, especially as consumer products that can ensure the physical and mental safety of humans in the operating environment \cite{chen2021HCIeffects,adaptive_planner_w_feedback}. On the other hand, humans are capable of making accurate perceptions and planning decisions under such extremely challenging social and environmental settings for robots. Hence, recent reinforcement learning (RL) studies highlight the advantage of learning from human feedback for RL agents to enhance real-world deployments \cite{hcrl,interactive_rl}. 
Such feedback is significantly useful for the robots to learn features that are difficult to learn from raw sensory observations \cite{trust_driven_navigation}. For instance, human feedback on socially compliant actions \cite{socially_aware_navigation,ritschel2019health_companion} or perceptually challenging environments \cite{agnisarman2019visual_inspection} improves the robot's navigation performance. In \cite{A-SLAM}, an augmented SLAM algorithm is proposed that allows users to interact with the robot and correct pose and map errors. A shared control and active perception framework is proposed in \cite{perception_aware_human_assisted_nav} to combine the skills of human operators with the capabilities of a mobile robot. 

\noindent \textbf{Natural Language for Robot Navigation:} Recent studies demonstrate that leveraging human knowledge of the environment can significantly enhance a robot's navigation performance \cite{nguyen2022framework,chen2011learning}. However, communication skills to understand human instructions are essential to bridge the gap between the natural language and the robot's actions\cite{peng2016need}. In \cite{nguyen2019vision}, imitation learning with an indirect intervention approach is proposed for vision-based navigation with language-based assistance. Similarly, several grounded parsing models that interact with robots through natural language queries against a knowledge base are presented in recent studies\cite{matuszek2012joint,tellex2020robots,zang2018translating}.

\noindent \textbf{Navigating in Noisy Sensor Data Environments:} Accurate robot perception plays a significant role in efficient autonomous robot navigation. The majority of robot navigation failures occur due to erroneous state estimations provided by inaccurate perception models. Such errors often arise in challenging environmental settings that can deceive the perception sensors. For instance, lighting changes \cite{wang2022applications_vision}, motion blur \cite{visual_navigation}, and occlusions significantly reduce the perception capabilities of vision-based models \cite{stein2021partially_reveled_env}. Similarly, cluttered environments can provide misleading object detection(i.e., pliable objects such as curtains, hanging clothes, etc as obstacles) from LiDAR sensors\cite{2022arXivgraspe}. To this end, multi-sensor perception methods are incorporated in the literature \cite{hu2020_multi_sensor_survey,wang2019multi_sensor}. However, oftentimes such methods are computationally expensive and require multiple expensive sensors for perception. %Hence, in this work, we incorporate only a 2D liDAR scan with minimum human feedback to perform efficient navigation under perceptually deceiving settings.

\section{Problem Formulation and Notations}
Mathematically, we formulate the robotic navigation problem as a Markov Decision Process (MDP) defined as $\mathcal{M}:=\{\mathcal{S}, \, \mathcal{A}, \, {P},\,r,\, \gamma \}$, where $\mathcal{S}$ denotes the state space,  $\mathcal{A}$ is the actions space, ${P}(s'|s,a)$ is the transition kernel, $r(s,a)$ is the reward, and $\gamma\in(0,1)$ denotes the discount factor. The objective of the robot is to learn a navigation policy $\pi_{\theta}(a |s)$ (probability of taking action $a$ in state $s$) parameterized by $\theta$ to perform efficient navigation. Since we are interested in solving the problem of test time adoption of a trained policy for a robotic navigation task, we first discuss the method we consider to obtain a trained policy. 
\begin{figure}
    \centering
\includegraphics[scale=0.5]{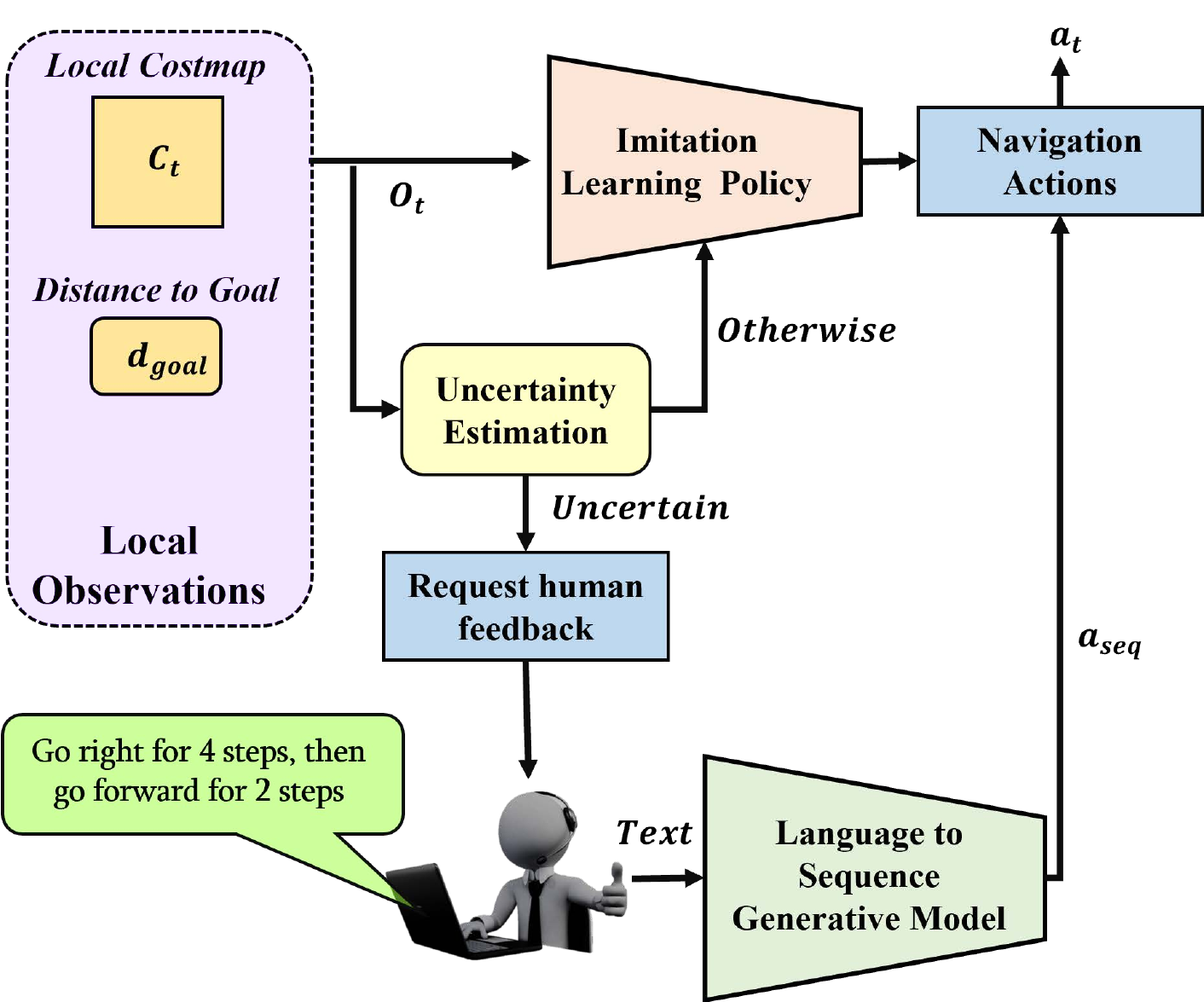}
    \caption{\small{Algorithmic Flow Diagram of RE-MOVE.}}
    \label{fig:sys-arch}
    \vspace{-15pt}
 \end{figure}
We consider an imitation learning framework to leverage expert demonstrations to learn the behavioral policy. For simplicity, we illustrate our framework using imitation data for the trained policy, but it can be seamlessly applied for any RL algorithm without loss of generality.
Let us denote expert demonstrations by $\mathcal{D}:=\{\tau_i\}_{i=1}^N$, a collection of $N$ trajectories collected by an expert to solve the underlying navigation task. Here, $\tau_i$ represents the $i^{th}$ trajectory given by $\tau_i=\{s_0^i,a_0^i,s_1^i,a_1^i,\cdots,s_T^i,a_T^i\}$ where $(s_j^i, a_j^i)$ denotes the state-action pair. 

The objective of learning behavioral policy can be re-formulated as maximizing the likelihood of the trajectories under the demonstration as 
where $P_{\theta}(\tau_i)$ denotes  the probability of the $i^{th}$ trajectory of the expert. We note that $P_{\theta}(\tau_i) = P(s_0^i) \prod_{t=1}^{T} \pi_{\theta}(a_t^i|s_t^i)P(s_{t+1}^i|s_t^i,a_t^i)$. 
    Using $P_{\theta}(\tau_i)$ , we can obtain the gradient of 
the objective $\log L (\theta)$ with respect to $\theta$ as
%
    %
   % \begin{align}\label{gradient}
  $\nabla_{\theta}  \log L (\theta) = \sum_{i=1}^{n} \sum_{t=1}^{T} \nabla_{\theta}  \log \pi_{\theta}(a_t^i|s_t^i)$.
%\end{align}
%
Thus, with the imitation learning-based formulation, we can compute the gradient w.r.t. to the loss function $ \nabla_{\theta} \log L (\theta)$. However, the trained policy $\pi_{\theta}$ works well in training scenarios but fails to adapt to unseen test scenarios \cite{8794134}. {A significant factor contributing to failure is the inability of the robot to stop at the right time while confronting dynamic barriers in the test time, leading to catastrophic failures. An effective remedy involves enabling the agent to stop immediately upon encountering new, unexpected deviations from its training environment, which requires well-calibrated and uncertainty-aware frameworks but is challenging to design.} Hence, an efficient characterization of uncertainty is critical such that once it reaches a certain threshold, the agent can stop and ask for feedback. However, the estimation of uncertainty is entangled with the design of the input space and a generic input space might fail to provide an evident well-calibrated uncertainty, as we demonstrate in Sec. \ref{sec:experiments}.
\section{Proposed Approach}

Before delving into the details of characterizing the uncertainty, we first discuss the design of observation space for the RL agent or robot. We note that developing a well-calibrated policy for an agent requires a careful design of the observation space of the agent since the epistemic uncertainty of the policy is directly related to the feature/state-space of the agent. 
\subsection{Designing the observation space} \label{sec:observation_space}
In this work, we analyzed three unique configurations of the observation space. Further details are as follows:

\begin{itemize}
    \item \textit{Global visibility observation:} This combines global and local observations of the agent. The global observation provides a representation of the entire environment, including the location of the agent and obstacles. This observation can be thought of as a complete map of the environment in which the agent is navigating. 
    
    \item \textit{Goal-conditioned observation:} This utilizes only the local observation of the agent and includes a measure of the distance between the agent and the goal.

    \item \textit{Partial global visibility observation:} %
    This combines global and local observations similar to the global visibility observation. However, unlike the latter, global observation does not contain any information about unseen environmental obstacles. As a result, the agent can only detect obstacles when they are in close proximity.
\end{itemize}

\subsection{Characterization and Representation of the Uncertainty}

The total uncertainty in the prediction can be decomposed into aleatoric and epistemic uncertainty \cite{uncertainty2, uncertainty3, uncertainty4, verma2021uncertaintyaware}. Epistemic uncertainty refers to the uncertainty in a model parameters, which can be estimated by measuring the variance in the parameter distribution space. 
Next, we try to emphasize the significance of monitoring the epistemic uncertainty specific to our dynamic navigation scenario as described below.

Consider our navigation scenario, and let us say instead of modeling the epistemic uncertainty, i.e., uncertainty w.r.t. the distribution over the parameters, we monitor the uncertainty over the action probability by the policy $\pi(a|s)?$. However, this approach fails to produce an efficient representation of uncertainty. This is because multiple common points in the trajectory lead to different actions, all of which are optimal considering their respective trajectories. Therefore, if we consider entropy in the probability space by monitoring $\pi(a|s)$, it will be high in all those regions, indicating multiple routes to travel from that path and not because these are unseen observations.
Consequently, it becomes difficult to segregate such uncertainty when an appropriate division between the two is not considered, leading to misleading and erroneous inferences. 

\textbf{Decomposition of Predictive Uncertainty.} 
Here we emphasize the decomposition of the uncertainties for our parameterized policy $\pi_{\theta}$ with demonstration data. In our setting, the total uncertainty can be represented as the entropy of the softmax probability in the action space as $H(\hat{a}|s, \mathcal{D})$, which can be decomposed as \cite{uncertainty3}
\begin{align}
    H(\hat{a}|s, \mathcal{D}) = I(\hat{a}, \theta|s, \mathcal{D}) + \mathbb{E}_{\theta \sim P(\theta|\mathcal{D})} [H(\hat{a}|s, w)],
\end{align}
where $I(\hat{a}, \theta|s, \mathcal{D})$ represents the mutual information, $\hat{a}$ represents the output predicted probability from the model after the softmax layer. This represents the decomposition into aleatoric and epistemic uncertainty as \cite{uncertainty1, uncertainty3}, where $H(\hat{a}|s, \mathcal{D})$ represents the predictive uncertainty, $\mathbb{E}_{\theta \sim P(\theta|\mathcal{D})} [H(\hat{a}|s, w)]$ represents the average entropy where the effect of the parameters are marginalized and hence depends primarily on the input state $s$ and hence can be thought of as the aleatoric uncertainty. $I(\hat{a}, \theta|s, \mathcal{D})$ represents a notion of the epistemic uncertainty, which is most important for our scenario. Next, we emphasize on estimating the epistemic uncertainty with an ensemble-based approximate Bayesian method as
\begin{align}
    \mathbb{E}_{\theta \sim P(\theta|\mathcal{D})} & [H(\hat{a}|s, w)] \\ \nonumber
    &= - \int_{\theta} P(\theta|\mathcal{D}) \sum_{a \in \mathcal{A}} P(\hat{a}|s, \theta) \log P(\hat{a}|s, \theta) d\theta \\ \nonumber
    & \approx \int_{\theta} \hat{P}(\theta) \sum_{a \in \mathcal{A}} P(\hat{a}|s, \theta) \log P(\hat{a}|s, \theta) d\theta \\ \nonumber
    & = \frac{1}{k} \sum_{k} \sum_{a} P(\hat{a}|s, \theta^k) \log P(\hat{a}|s, \theta^k).
\end{align}
where, $P(\hat{a}|s, \theta)$ represents the predicted softmax output probability for action $a$ using the $k^{th}$ model in the ensemble. $\hat{P}(\theta)$ represents the approximate posterior. 
Now, the epistemic uncertainty thus can be computed as the difference between the two estimates as 
\begin{align}
     I(\hat{a}, \theta|s, \mathcal{D}) =  \mathbb{E}_{\theta \sim P(\theta|\mathcal{D})} [H(\hat{a}|s, w)] - H(\hat{a}|s, \mathcal{D})
\end{align}
In our experiments, we compute the epistemic uncertainty at every time step and use a changepoint detection algorithm to identify if there is a significant increase in the uncertainty to prompt the human expert for feedback. We use the Bayesian changepoint detection algorithm described in \cite{Fearnhead2006} and thus improve the query efficiency of our framework. 
\begin{figure}[t]
    \centering
\includegraphics[width=\columnwidth]{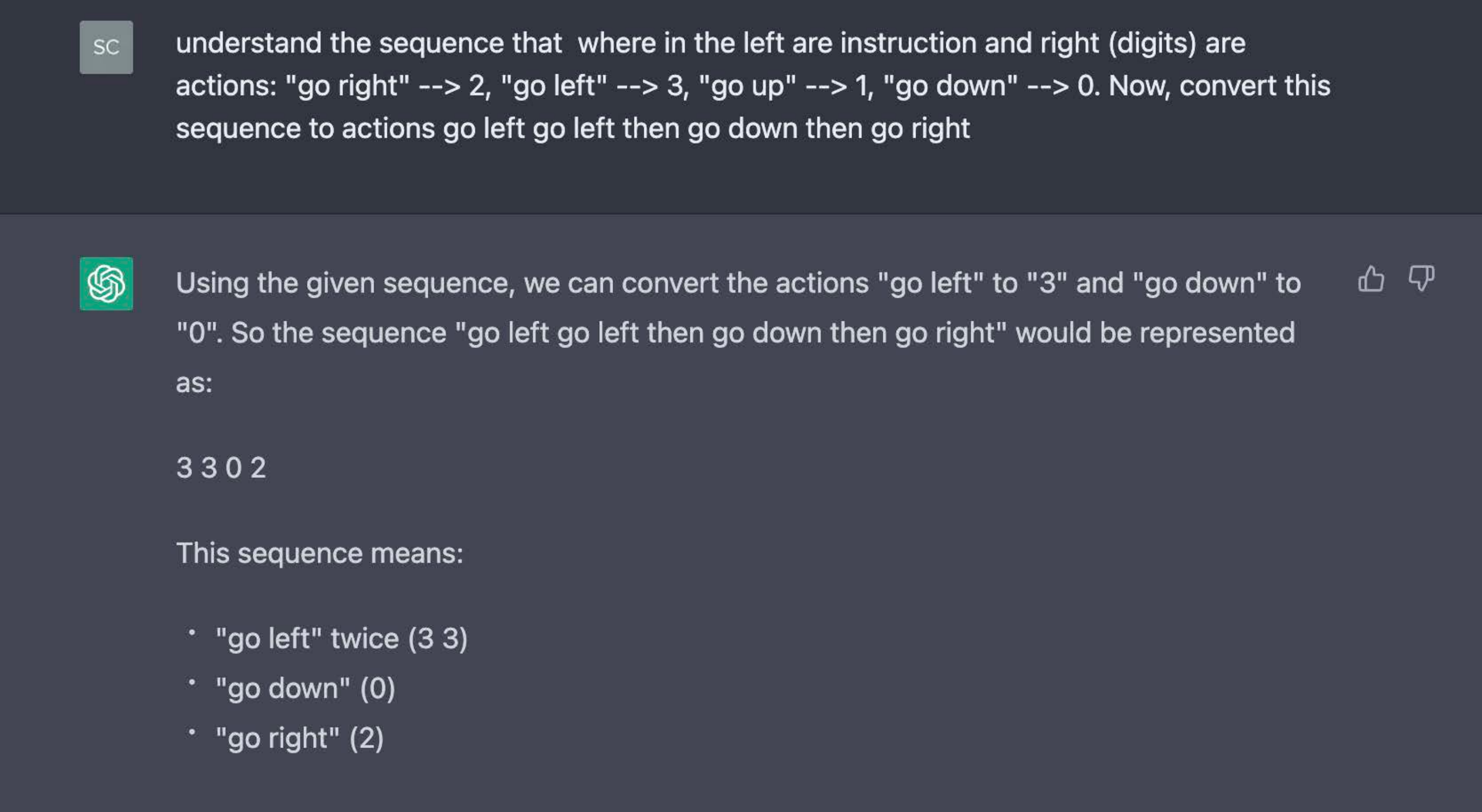}
    \caption{ \small {Utilizing the zero-shot generalization capabilities of ChatGPT to convert textual instructions from the human expert, into a sequence of actions. The model is first familiarized with the action-to-integer mappings and then tasked to convert a textual prompt into a sequence of integers that correspond to the actions specified in the prompt.}}
\label{chatGPT_figure}
 \end{figure}

\subsection{Incorporating Human Language Feedback}
%\subsubsection{Sequence to Sequence Generative Model for Grounded Language to Sequence}
One major challenge in the framework lies in interpreting the language instruction provided by humans and converting it to a sequence of activities that can be incorporated into the framework of the agent and that too with limited supervision. Hence, the broader goal is to model the condition generative process $P(a_1, a_2 \cdots a_n| h_1, h_2, \cdots)$ where $a_i, h_i$ represents the $i^{th}$ action sequence and the human grounded language instruction respectively, $T, S$ represent the output action and input grounded instruction sequences respectively. Then the training objective of maximizing the log-likelihood can be modeled in the sequence-sequence framework as $\max {\frac{1}{|H|} \sum_{(T, H) \in \mathcal{H}} \log P(T|H)}$. Once the training is done, the most likely output sequence of actions given the input instruction sequence is modeled as $\hat{T} = \argmax_{T} P(T|H)$. While various approaches exist for modeling sequence-sequence frameworks like BART \cite{bart}, and T5 \cite{t5} these methods typically require large amounts of supervised data, which may not always be always a reasonable assumption. To address this issue, we propose a near zero-shot generalization framework that leverages the discretization of the action space in our navigation scenarios. By exploiting this property, we are able to design efficient prompts for language models, providing a more feasible and effective solution. In this work, 
we utilize the zero-shot generalization capability of ChatGPT \cite{chatgpt, chatgpt1, chatgpt2} by a suitable prompt design and achieve improved zero-shot performance (cf. Figure \ref{chatGPT_figure}). We are one of the first to utilize the efficacy and generalization capability of ChatGPT in uncertainty-driven navigation tasks. 
\begin{figure}
    \centering
 \includegraphics[width=0.7\columnwidth]{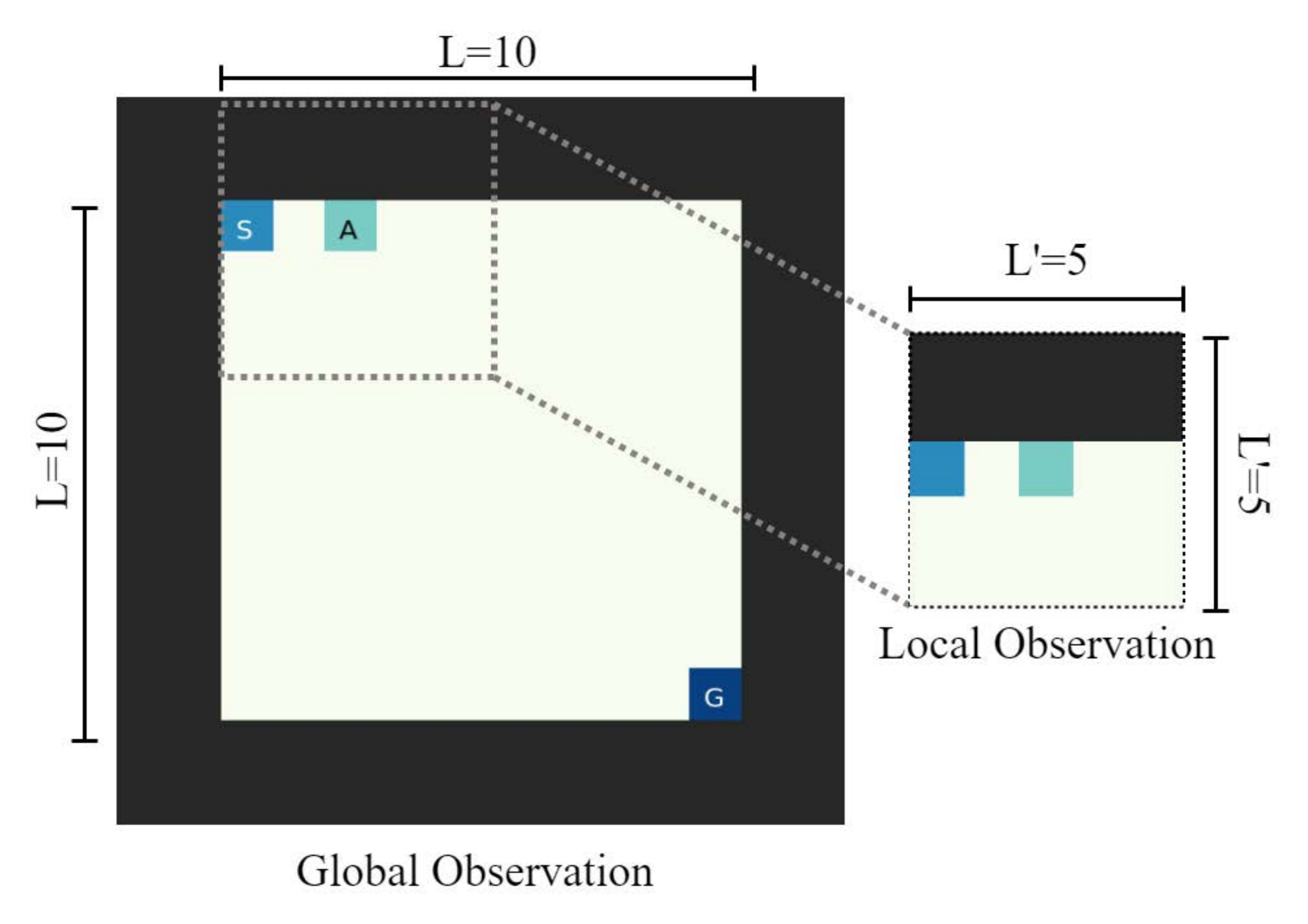}
    \caption{{Pictorial representation of the discrete grid world.
    }}
    \label{fig:env_sample}
    \vspace{-18pt}
 \end{figure}

  \subsection{Overall Algorithmic Description} As in Figure \ref{fig:sys-arch}, the algorithm collects a limited set of expert demonstration trajectories and designs a localized goal-conditioned input space for the agent. It then trains a behavioral policy using imitation learning by maximizing the log-likelihood of the probability of occurrence of the trajectories. The policy is deployed to test environments, and the algorithm monitors epistemic uncertainty wrt model parameters using a Bayesian change point detection algorithm. If the agent is uncertain, it asks the human for help and converts language into actions. 

\section{Experiments} \label{sec:experiments}

% \subsection{Synthetic Experiments}
We conduct experiments in both synthetic and real-world environments. {In particular, we consider (1) Discrete grid world, (2) Visual grid world, (3) Simulated outdoor environment, and (4) Real-world indoor navigation for training and testing.}
\subsection{{Discrete \& Visual grid world}}
\begin{figure*}
\centering
  \subfloat[]{\label{fig:s1_no_obs} \includegraphics[width=8cm]{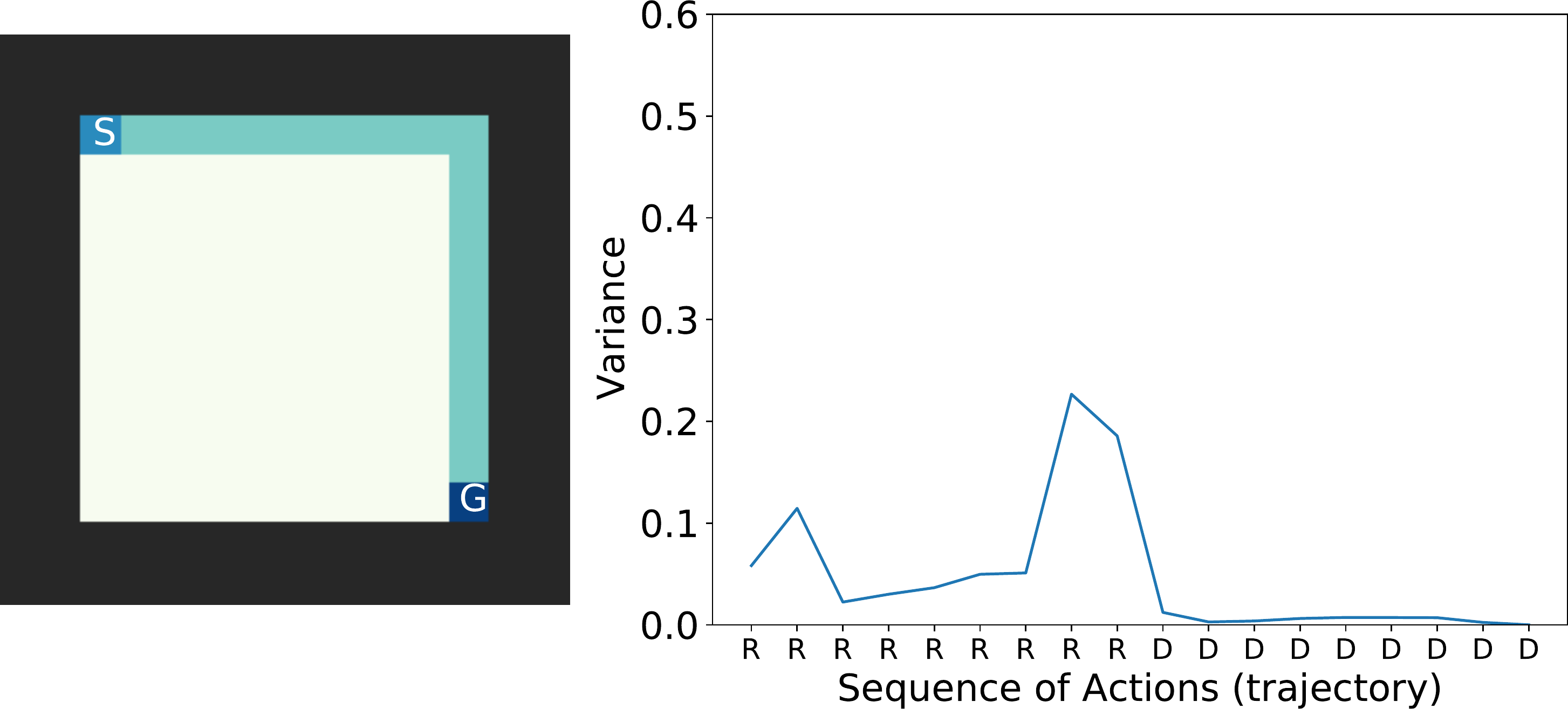}}
  \hspace{0.2cm}
  \subfloat[]{\label{fig:s1_obs}
  \includegraphics[width=8cm]{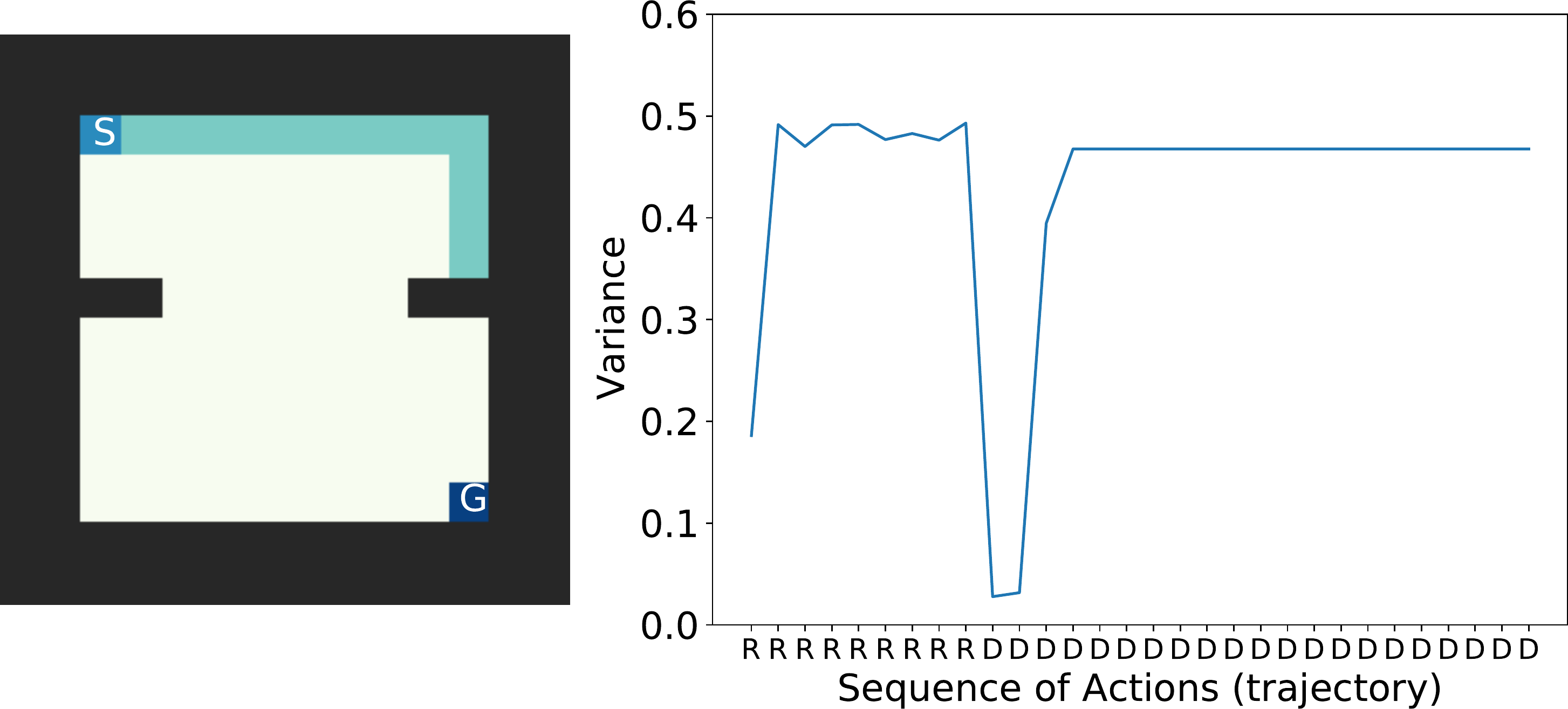}}
  \newline
    % \centering
  \subfloat[]{\label{fig:s2_obs_fail_2} \includegraphics[width=8cm]{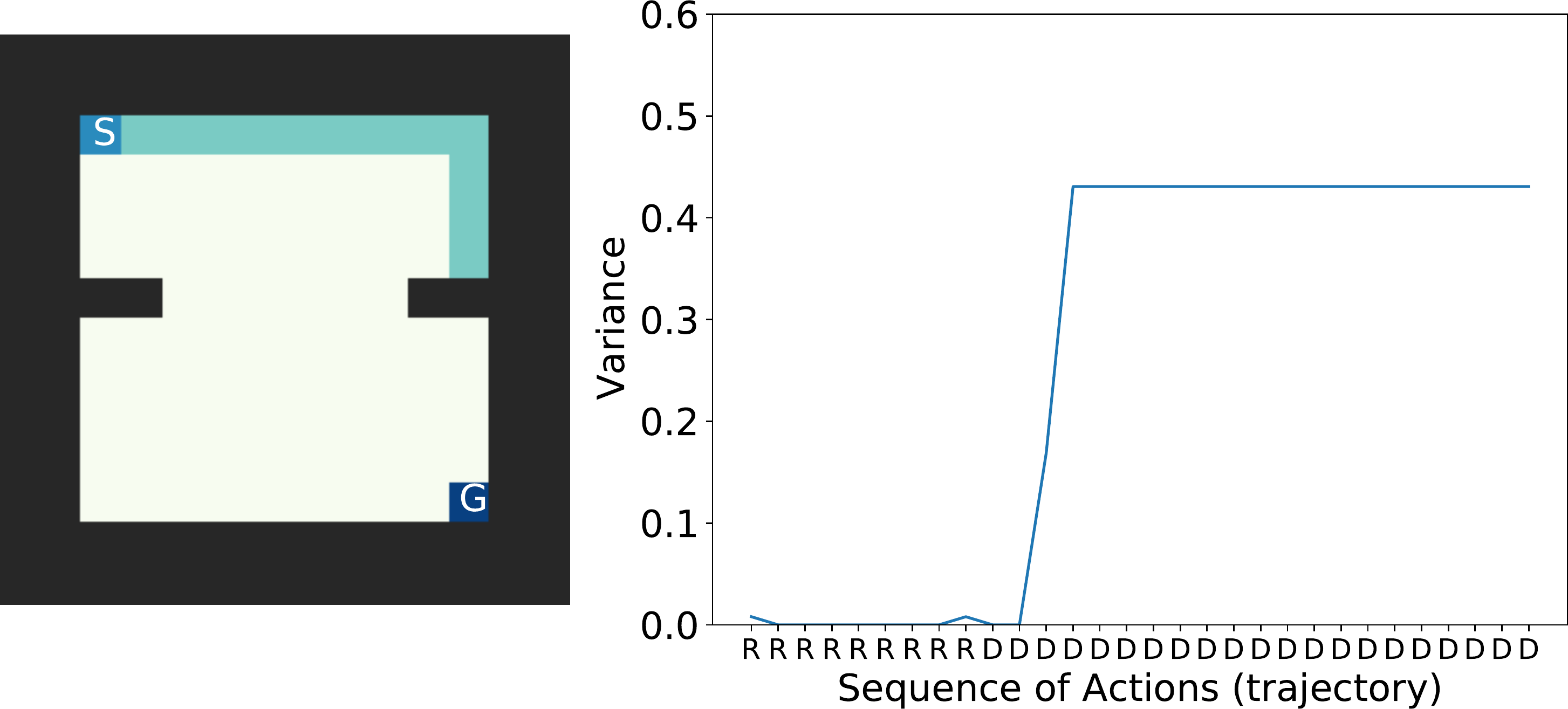}}
  \hspace{0.5cm}
  \subfloat[]{\label{fig:s2_obs_jit}
  \includegraphics[width=8cm]{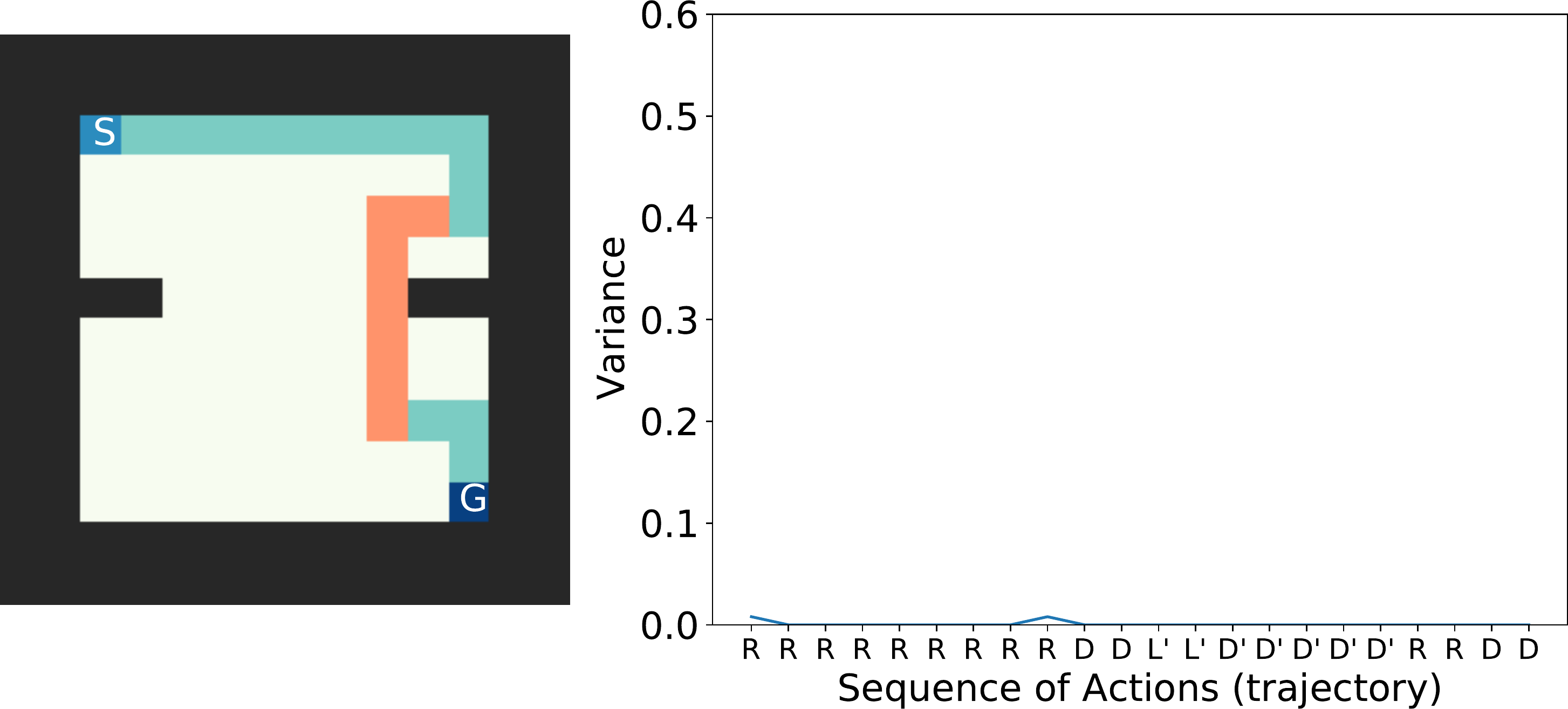}}
% \\
    \caption{ \small{(a) Shows the trained policy with no obstacles, using the global visibility observation space. (b) This figure shows the same policy but with unseen obstacles during the testing. The variance in the action predictions made by the ensembles represents the uncertainty in the model's prediction (higher variance correlates to higher uncertainty). The results obtained in (b) reveal that the policy demonstrates high uncertainty predictions during the initial stages of the trajectory when the agent is far from any obstacle. These characteristics render the configuration unsuitable for the RE-MOVE algorithm, as the high uncertainties are only desired when the agent is in close proximity to the obstacle. (b) and (c) represent the trajectories followed by the RE-MOVE agent (in an environment with previously unseen obstacles) with and without human feedback. The \textit{blue} trajectory is followed by the trained policy, while the \textit{orange} trajectory is generated from human feedback. (c) Shows that without human feedback, the agent gets stuck behind the obstacle that lies on the optimal path. (d) Shows the trajectory followed by the agent with feedback from the human eventually avoiding the obstacle.}}\label{fig:goal-cond-new-conf}
   % \vspace{-15pt}
\end{figure*}

{The grid world environments in our experiments ((1)-(2)) include a 2D maze-like environment where the agent can take discrete steps and needs to navigate from the start to the goal position. In the discrete grid-world,} we test our framework by considering two variations of the input space: the global visibility and the goal-conditioned observation spaces as shown in Figure \ref{fig:env_sample} (refer to \cite[Appendix \ref{discrete_grid}]{chakraborty2023re} for details). The global visibility observation space considers the state of the entire grid as the input, while the goal-conditioned observation space considers only the local observation of the agent. The evaluation results for both cases are shown in Fig.\ref{fig:goal-cond-new-conf}. Fig.\ref{fig:s1_no_obs} and Fig.\ref{fig:s1_obs} show how the global visibility observation space is not suitable for RE-MOVE while Fig.\ref{fig:s2_obs_fail_2} and Fig.\ref{fig:s2_obs_jit} show the performance of RE-MOVE, with and without human feedback, using the goal-conditioned observation space. In the \textbf{visual grid world}, we replace tabular states with image-based inputs, adding complexity and making the setup more reflective of real-world conditions. This environment is particularly useful for evaluating image-based navigation policies. For the discrete grid, we use boosted ensembles and vanilla Neural networks as the agent policy networks and boosted uncertainty to estimate the epistemic uncertainty. For the visual inputs on the other hand, we use CNN policies with MC-dropout for uncertainty estimation.

\textbf{The simulated outdoor environment} (3) includes different types of trees, buildings, etc., and a Clearpath husky robot with a LiDAR. The outdoor high-fidelity simulated environment closely replicates the real-world navigation, refer to \cite[Appendix \ref{simulation_environment}]{chakraborty2023re} for further details.

\begin{figure*}[t] 
    \centering
  \subfloat[]{
  \includegraphics[width=0.075\textwidth]{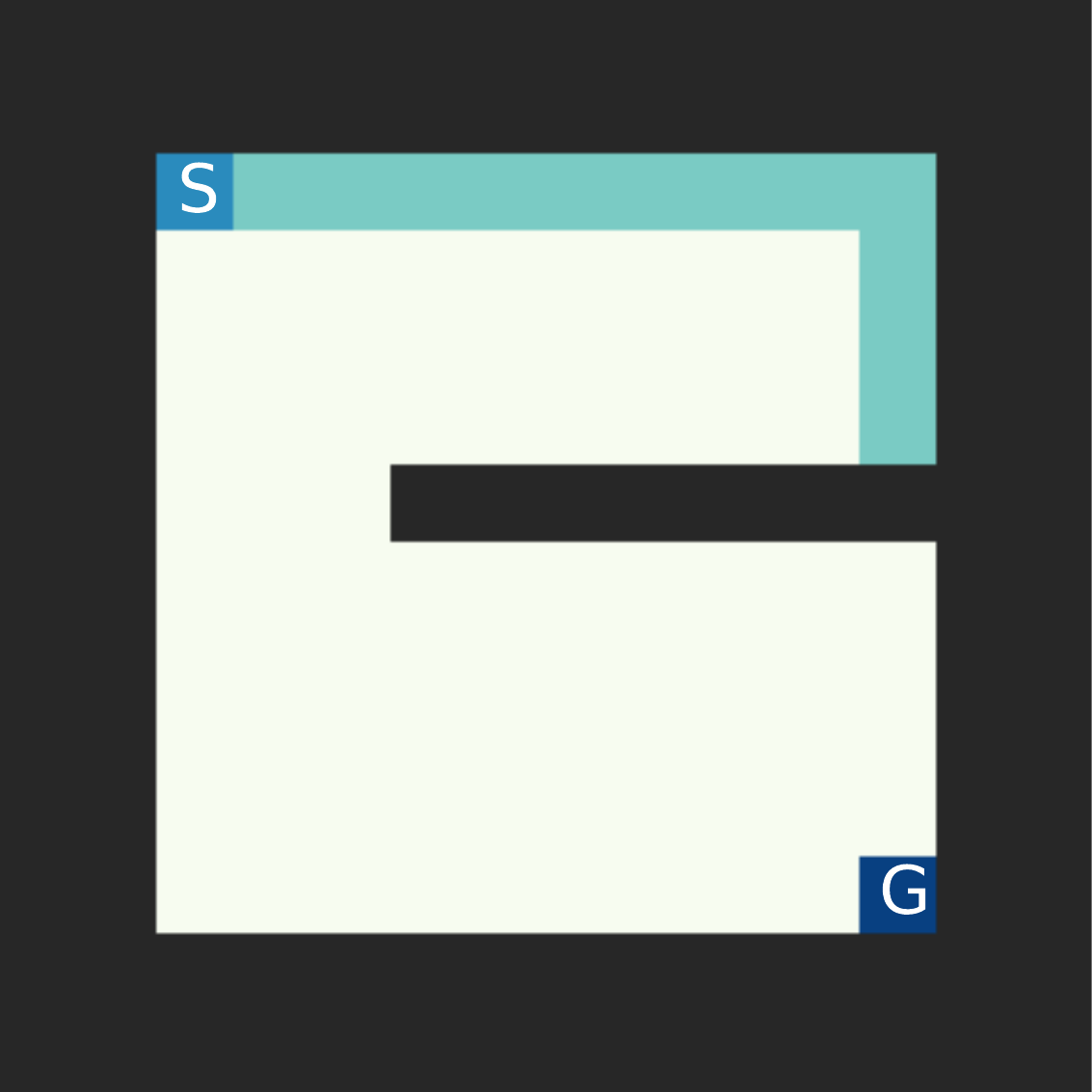}
  %\hspace{0.1cm}
  \includegraphics[width=0.075\textwidth]{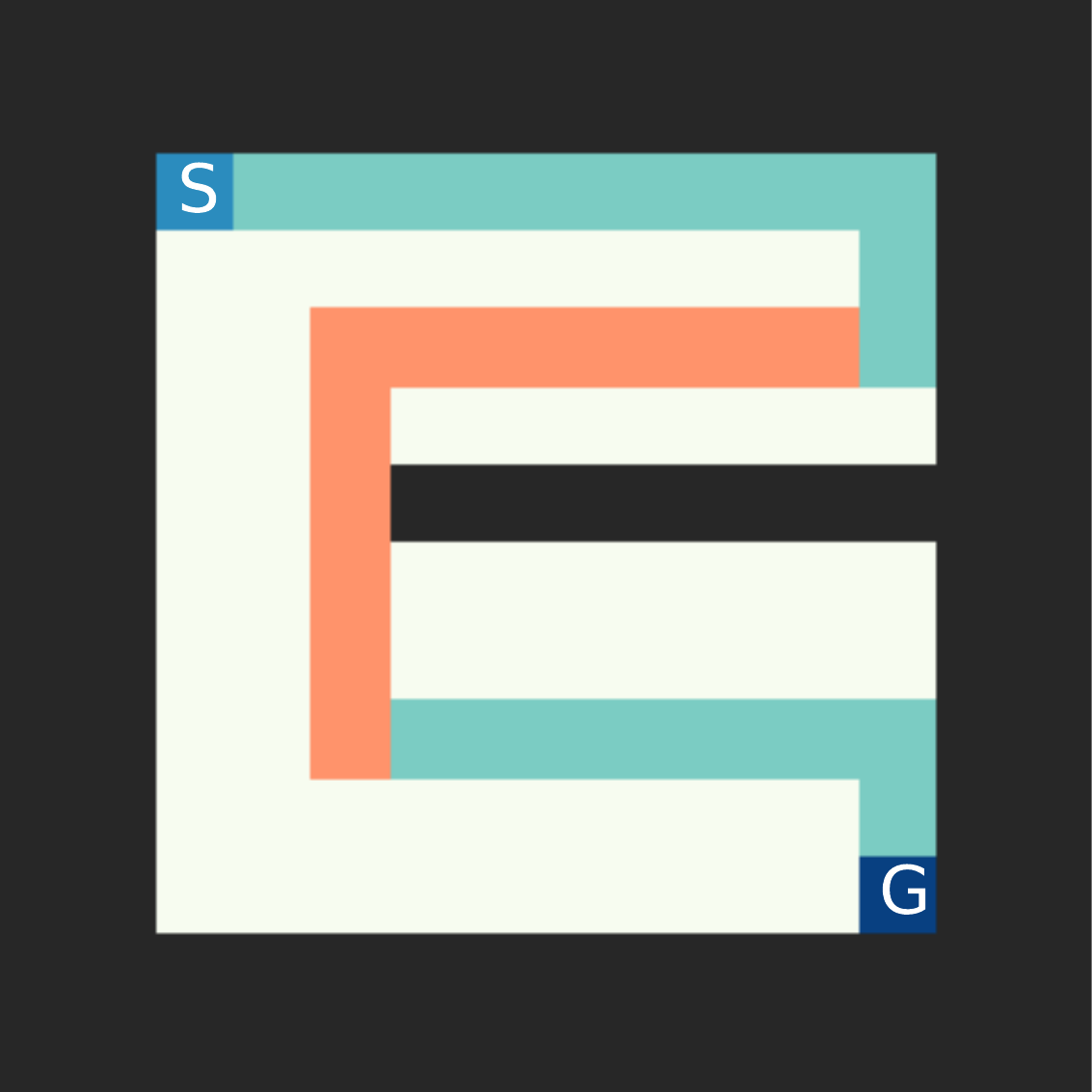}}
  % \includegraphics[width=4.7cm]{s2_img/s_2_obs_config_1_fail_variance.pdf}}
  %\hspace{0.2cm}
  \subfloat[]{
  \includegraphics[width=0.075\textwidth]{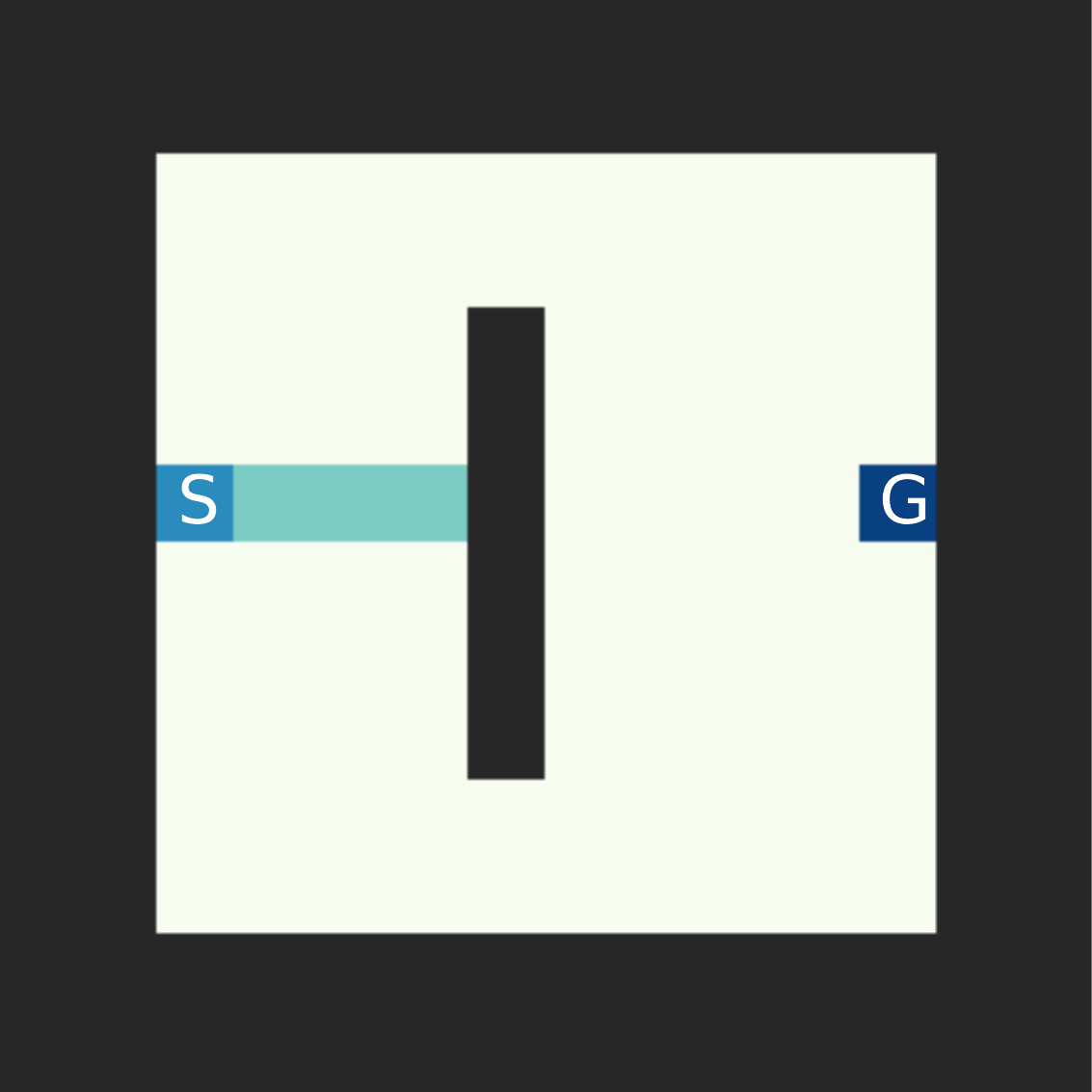}
  %\hspace{0.1cm}
  \includegraphics[width=0.075\textwidth]{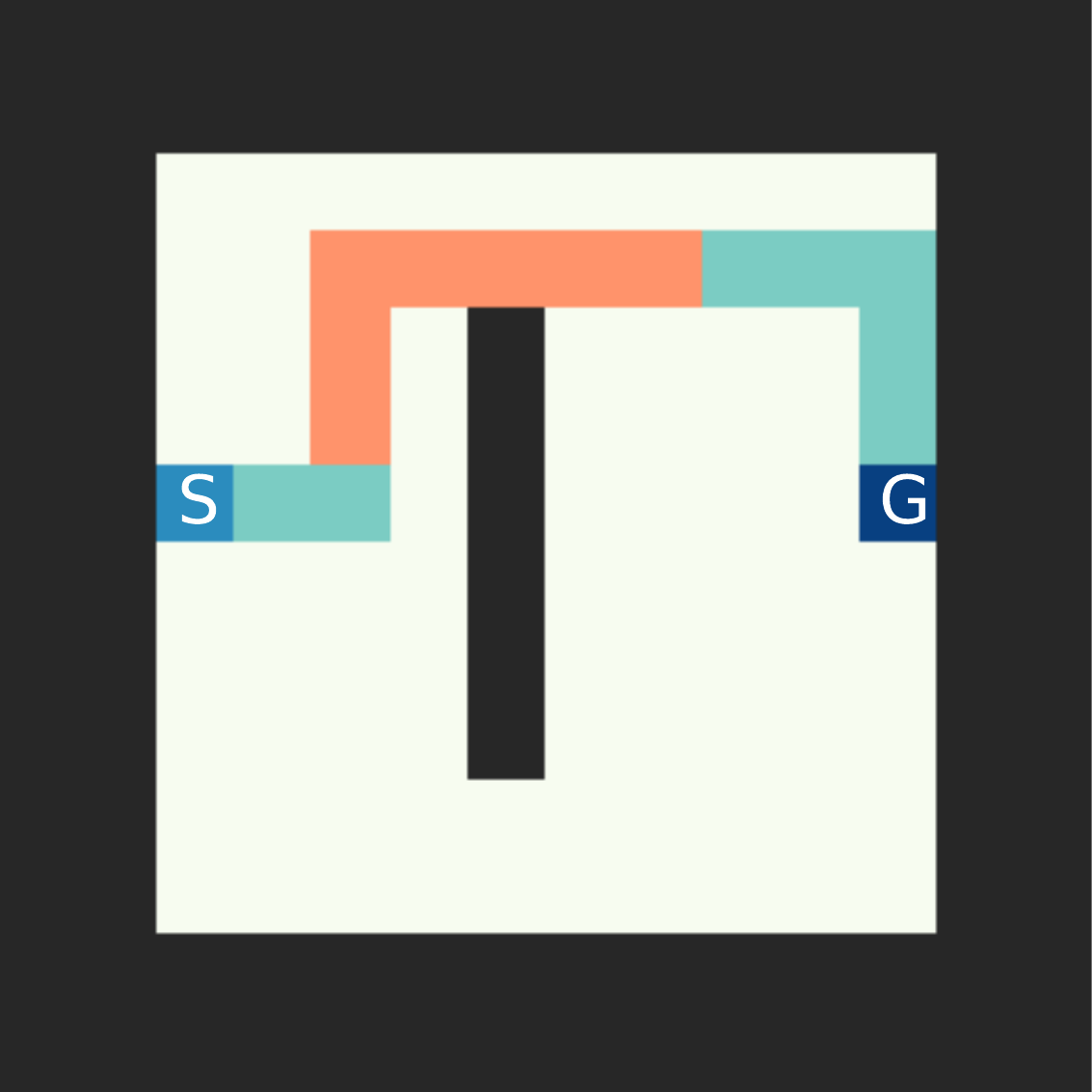}}
  %\\
  \subfloat[]{
  \includegraphics[width=0.075\textwidth]{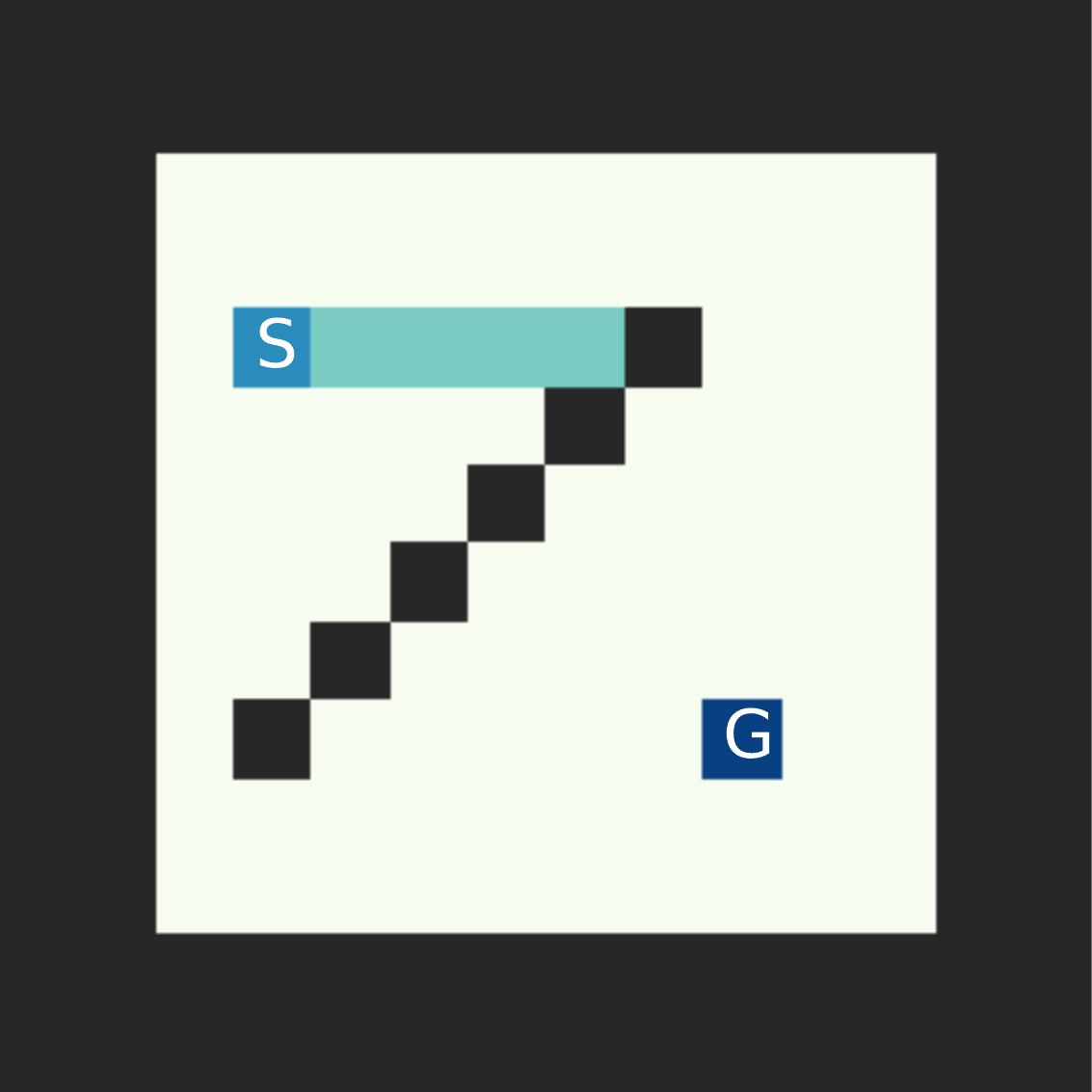}
  %\hspace{0.1cm}
  \includegraphics[width=0.075\textwidth]{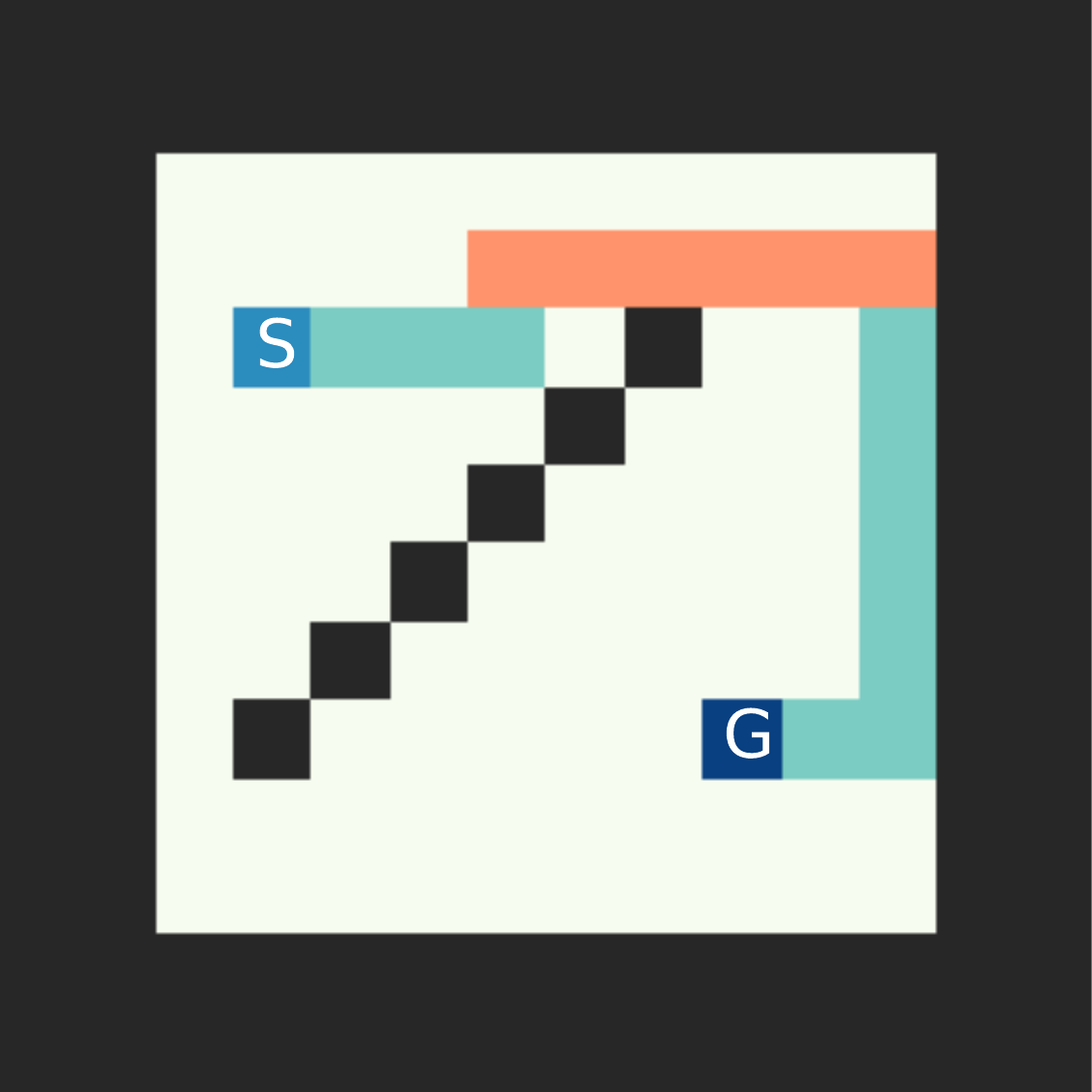}}
  %\hspace{0.2cm}
  \subfloat[]{
  \includegraphics[width=0.075\textwidth]{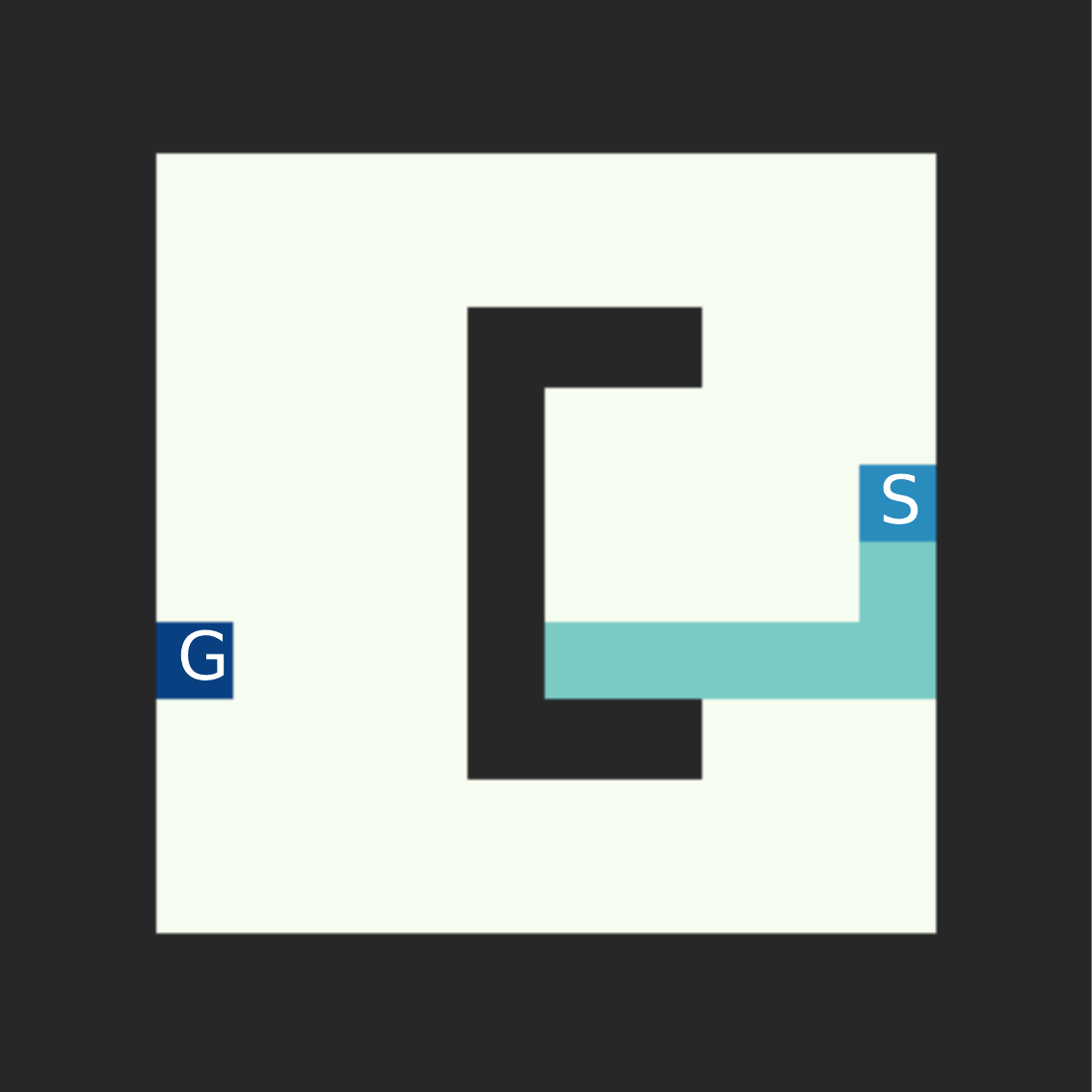}
  %\hspace{0.1cm}
  \includegraphics[width=0.075\textwidth]{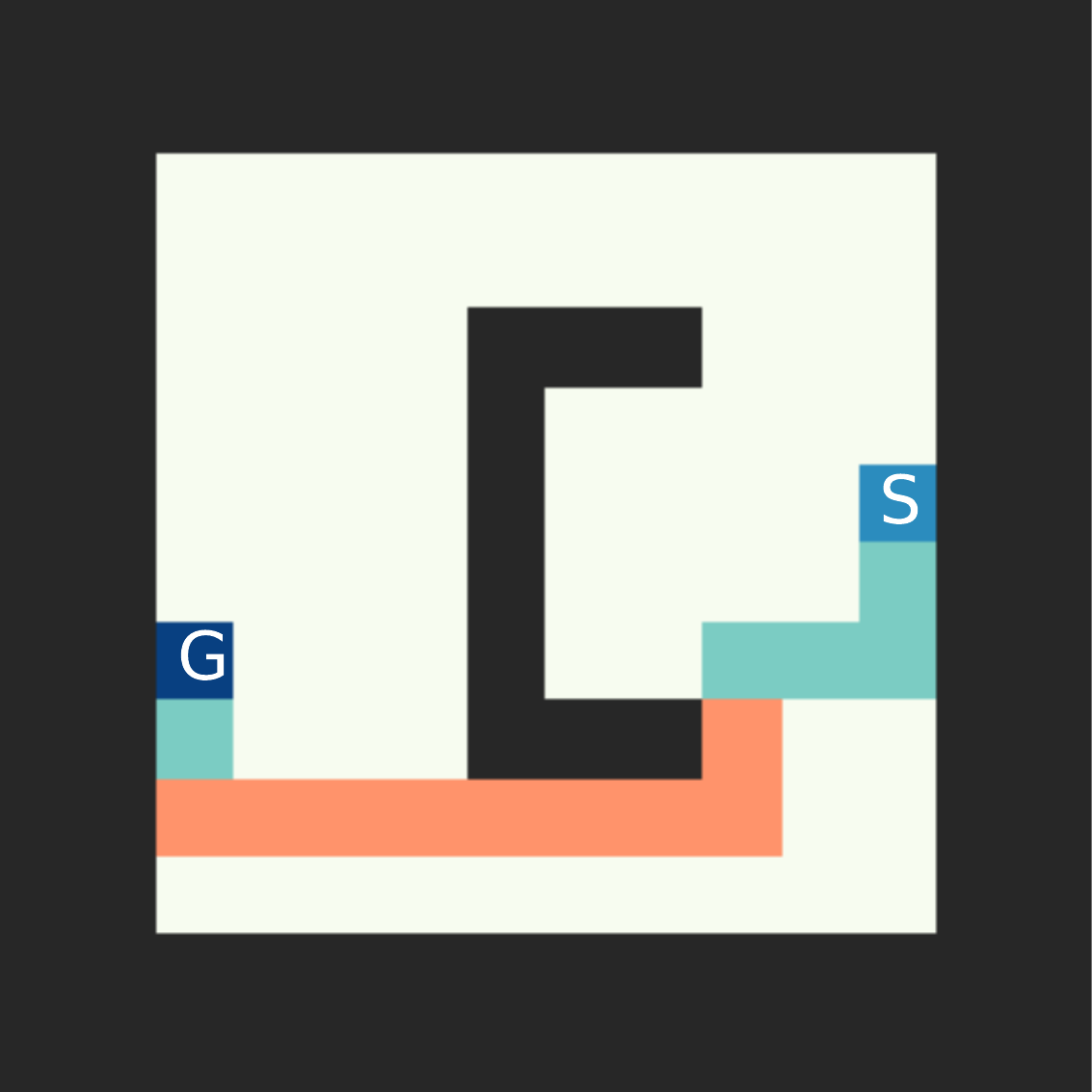}}
  %\\
  \subfloat[]{
  \includegraphics[width=0.075\textwidth]{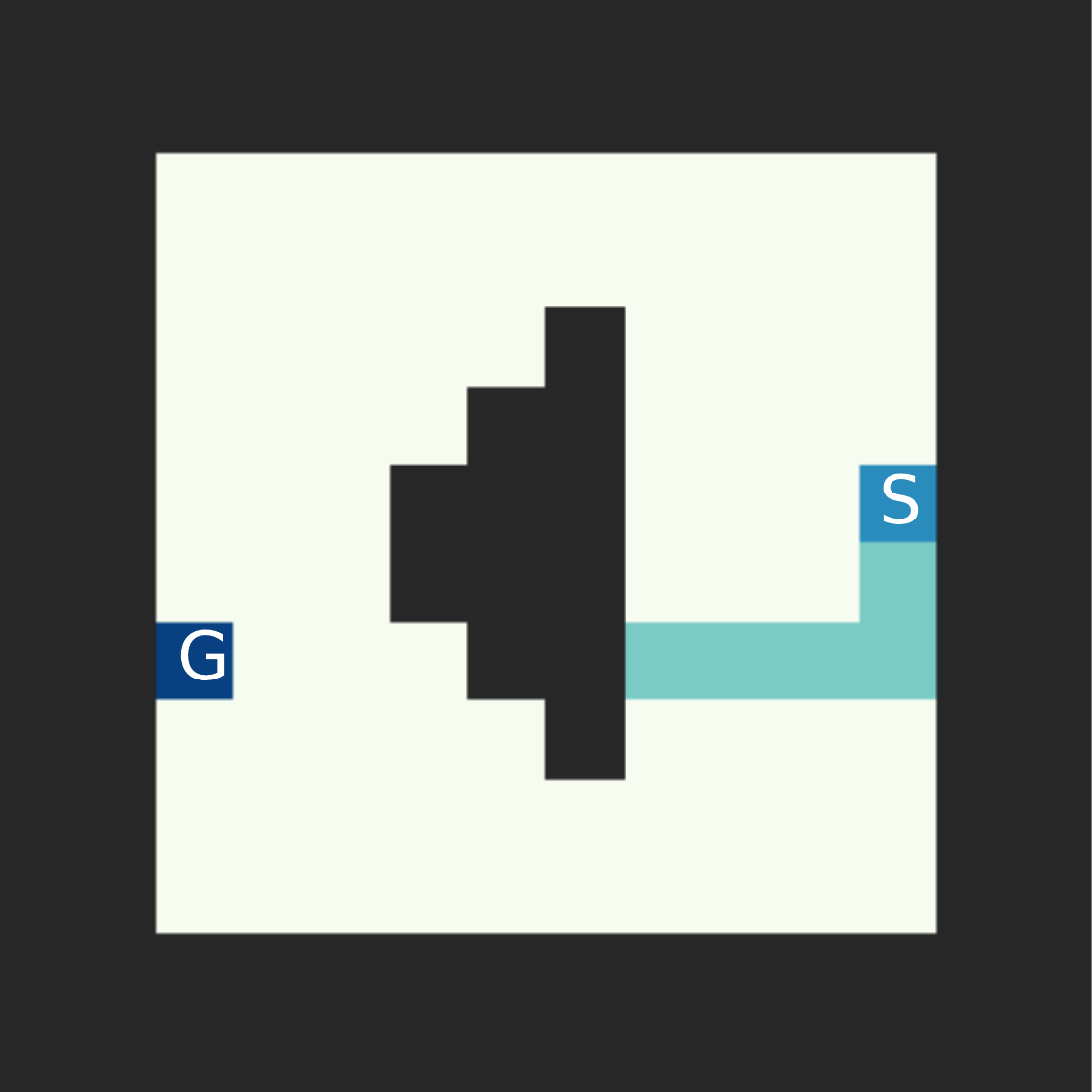}
  %\hspace{0.1cm}
  \includegraphics[width=0.075\textwidth]{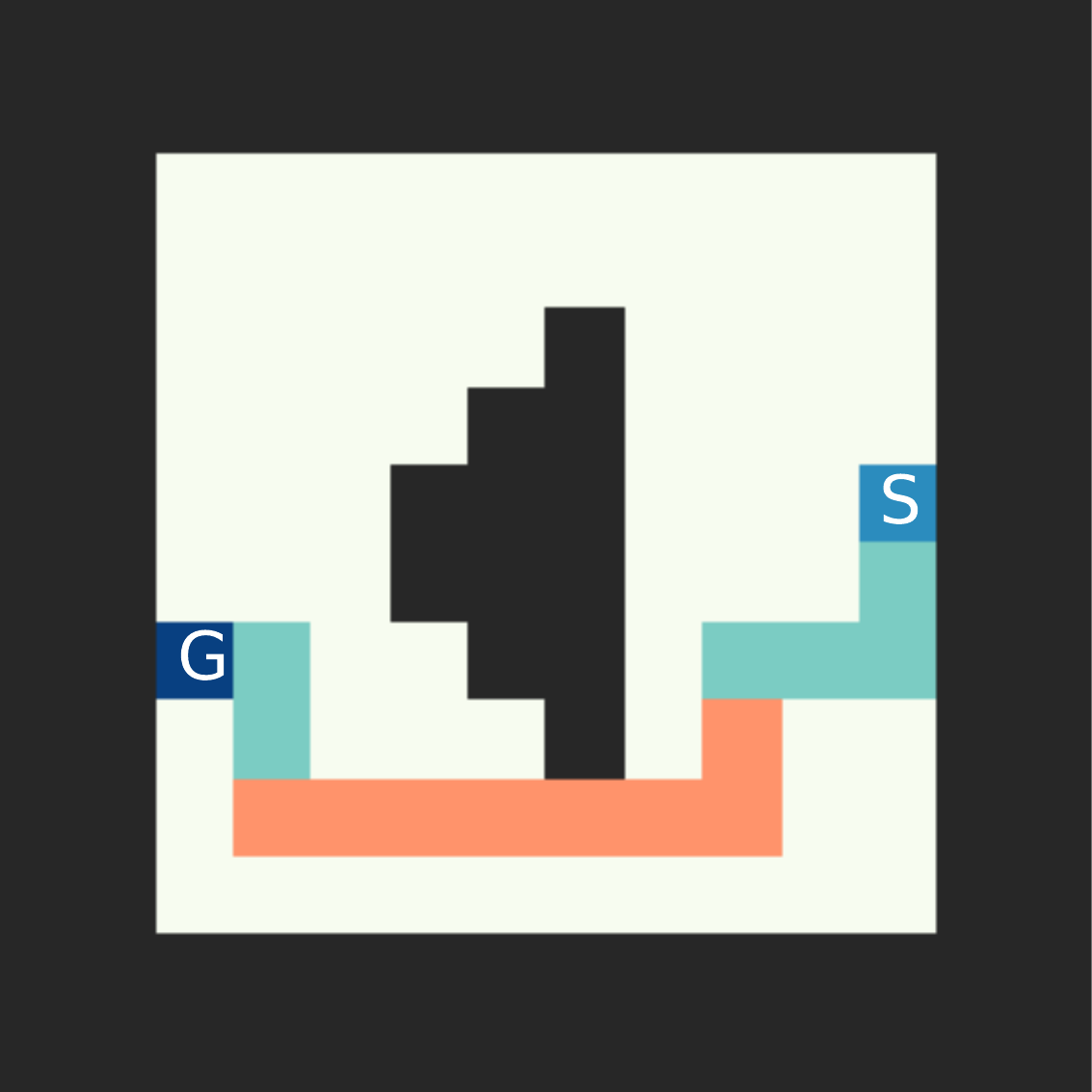}}
  %\hspace{0.2cm}
  \subfloat[]{
  \includegraphics[width=0.075\textwidth]{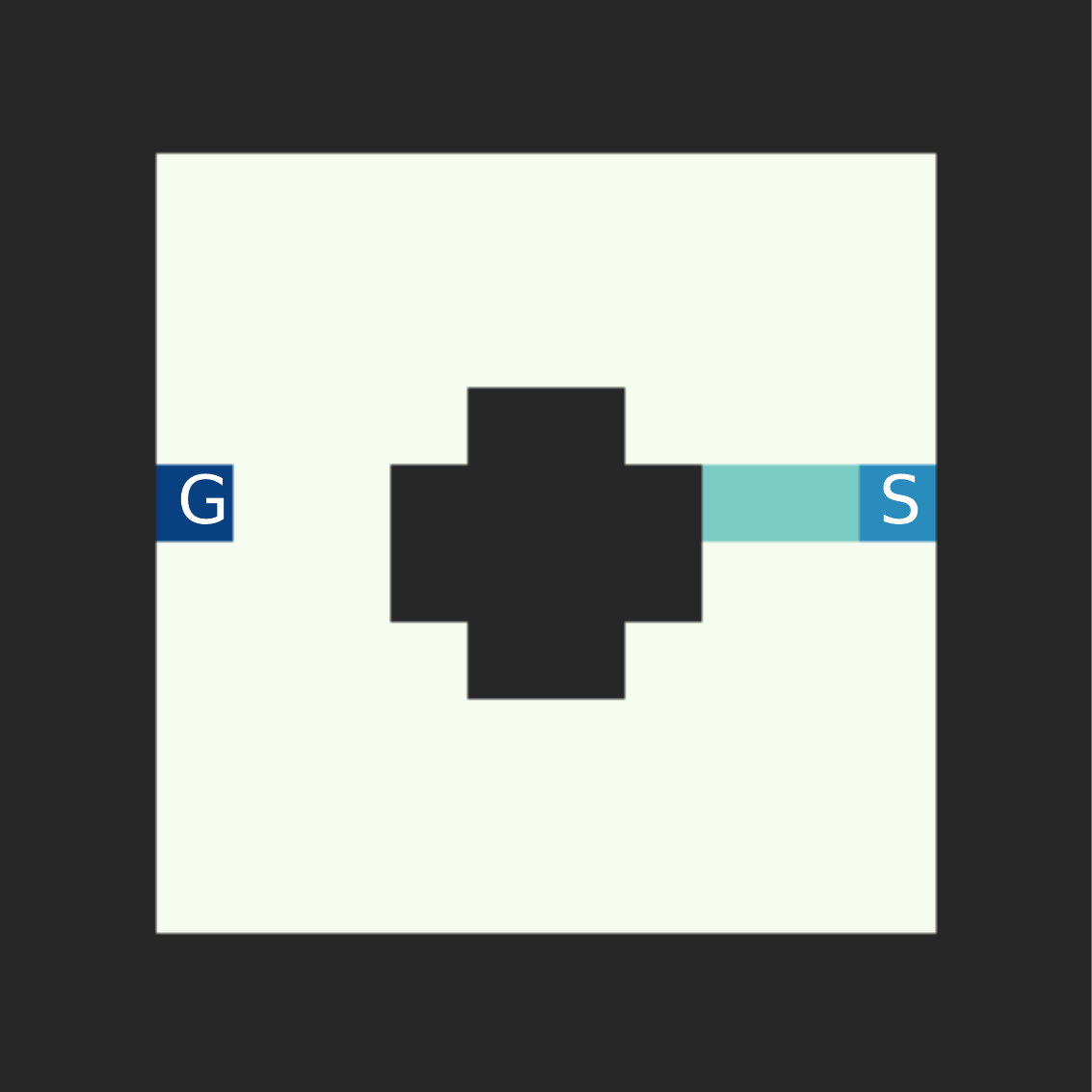}
  %\hspace{0.1cm}
  \includegraphics[width=0.075\textwidth]{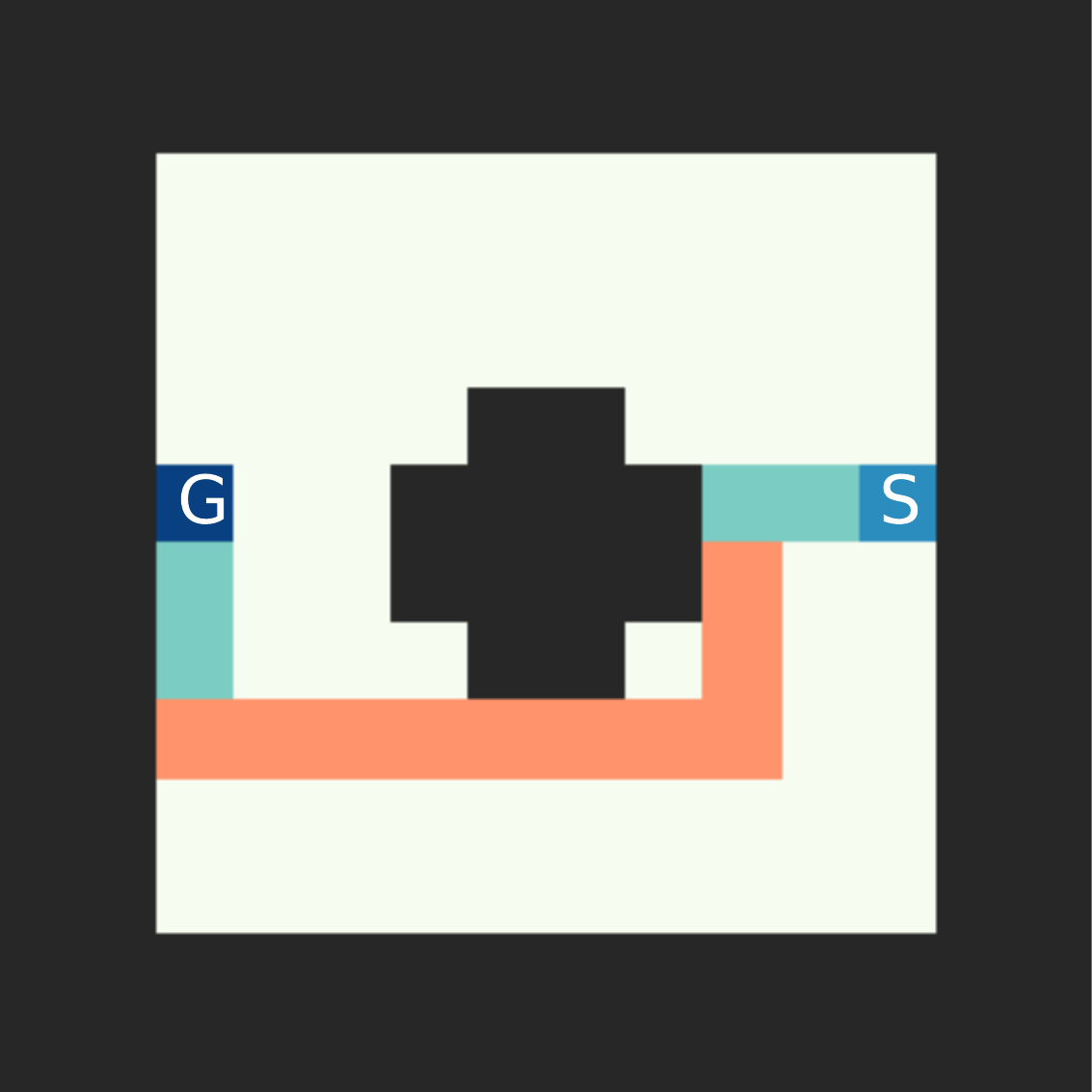}}
    \caption{\small {Testing RE-MOVE with different obstacle configurations and the goal-conditioned observation Space. The left renders show RE-MOVE without human feedback. The right renders show RE-MOVE with human feedback. The orange trajectory shows the human feedback-based path.}}\label{fig:s2_additional}
\end{figure*}

\begin{figure}[t]
    \centering
\includegraphics[scale=0.35]{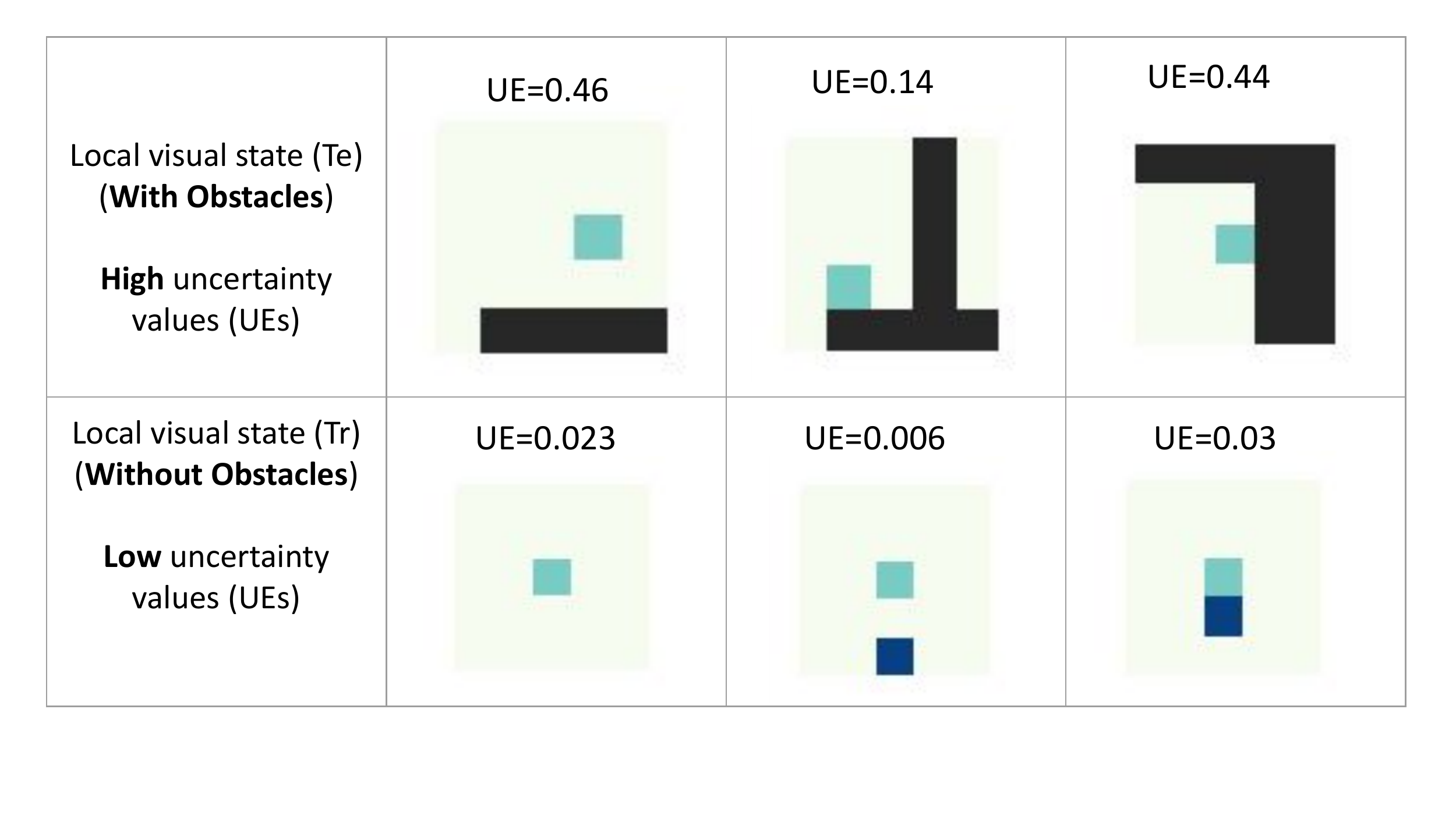}
    \caption{ \small {This figure shows the uncertainty estimates (UE) using MC Dropout for visual navigation tasks with local observations during training without obstacles (\textbf{BOTTOM}) and testing with obstacles (\textbf{TOP}) at various segments of the trajectory. We note that when the local observation image has no obstacles, the uncertainty is low, indicating that the agent policy is confident, whereas during test time, with obstacles/barriers (shown in black in the top row), the uncertainty estimates are high where the policy is underconfident, and the agent asks for help. We want to emphasize that when the robot comes in proximity of the obstacle, then the uncertainty shoots up (10-50) times, and hence, we can utilize this jump for thresholding to ask for human feedback}}
\label{chatGPT_figure}
 \end{figure}

\begin{figure}
    \centering
\includegraphics[width=\columnwidth,height=3.4cm]{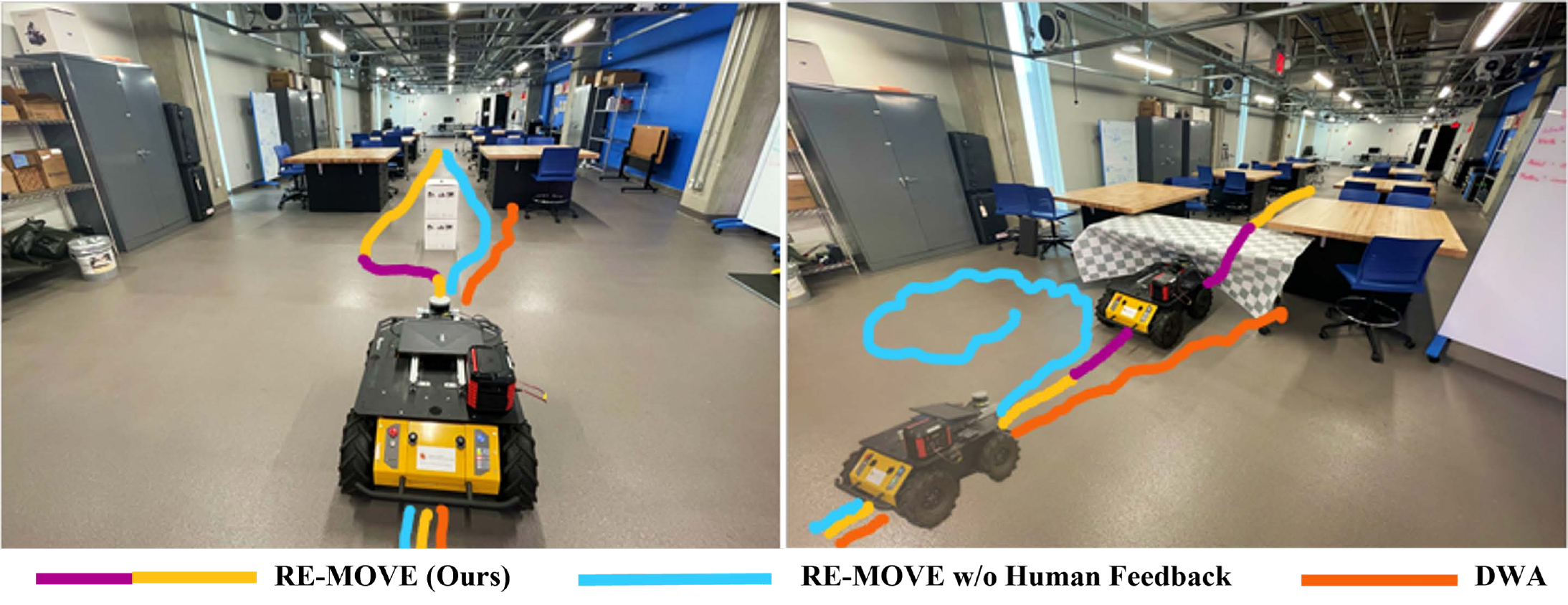}
    \caption{\small{Navigation trajectories generated by RE-MOVE(Ours), RE-MOVE w/o human feedback and DWA\cite{DWA} in real-world environments. \textbf{[Left]:} Scenario 1;\textbf{[Right]:} Scenario 3. We observe that both RE-MOVE and DWA are able to reach the goal by avoiding obstacles and through the free spaces in scenario 1. However, DWA or similar LiDAR-based navigation methods fail in scenario 3 since no free spaces are observable from the LiDAR scan (i.e., leads to freezing). In contrast, our RE-MOVE formulation seeks human assistance to navigate through the pliable region (i.e., through the tablecloth) instead of completely depending on the robot's LiDAR sensor-based observations. Hence, our methods can be generalized to address the perception challenges that occur in real-world environments where sensory observations are ambiguous due to perceptually deceptive objects or obstacles.}}
    \label{fig:real_trajs}
 %   \vspace{-5pt}
 \end{figure}

\subsection{Real-world Experiments}
We deploy our RE-MOVE algorithm designed using the outdoor simulation environment on a real Clearpath Husky robot to evaluate its performance in real-world environments after training and testing with the outdoor simulator described in the previous sub-section. The robot is equipped with a VLP16 LiDAR, and a laptop with an Intel i9 CPU and an Nvidia RTX 2080 GPU.

\noindent \textbf{Evaluations.} We use the following evaluation metrics to compare our method's navigation performance against the existing methods and the ablation studies. We use Dynamic Window Approach (DWA)\cite{DWA} as the baseline comparison method since it is a local planner that utilizes a 2D LiDAR scan for obstacle avoidance. However, any state-of-the-art 2D laser scan-based classical planner can be utilized as the baseline method. We further use RE-MOVE without human feedback as an ablation study to highlight the benefits of our method.

\begin{itemize}
    \item \textbf{Success Rate}: The number of successful goal-reaching attempts without collisions over the total number of trials. 
    \item  \textbf{Normalized Trajectory Length}: The robot's trajectory length is normalized by the straight-line distance to the goal in both successful and unsuccessful trajectories.

\end{itemize}

\noindent \textbf{Test Scenarios.} We consider the following two test scenarios to evaluate our method's navigation performance in real-world environments.

\begin{itemize}
    \item \textbf{Scenario 1:} A large indoor room that includes several obstacles and obstacle-free spaces large enough for the robot to navigate through (see Fig. \ref{fig:real_trajs}[Left]). 

\item \textbf{Scenario 2:} A large indoor room with a long table covered with a tablecloth hiding the ground clearance region of the tables. The room also has several obstacle-free spaces large enough for the robot to navigate through (see Fig. \ref{fig:cover}). 

\item  \textbf{Scenario 3:} The same large room is occupied with more objects and a table covered with clothes. No free space is available for the robot to navigate, with the only option of penetrating through the table clothes and navigating under the table(see Fig. \ref{fig:real_trajs}[Right]). 
\end{itemize}

\begin{figure}
    \centering
    \includegraphics[scale=0.4]{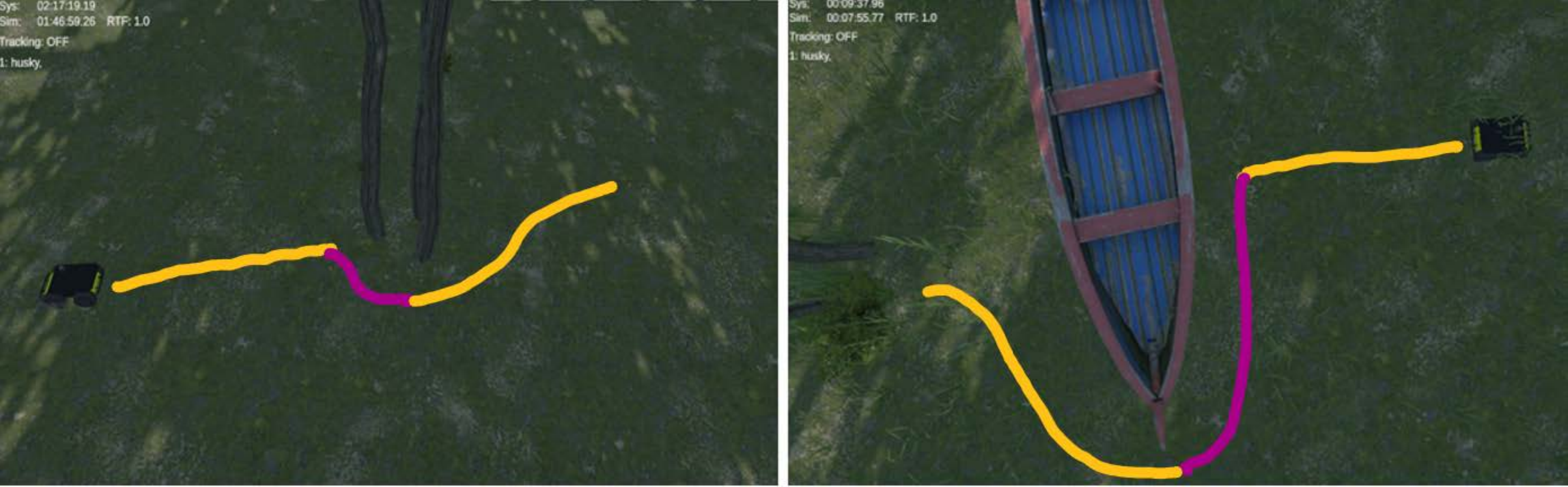}
    \caption{{Trajectories generated by our algorithm in simulated outdoor environments with a Husky robot model. Yellow colored parts indicate the trajectory generated by the Imitation Learning based navigation policy. The violet color part indicates the deviated trajectory after the human feedback. Once the estimated uncertainty is high, the robot asks for human feedback and follows the feedback instructions to avoid possible collisions. After following the instructed actions, the robot continues to reach the original goal location using the navigation policy.}}
    \label{fig:sim_trajs}
   % \vspace{-5pt}
\end{figure}

\noindent \textbf{Analysis and Discussions.} We evaluate our method's performance qualitatively and quantitatively for both synthetic and real-world scenarios. 
 To validate the navigation performance of our method trained using the simulation dataset, we consider two obstacle scenarios in the outdoor simulator as shown in Fig. \ref{fig:sim_trajs}. Both scenarios include solid objects (i.e., a tree and a boat, respectively) that the robot must avoid. We observe that RE-MOVE is able to successfully identify the uncertainties that occurred in the presence of those obstacles and request a human assistant to deviate from the original trajectory (See the violet color portion of the trajectories in Fig. \ref{fig:sim_trajs}).   Next, we deploy this algorithm into a real Husky robot for further evaluation since the outdoor simulator does not include any perceptually deceiving objects for testing.

\begin{table}
\resizebox{\columnwidth}{!}{
\begin{tabular}{ |c |c |c |c | c|} 
\hline
\textbf{Metrics} & \textbf{Method} & \multicolumn{1}{|p{1cm}|}{\centering \textbf{Scenario} \\ \textbf{1}} & \multicolumn{1}{|p{1cm}|}{\centering \textbf{Scenario} \\ \textbf{2}} & \multicolumn{1}{|p{1cm}|}{\centering \textbf{Scenario} \\ \textbf{3}} \\ [0.5ex] 
\hline
{\textbf{SR}}
 & DWA \cite{DWA} & 100 & 100 & 0 \\
 & RE-MOVE w/o feedback  & 40 & 30  & 10 \\
 & RE-MOVE (ours)  & \textbf{100} & \textbf{90} & \textbf{80} \\
\hline
{\textbf{TL}} 
 & DWA \cite{DWA}  & 1.241 & 1.311 & 0.672\\
 & RE-MOVE w/o feedback  & 0.951 & 0.863  & 0.769\\
 & RE-MOVE (ours) & \textbf{1.298} & \textbf{1.134} & \textbf{1.093}  \\
\hline
\end{tabular}
}
\caption{\small{\textbf{Navigation Performance Comparisons (SR denoted success rate (in \%) and RL denotes trajectory length):} We observe that the classical obstacle avoidance algorithms, such as DWA can perform well in environments with free spaces (e.g., scenario 1). However, such methods cannot navigate well under dense and perceptually deceiving environments such as Scenario 2 and 3 due to misleading perceptions. In contrast, RE-MOVE is able to perform well consistently in all three scenarios.}
}
\label{tab:comparison_table}
%\vspace{-12pt}
\end{table}

 We present our real-world experiment results qualitatively in Fig. \ref{fig:real_trajs} and Fig.\ref{fig:cover}, and quantitative results in Table \ref{tab:comparison_table}. We observe that both DWA and RE-MOVE perform equally well in Scenario 1 due to the availability of obstacle-free spaces for the robot to navigate. However, with the presence of a table covered with a cloth (i.e., a deceiving object to the LiDAR 2D scan) in Scenario 2, DWA takes a significantly longer trajectory to avoid all the obstacle regions that appeared in the cost map. In contrast, RE-MOVE identifies the tablecloth region as highly uncertain and requests human feedback to proceed to the goal. With assistance from the human expert, RE-MOVE continues to navigate through the hanging cloth which results in a significantly lower normalized trajectory length compared to the DWA. However, RE-MOVE's success rate is slightly lower than the DWA in Scenario 2 due to the errors in uncertainty estimation and language-to-sequence translation.  
 We further notice that DWA cannot reach the goal in Scenario 3 since it considers both deceiving and solid objects as obstacles from the LiDAR scan. On the other hand, RE-MOVE is able to proceed to the goal with the assistance of a human expert through the only possible path available in the environment (i.e., through the tablecloth). Please refer to \cite[Appendix \ref{additional_real_world}]{chakraborty2023re}.

\section{Conclusions and Future Work}
We proposed a novel approach called RE-MOVE (REquest and MOVE on) to enable reinforcement learning policies to exhibit test time adaptability. RE-MOVE is able to identify real-time objects/obstacles which were not present during the training, and then utilize language-based human feedback to adapt its policy.  We optimize the feedback by designing a novel decision function based on the uncertainty  quantification of model parameters. We have shown the efficacy of our proposed approach on a variety of synthetic and real-world navigation tasks. As a valid future scope, it would be interesting to develop a dedicated language model for the robotic navigation task and also incorporate feedback in terms of speech or images as well. 

\bibliographystyle{IEEEtran}
\bibliography{iclr2021_conference}

\clearpage
\onecolumn
\appendices

\section{{Discrete Grid world Experiments in Details}}\label{discrete_grid}

Figure \ref{fig:env_sample} pictorially represents the grid world environment (without obstacles) that is used for this experiment. We consider a two-dimensional grid world with dimensions $(L+4)\times(L+4)$, where the central portion of the grid, measuring $L\times L$, is accessible to agents, while the outer regions are impassable walls. As illustrated in Figure \ref{fig:env_sample}, the start and goal cells are marked as "S" and "G", respectively, with the agent's current position denoted as "A". The objective of the agent is to reach the goal "G" from the start "S" by taking the shortest possible path. 

As mentioned in Section \ref{sec:observation_space}, we test our algorithm using two observation space configurations. In the discrete grid world, the global observation is a $(L+4)^2$ dimensional vector $O^G_t = [G^{1,1}_t, G^{1,2}_t, \cdots,G^{(L+4),(L+4)}_t]^T$ where $G^{i,j}_t$ represents the value of the cell at the $i^{th}$ row and $j^{th}$ column of the grid world at time $t$. The values of the cells of the grid world are set as follows: $G^{i,j}_t = 0$ if the cell is unoccupied, $G^{i,j}_t=10$ if the agent occupies the cell, $G^{i,j}_t=20$ if the cell is the goal and $G^{i,j}_t=30$ if an obstacle or wall occupies the cell. $O^G_t$ is essentially the flattened image of the entire environment (where each pixel represents a cell of the grid). The local observation is obtained by taking a $L'\times L'$ s.t. $L'<L$ subgrid around the agent's location, see Figure \ref{fig:env_sample}.

\subsubsection{Generating Demonstration Data and Imitation Learning Policy}

To determine the shortest path, we employ Dijkstra's algorithm. The resulting trajectory is then incorporated into the expert demonstration as part of the model. An expert demonstration, denoted as $\mathcal{D}$, is a collection of trajectories $\tau$, each of which is generated by recording the actions taken by an expert user to solve the task. Thus, $\mathcal{D}=(\tau_1, \tau_2,\cdots,\tau_N)$, where $\tau_i=(s_0,a_0,s_1,a_1,\cdots,s_T,a_T)$.
In this paper, we generate two sets of expert demonstrations, denoted as $\mathcal{D}_1$ and $\mathcal{D}_2$, that operate on the \textit{Global Visibility Observation} and the \textit{Goal-Conditioned Observation} respectively. Using these demonstrations, we train two Random Forest Tree Classifiers as our policies. Each policy is generated by an ensemble of 500 trees with a maximum depth of 50.
\newline
\subsubsection{Global Visibility Observation space vs. Goal-Conditioned Observation space}

We present the trajectory of an agent in an obstacle-less environment, trained using the imitation policy for the Global Visibility Observation space, in Figure \ref{fig:s1_no_obs}. The figure reveals that the policy's prediction has some uncertainty even when no obstacles exist. We also observe that in Figure \ref{fig:s1_obs}, the uncertainty level is highest during the initial stages of the trajectory when the agent is far from any obstacle. Interestingly, the uncertainty momentarily decreases when the agent is close to the obstacle, which is a counterintuitive observation. We hypothesize that this behavior is due to the fact that the global observation vector always contains information on unseen obstacles. During training, the policy was trained on global observations without any obstacles, resulting in high uncertainty at all times when there are obstacles in the environment, as the global observation vector is new to the policy at all times. 

\subsubsection{RE-MOVE with Human Feedback}

We evaluate the performance of the RE-MOVE algorithm in multiple environmental settings with various obstacle configurations. In particular, we present the results of applying RE-MOVE to navigate an agent in an environment with two distinct obstacle configurations in Figure \ref{fig:goal-cond-new-conf}. Without any feedback from the human expert, the agent gets stuck behind the obstacle, as shown in Figures \ref{fig:s2_obs_fail_2} and \ref{fig:s2_obs_new_fail}, and the uncertainty in the predictions increases concurrently. We use a changepoint detection algorithm \cite{Fearnhead2006} to detect these changes in the prediction uncertainty and prompt the human expert for feedback. The resulting trajectory from the human feedback is presented in Figures \ref{fig:s2_obs_jit} and \ref{fig:s2_obs_new_jit}, where the orange segment of the trajectory is generated based on the feedback.

\section{Visual grid world}
\begin{figure}[htp]
    \centering
    \subfloat[]{%
        \includegraphics[width=0.4\textwidth]{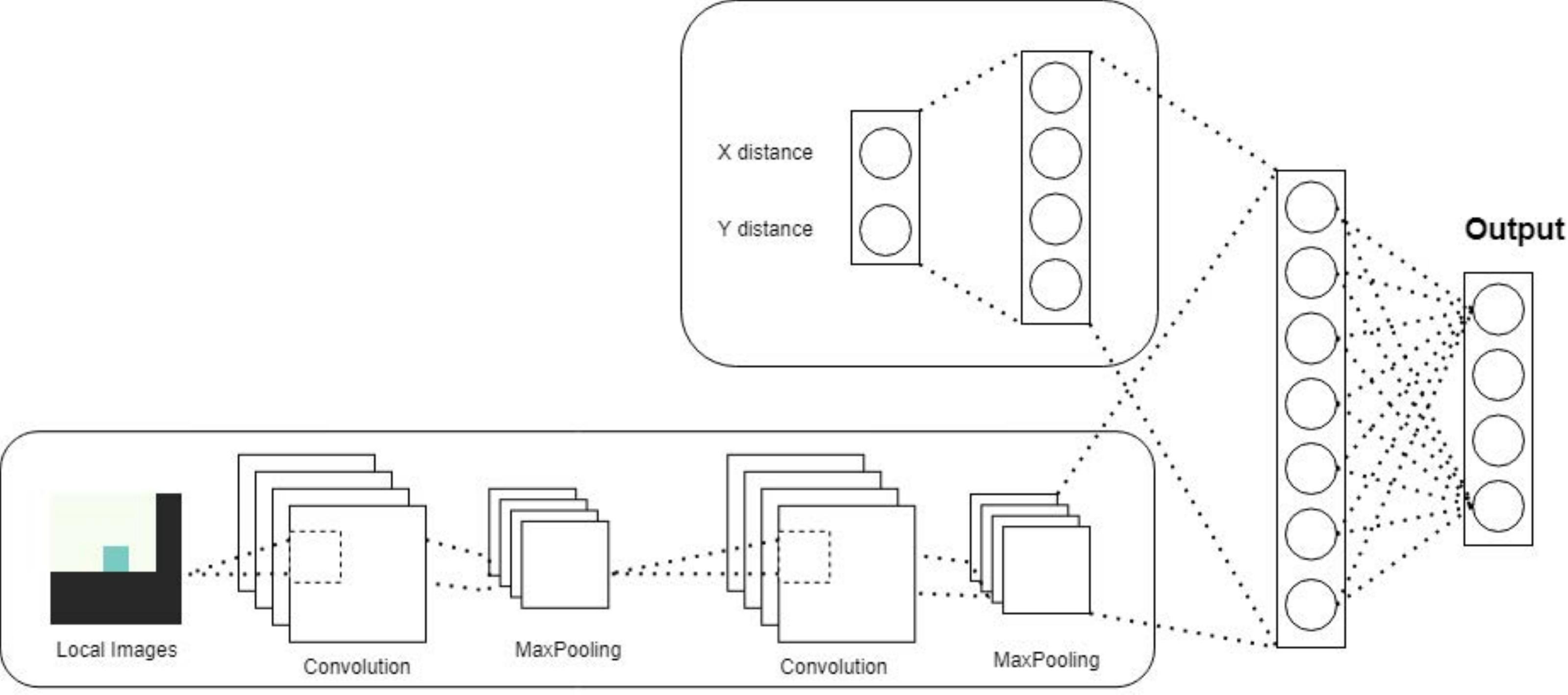}
    }
    \hspace{5mm} 
    \subfloat[]{%
        \includegraphics[width=0.4\textwidth]{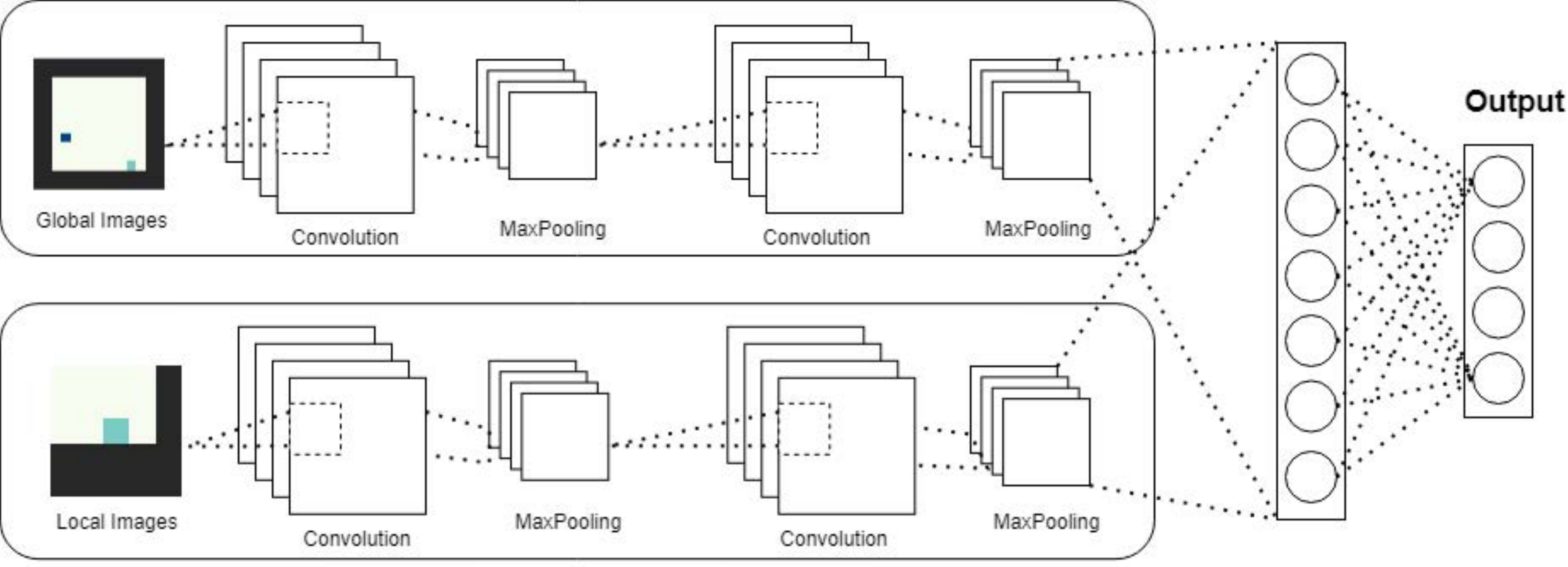}
    }
    \caption{(a) Goal-Conditioned: Architecture uses CNN for visual inputs and a dense layer for distance-to-goal, merging both through fully connected layers for final outputs. (b) Global Observation: Architecture processes both global map and local observations through separate CNN branches, then merges via dense layers.}

    \label{fig:arch}
\end{figure}

\begin{figure}[htp]
    \centering
    \subfloat[]{%
        \includegraphics[width=0.4\textwidth]{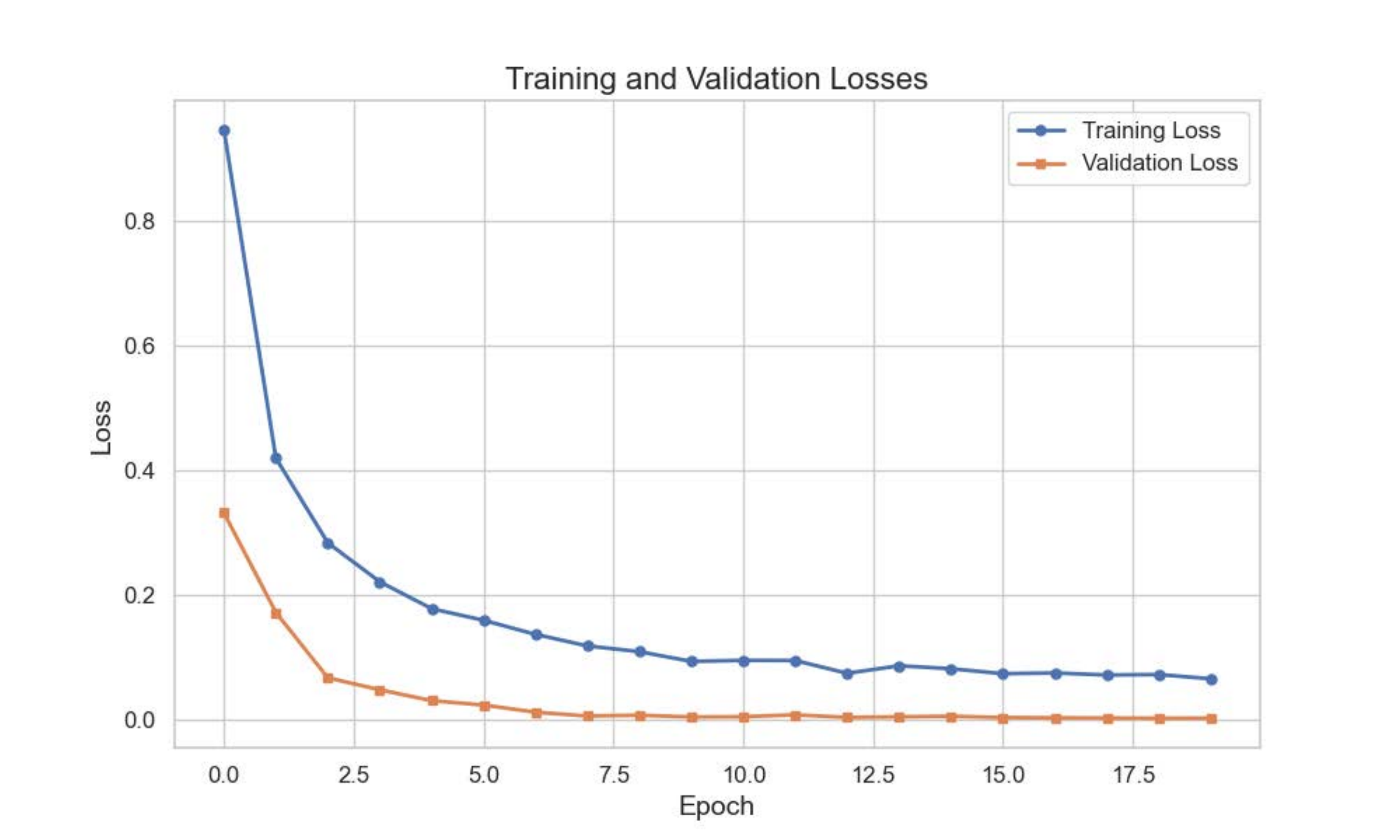}
    }
    \hspace{5mm} 
    \subfloat[]{%
        \includegraphics[width=0.4\textwidth]{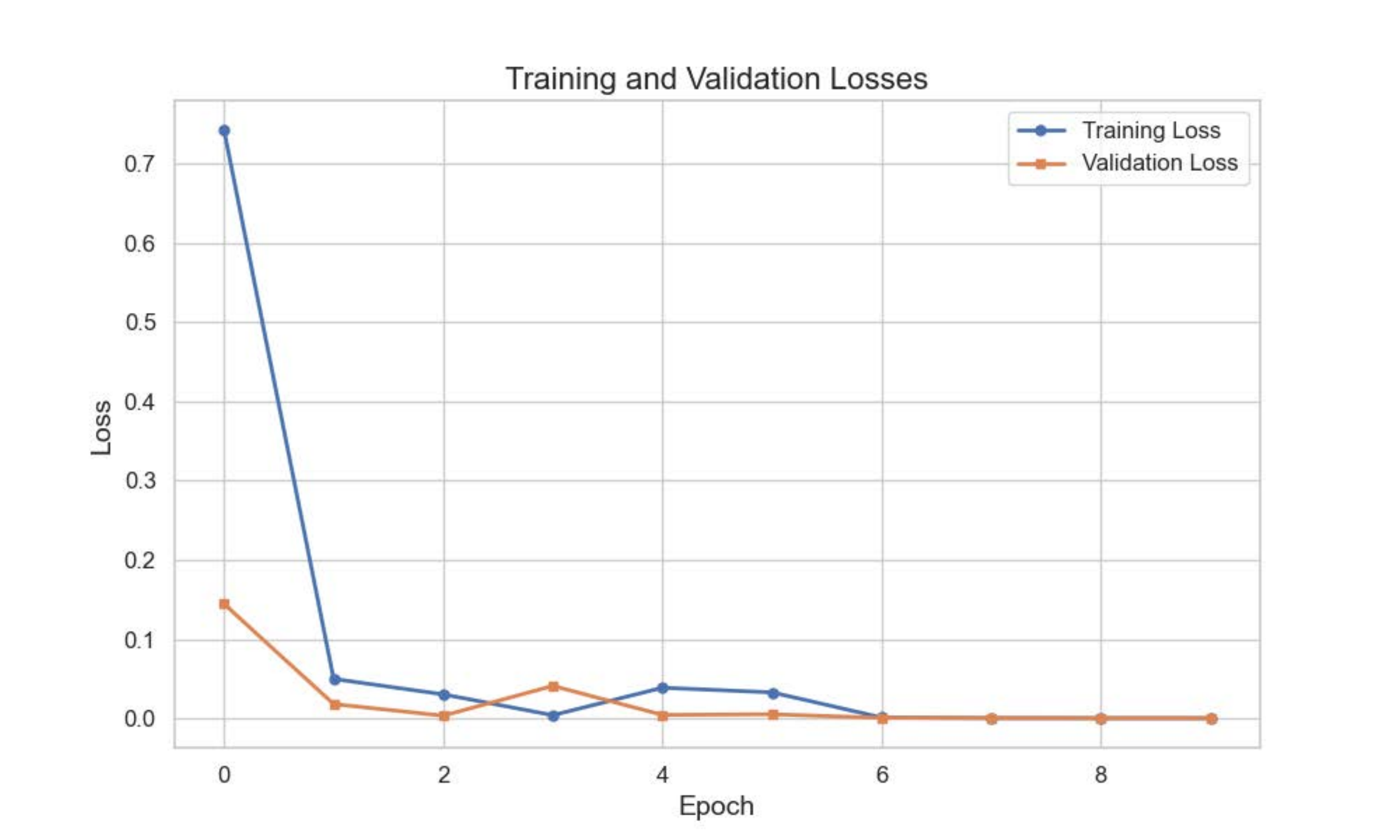}
    }
    \caption{Convergence of Training Loss for Reinforcement Learning Architectures. (a) Goal-Conditioned: Demonstrates loss convergence, validating the architecture's capability to integrate visual inputs and distance-to-goal via CNN and dense layers. (b) Global Observation: Showcases converging loss, confirming the model's ability to process and combine global and local observations effectively}

    \label{fig:model_loss}
\end{figure}

In the visual grid world, we replace tabular states with image-based inputs, adding complexity and making the setup more reflective of real-world conditions. This environment is particularly useful for evaluating image-based navigation policies. 
We transformed the tabular grid worlds into 100x100 pixel images to render our RE-MOVE framework compatible with real-world scenarios. To address the challenges if partially observable information in complex environment, we designed two distinct CNN architectures for two different scenarios—\textit{goal-conditioned} and \textit{global observation}. 
As shown in figure \ref{fig:arch}, the goal-conditioned model comprises a dual-branch design, efficiently integrating local observations and distance-to-goal values. Conversely, the global observation model handles both global maps and local observations, combining them to facilitate informed decision-making. Further validation of these models' effectiveness is evident from Figure \ref{fig:model_loss}, which displays the convergence of training loss for both scenarios.

\section{Simulated outdoor environment}\label{simulation_environment}
We incorporate a high-fidelity outdoor simulator to train an imitation learning-based navigation policy using a set of expert demonstrations. The simulator does not include perceptually deceiving objects such as thin pliable grass, which the 2D LiDAR scan can detect as obstacles. Similar to the grid world environment, we consider a robot-centric, $L'\times L'$ dimensional, 2D local cost map $C_t^L$ (i.e., a local grid map) and the robot's current distance to the goal $d_{goal}$ as local observations $O_t^L$. The grid map combines two $L'\times L'$ dimensional layers; 1. Obstacle layer ($\ell^{scan}$) from 2D LiDAR scan data; 2. Distance layer ($\ell^{dist}$) that reflects the distance to the goal information for each grid point of the grid map. We obtain the obstacle layer as an occupancy grid map using the gmapping\cite{gmapping} ROS package. The distance layer is obtained as follows.
Let $\ell_{i,j}$ be the grid point located in $i$ th row and $j$ the column of the distance layer $\ell^{dist}$. Then, $ \ell^{dist}_{i,j} = d_{i,j}$, where $d_{i,j}$ denotes the distance from the grid point $\ell_{i,j}$ to the goal location. We generate the final local cost map $C_t^L$ (i.e., local observation map) as a linear combination of the aforementioned two layers. Hence, $C_t^L =  \alpha_1 \times \ell^{scan} + \alpha_2 \times \ell^{dist}$, where $\alpha_1$ and $\alpha_2$ are tunable parameters. In obstacle-free settings, the distance layer dominates the local cost map.

\subsection{Generating the Low Dimensional Observations}
To reduce the high dimensionality of the 2D cost map $C_t^L$, we incorporate Principal Component Analysis (PCA)\cite{jolliffe2002pca} to obtain a reduced dimensional vector $V_t^{pca} = [v_t^{pc_1},v_t^{pc_2},..,v_t^{pc_k}]$ of length $k$. Finally, the obtained vector is concatenated with the robot's current distance value to the goal $d_{goal}$ to obtain the observation vector $O_t^L = [V_t^{pca},d_{goal}]$ of length $k+1$.

\subsection{Expert Navigation Data Collection and Imitation Learning Policy}
The expert demonstrations (local observations $O_t^L$ and corresponding actions $a_{t+1}$) are collected for a goal-reaching task in an obstacle-free environment of the simulator. In this context action, $a_t$ consists of linear and angular velocities $v_t,\omega_t$ of the robot (i.e.,$a_t = (v_t,\omega_t)$ ). During data collection, we consider each goal-reaching task as a trajectory $\tau$ similar to the grid world environment. Hence, for each trajectory, a human expert teleoperates the simulated Husky robot model to a random goal located up to 20 meters away from the robot's starting position. Then, we record the local cost map $C_t^L$ and the robot's current distance to the goal $d_{goal}$ information and the subsequent action in the next time step (i.e., $a_{t+1} = (v_{t+1}, \omega_{t+1})$) for each trajectory. We denote the resulting dataset as $\mathcal{D}_{sim}$.

\begin{figure}
    \centering    \includegraphics[scale=0.4]{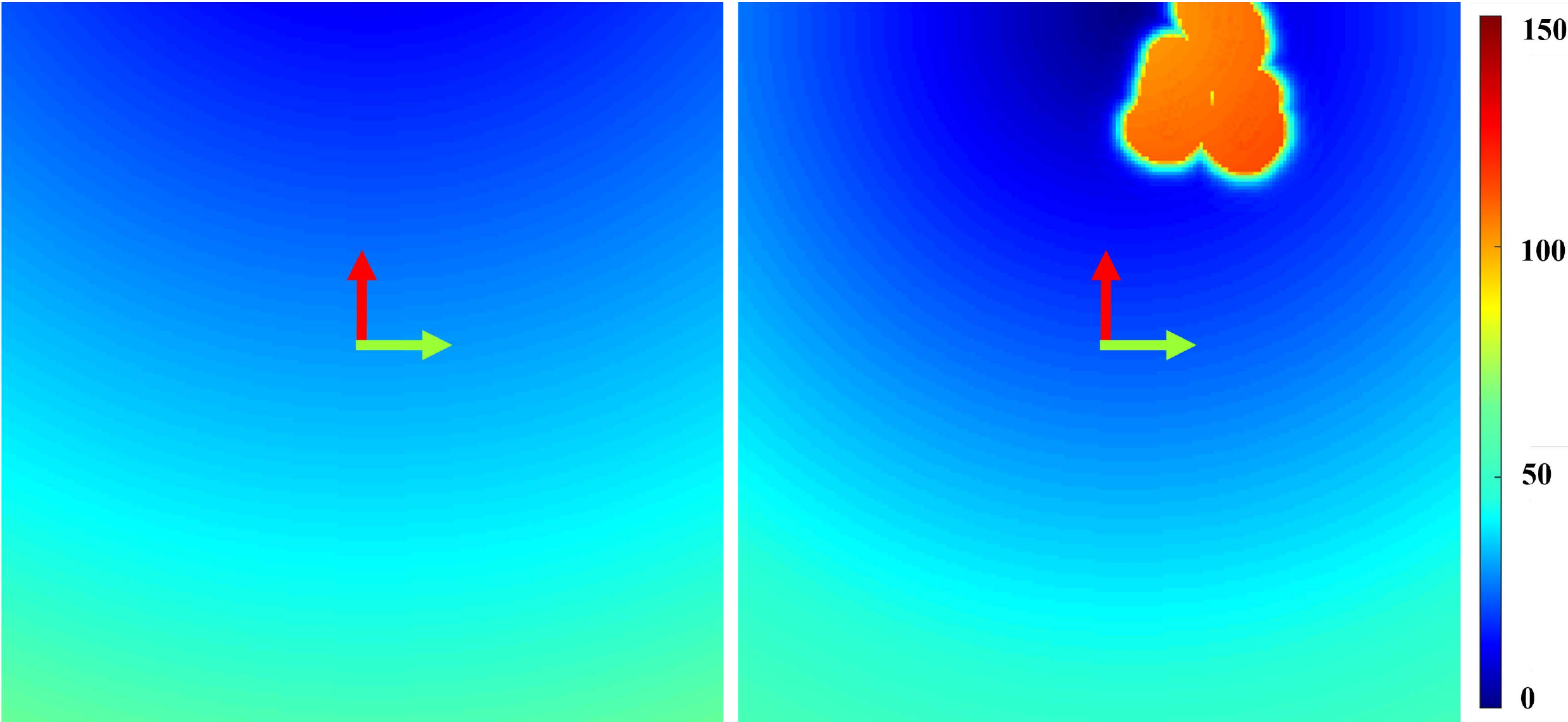}
    \caption{{Examples of local costmap $C_t^L$ obtained by combining laser scan based obstacle layer $\ell^{scan}$ and distance layer $\ell^{dist}$} in the simulated outdoor environment. Obstacles detected from the 2D laser scan lead to significantly high-cost regions (e.g., orange color blobs in the right-side image) in the local cost map. Our uncertainty estimation is capable of identifying such instances so that the robot can request assistance from a human. These estimations are significantly different from the standard obstacle detection in a cost map where the high-cost regions will always be avoided as obstacles without further evaluation (leads to robot freezing in dense obstacle scenarios). }
    \label{fig:sim_costmaps}
  %  \vspace{-15pt}
\end{figure}
 
\section{RE-MOVE with the Partial Global Visibility Observation Space}

\begin{figure}[H]
    \centering
  \subfloat[RE-MOVE without human feedback]{ \includegraphics[width=7cm]{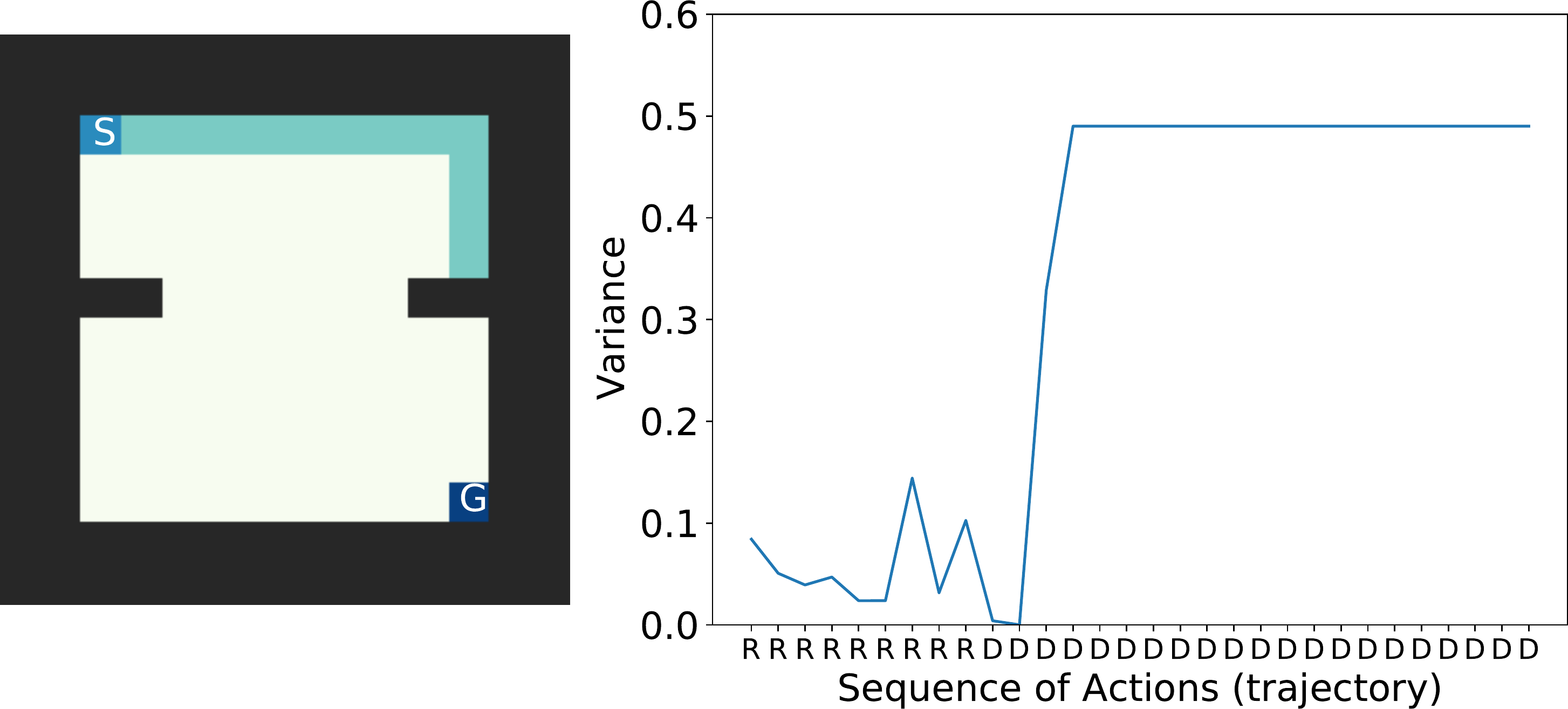}}
  \hspace{0.5cm}
  \subfloat[RE-MOVE with human feedback]{
  \includegraphics[width=7cm]{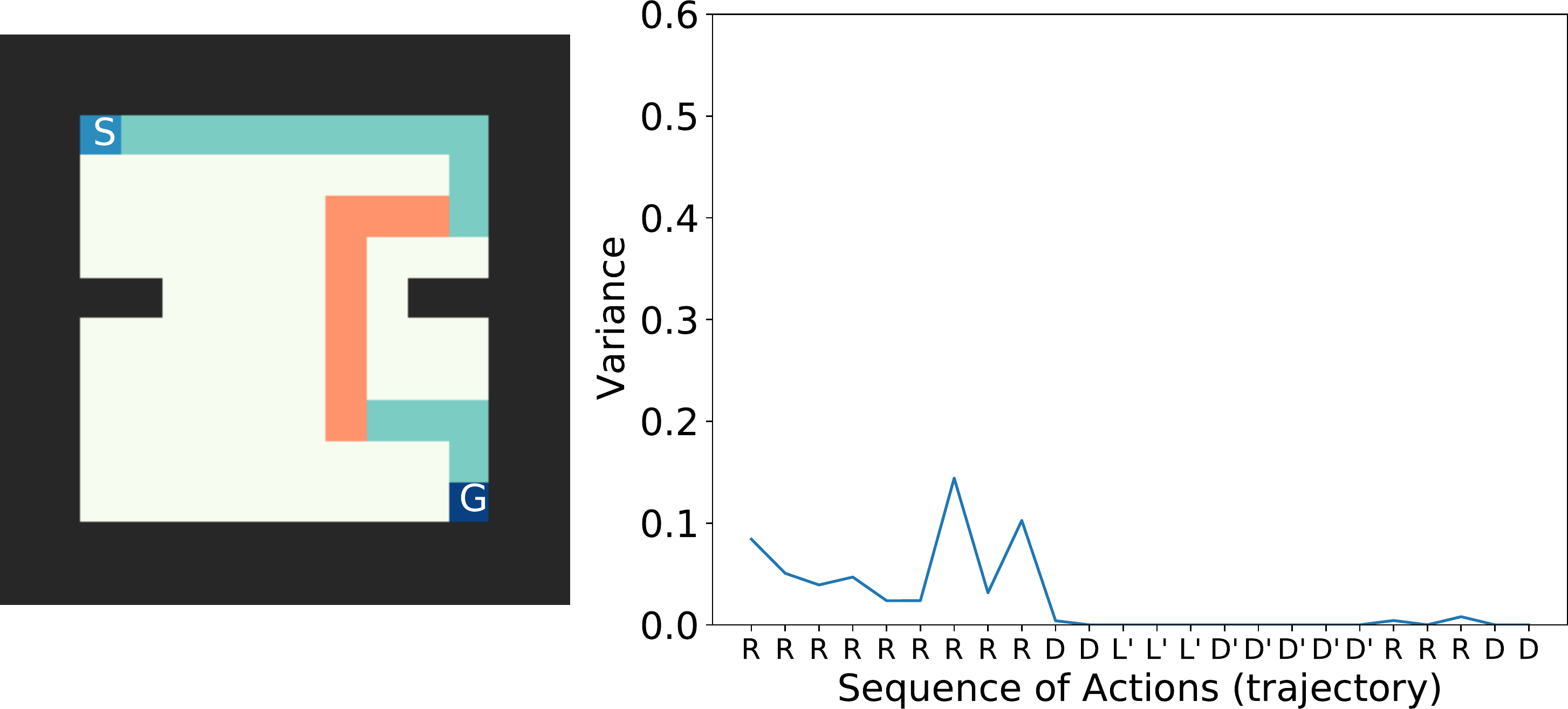}}
\\
  \subfloat[RE-MOVE without human feedback]{ \includegraphics[width=7cm]{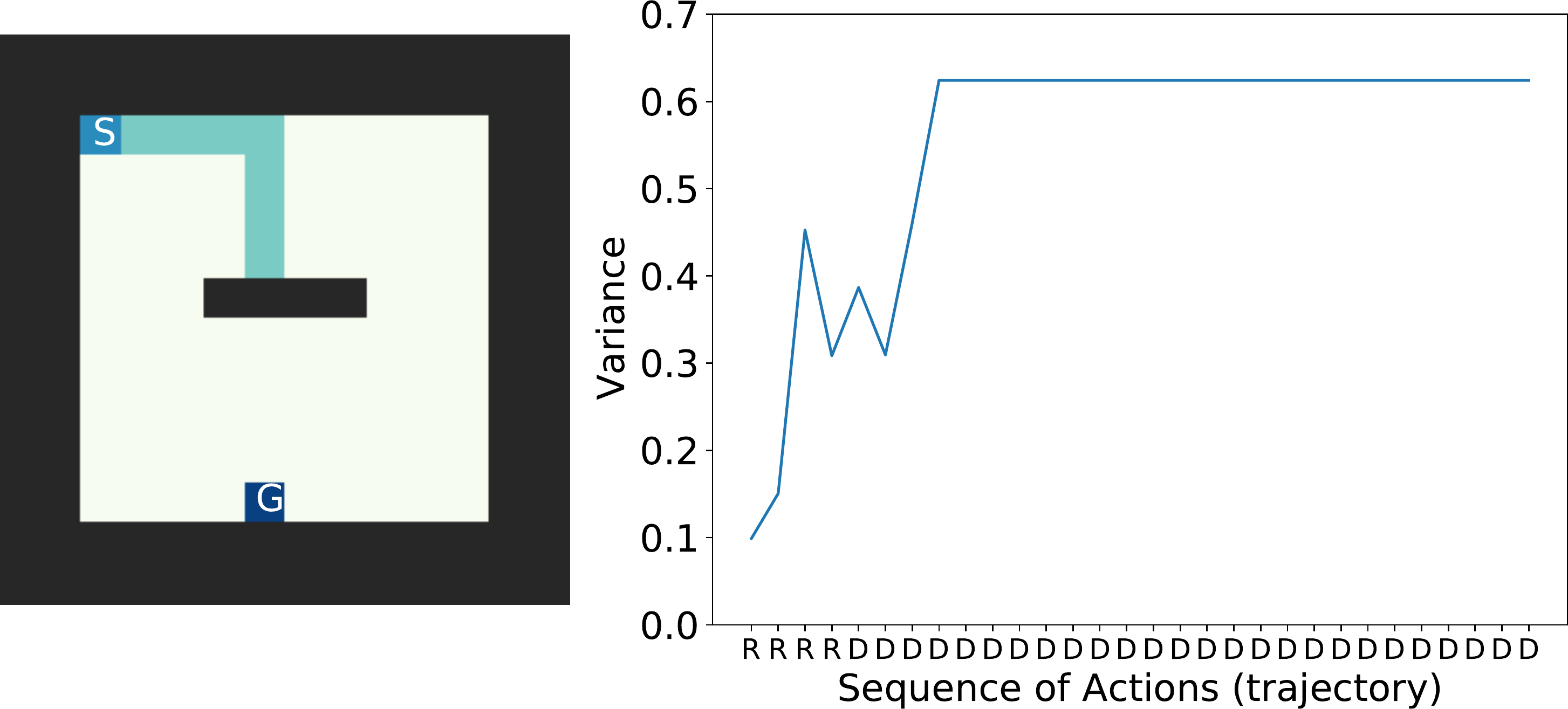}}
  \hspace{0.5cm}
  \subfloat[RE-MOVE with human feedback]{
  \includegraphics[width=7cm]{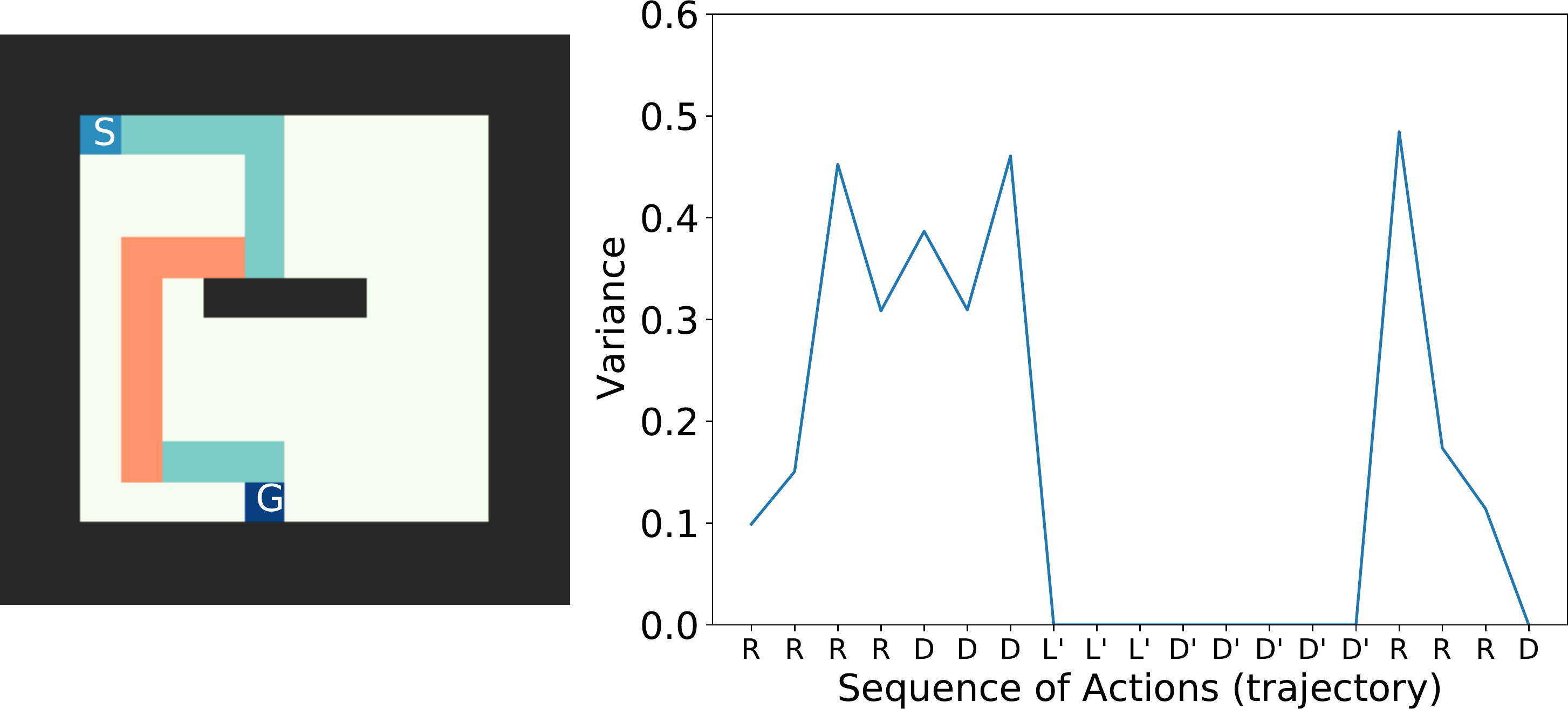}}
    \caption{Trajectories followed by the RE-MOVE agent (trained using the Partial Global Visibility Observation space) with and without human feedback in environments with unseen obstacles. It can be seen that there are intermittent spikes in the prediction uncertainties even when the obstacle is not within the agent's observable range. The increase in the uncertainty when the agent is close to the obstacle is still greater than these spikes. This makes this observation space usable, although with some misclassification where the agent asks for human feedback when it is not necessary.}\label{fig:partial_global_1}
\end{figure}

\begin{figure}[H]
    \centering  
  \subfloat[RE-MOVE without human feedback]{ 
  \includegraphics[width=7cm]{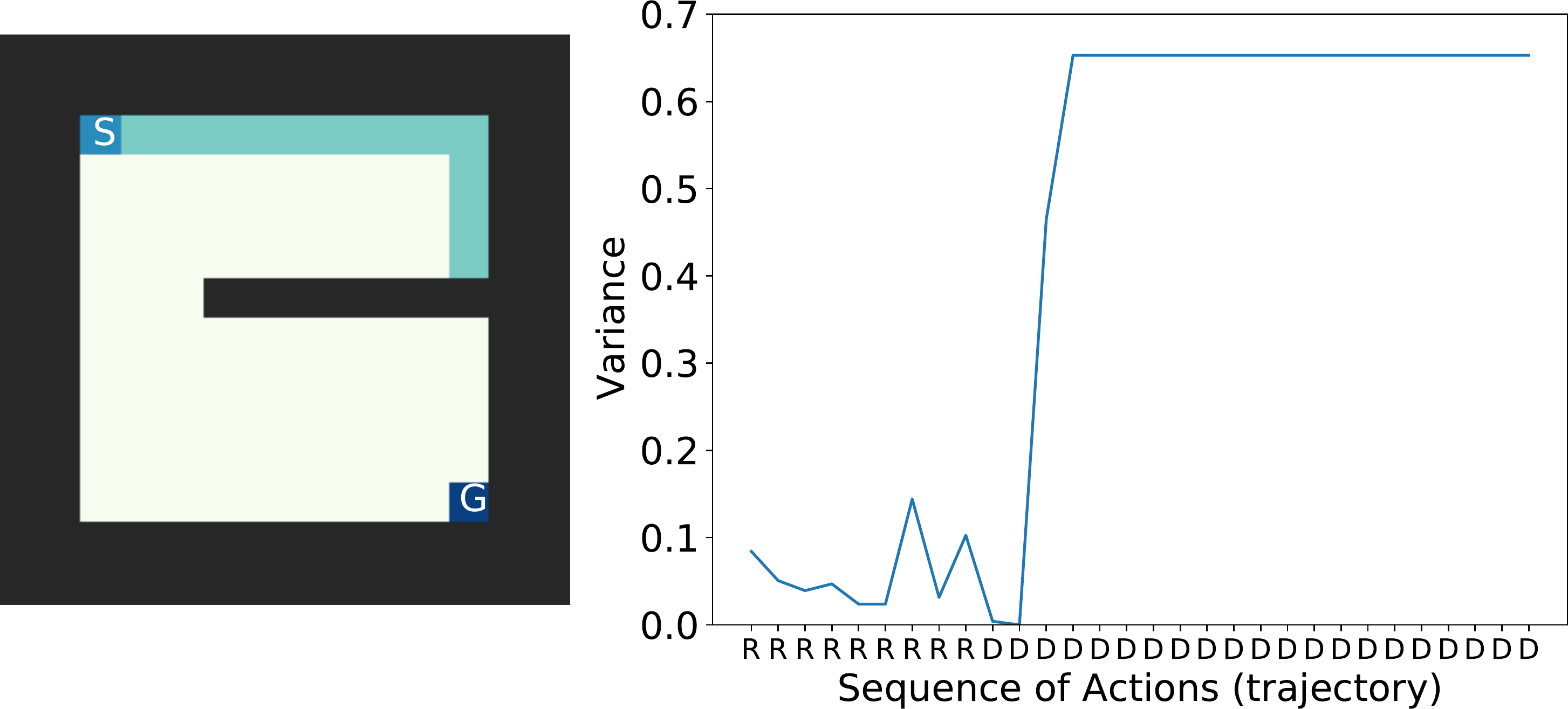}}
  \hspace{0.2cm}
  \subfloat[RE-MOVE with human feedback]{
  \includegraphics[width=7cm]{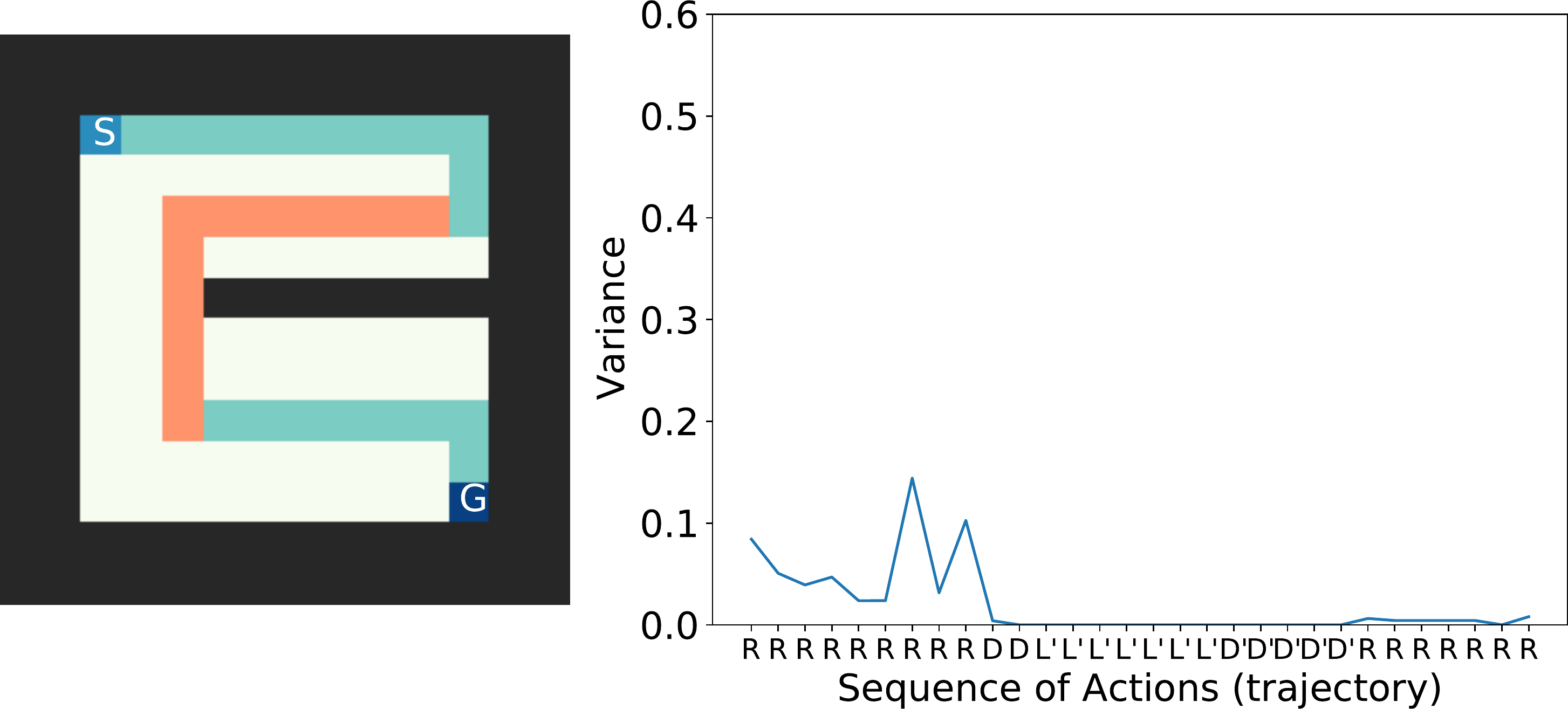}}
  \\
  \subfloat[RE-MOVE without human feedback]{ 
  \includegraphics[width=7cm]{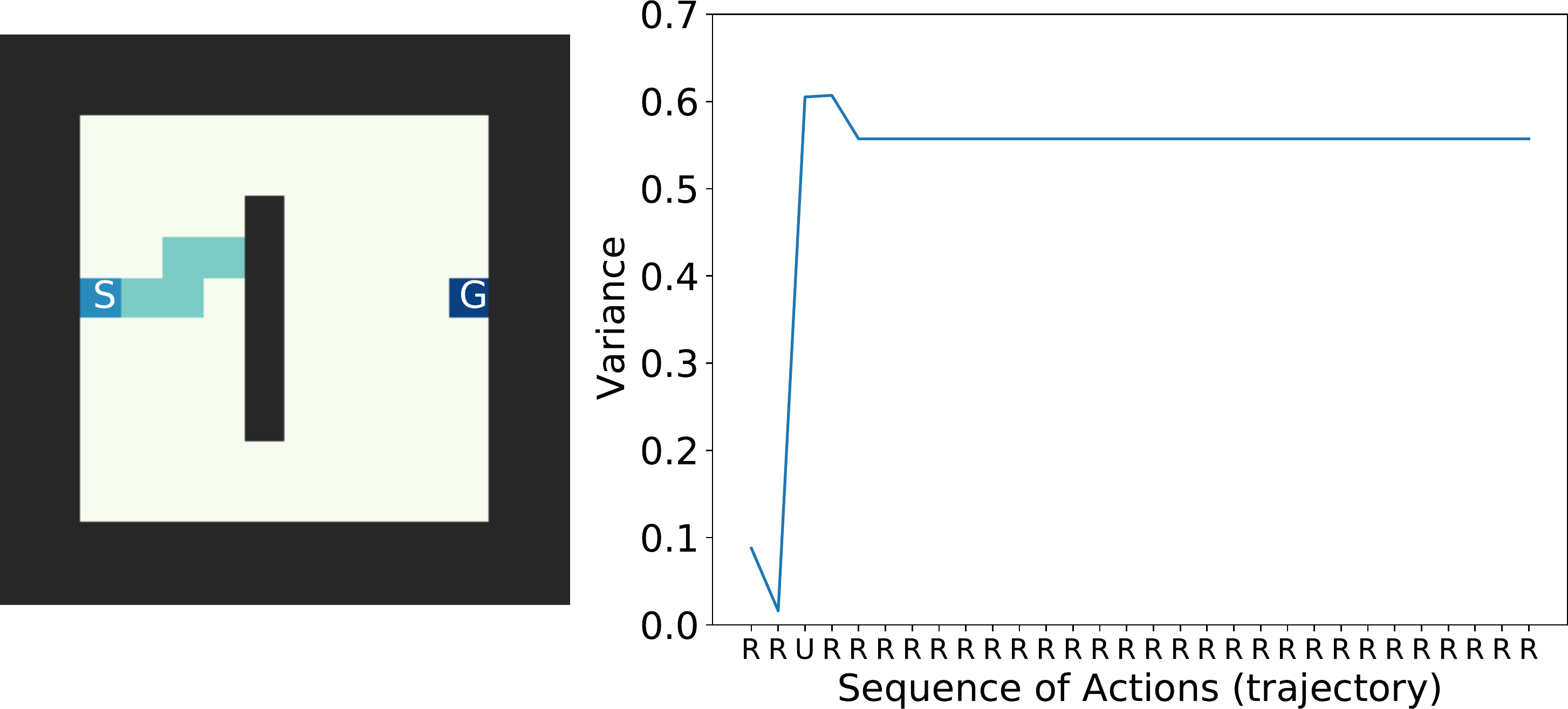}}
  \hspace{0.2cm}
  \subfloat[RE-MOVE with human feedback]{
  \includegraphics[width=7cm]{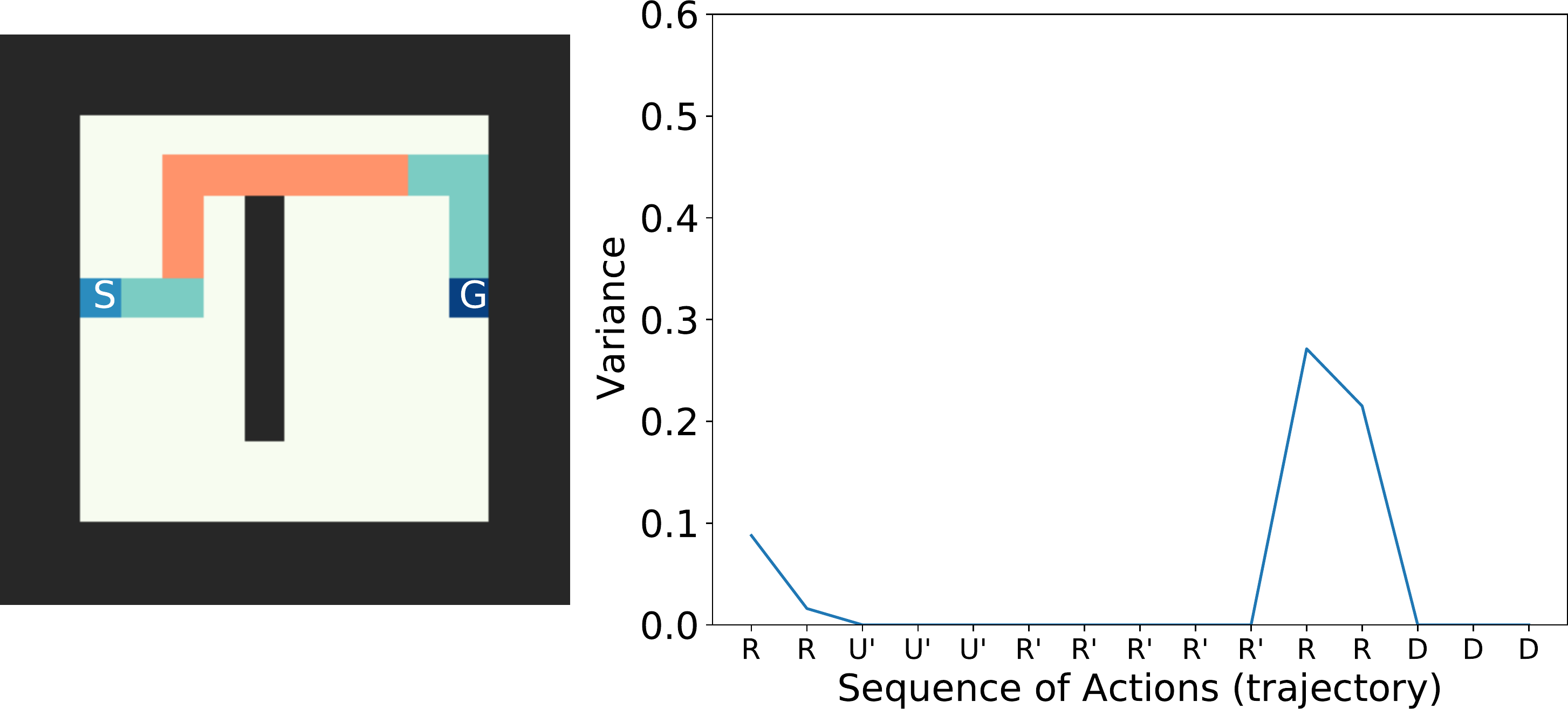}}
    \caption{Testing RE-MOVE with different obstacle configurations using the Partial Global Observation space}\label{fig:partial_global_20}
\end{figure}

\begin{figure}[H]
    \centering  
  \subfloat[RE-MOVE without human feedback]{ 
  \includegraphics[width=7cm]{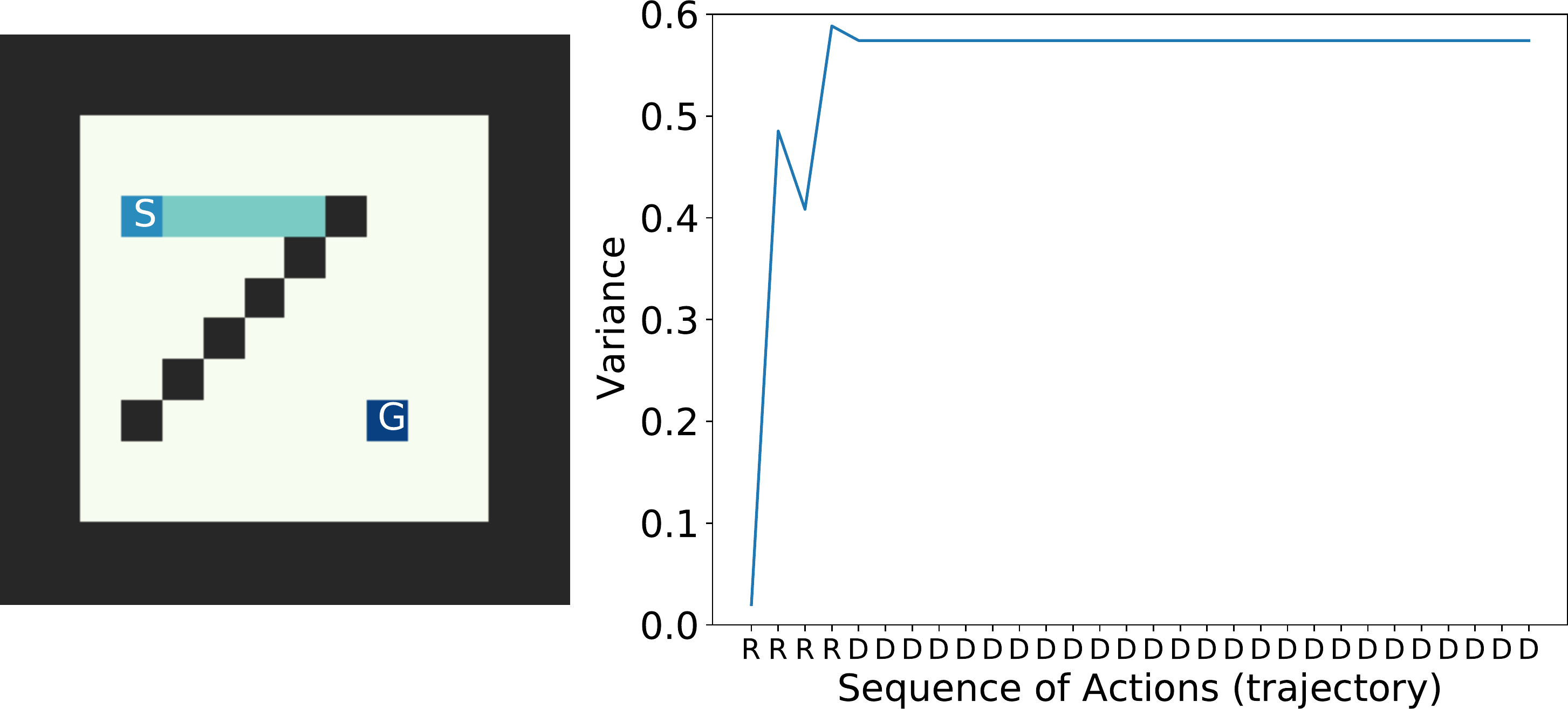}}
  \hspace{0.2cm}
  \subfloat[RE-MOVE with human feedback]{
  \includegraphics[width=7cm]{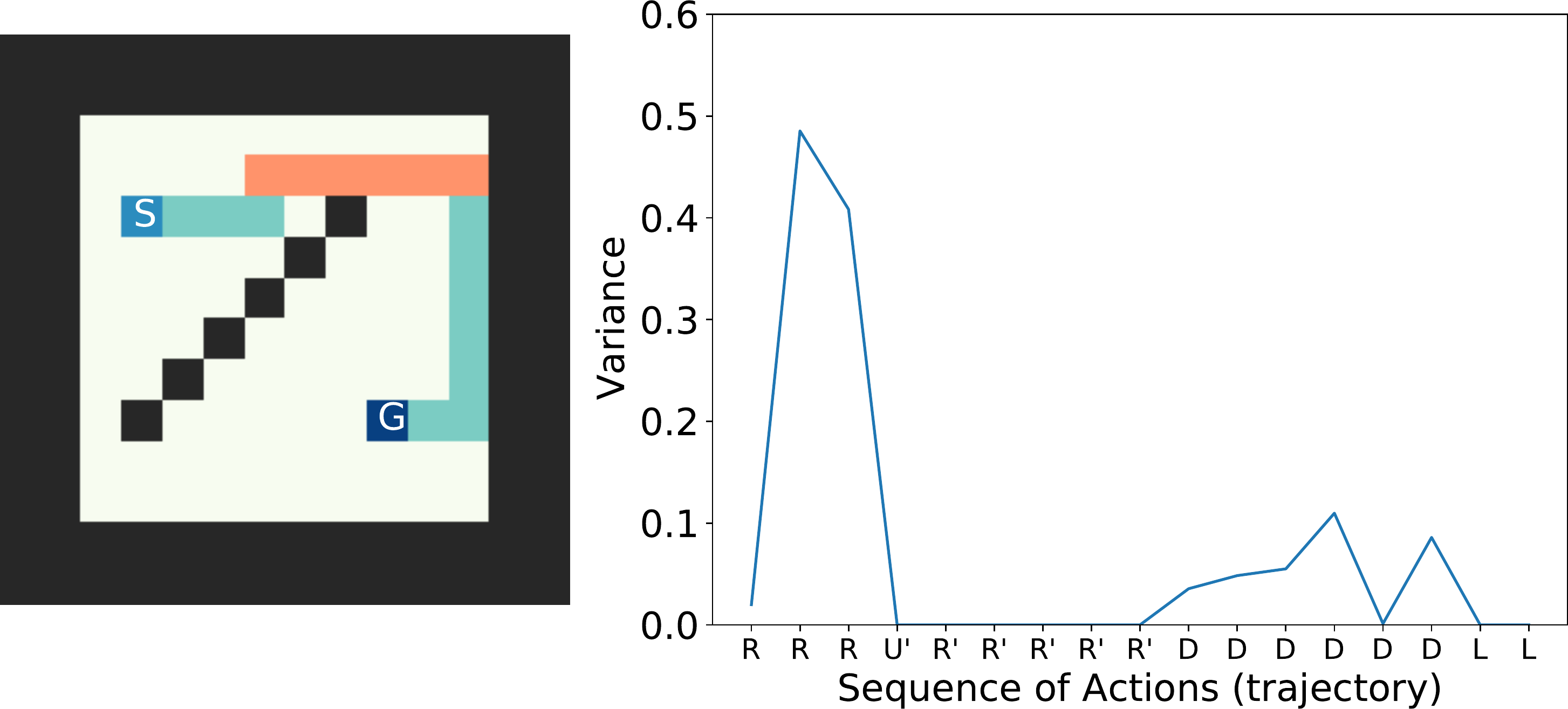}}
  \\
  \subfloat[RE-MOVE without human feedback]{\label{fig:sample2} 
  \includegraphics[width=7cm]{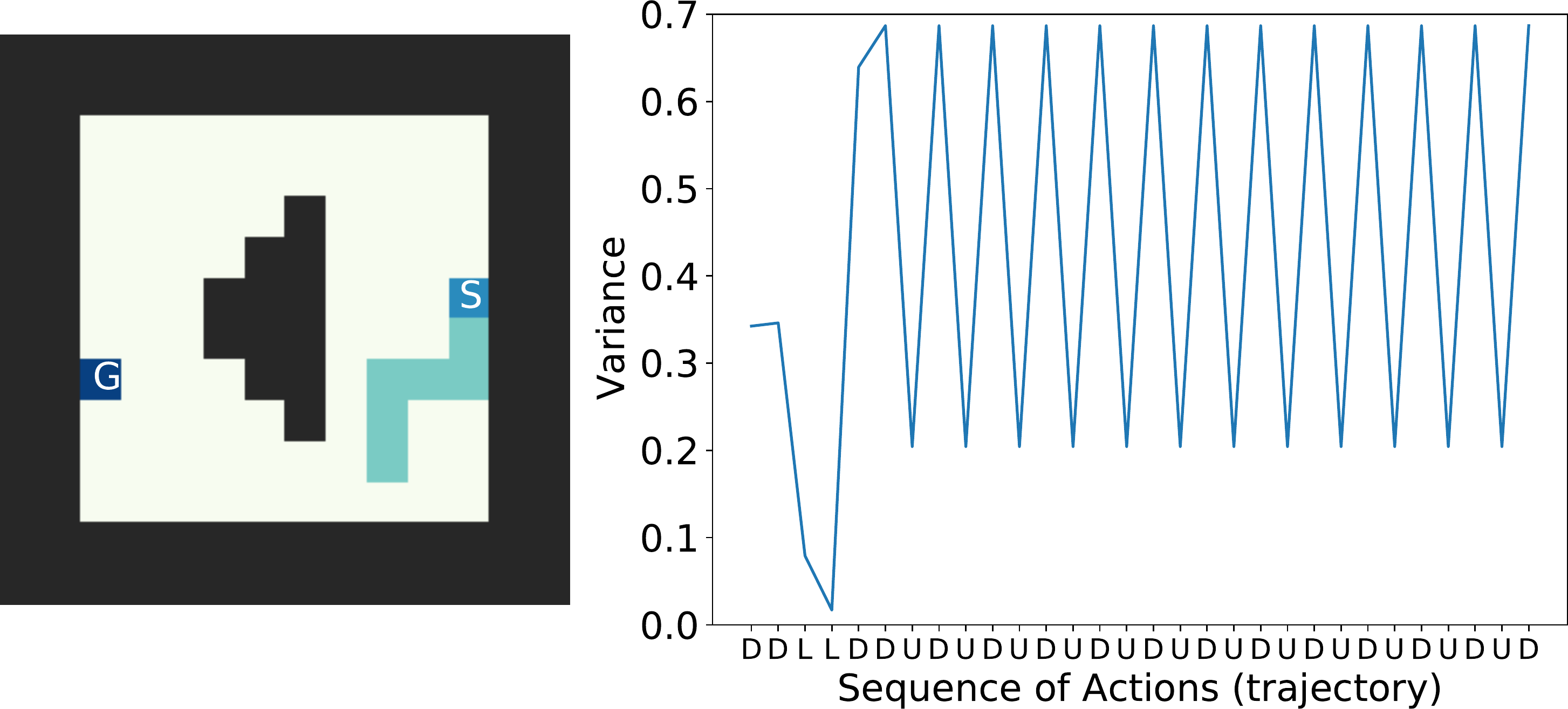}}
  \hspace{0.2cm}
  \subfloat[RE-MOVE with human feedback]{
  \includegraphics[width=7cm]{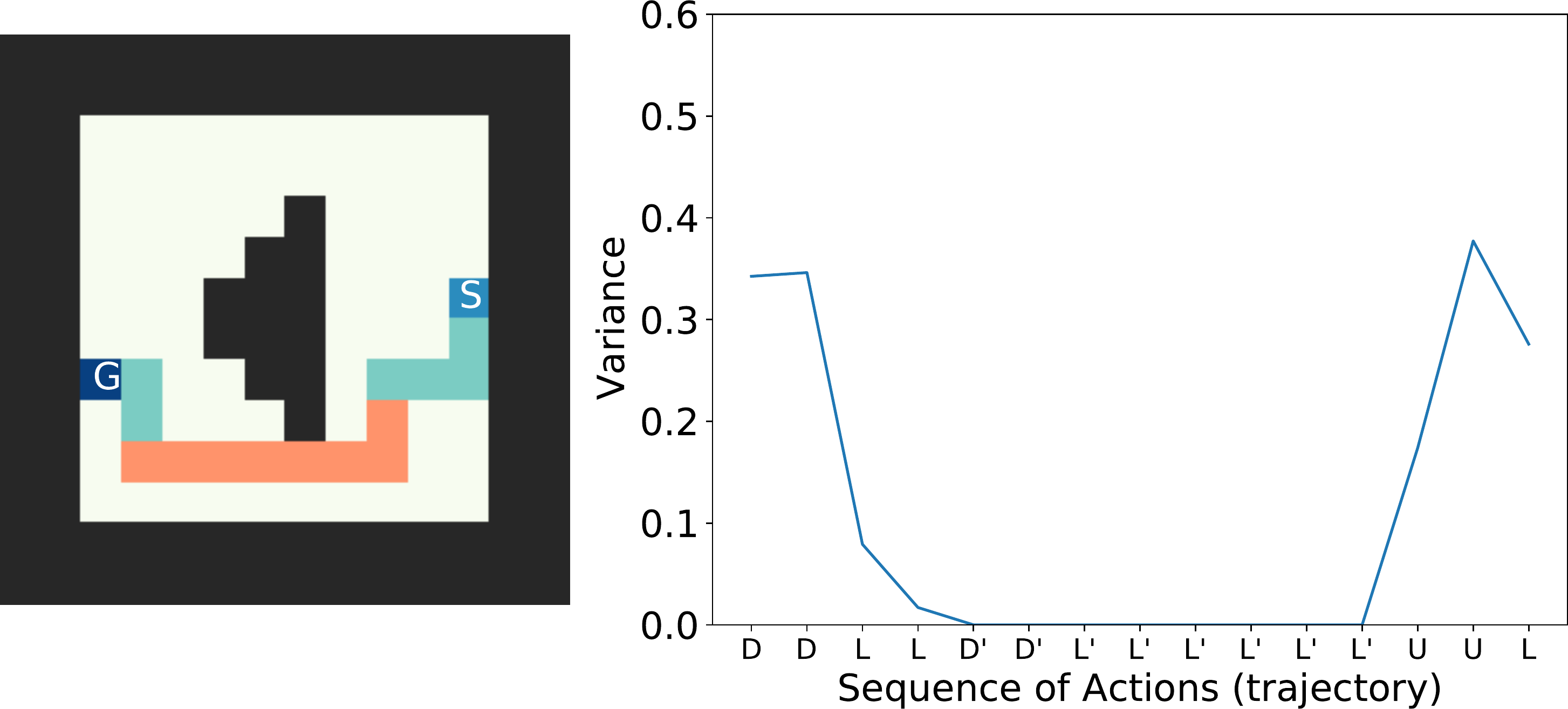}}
  \\
  \subfloat[RE-MOVE without human feedback]{ 
  \includegraphics[width=7cm]{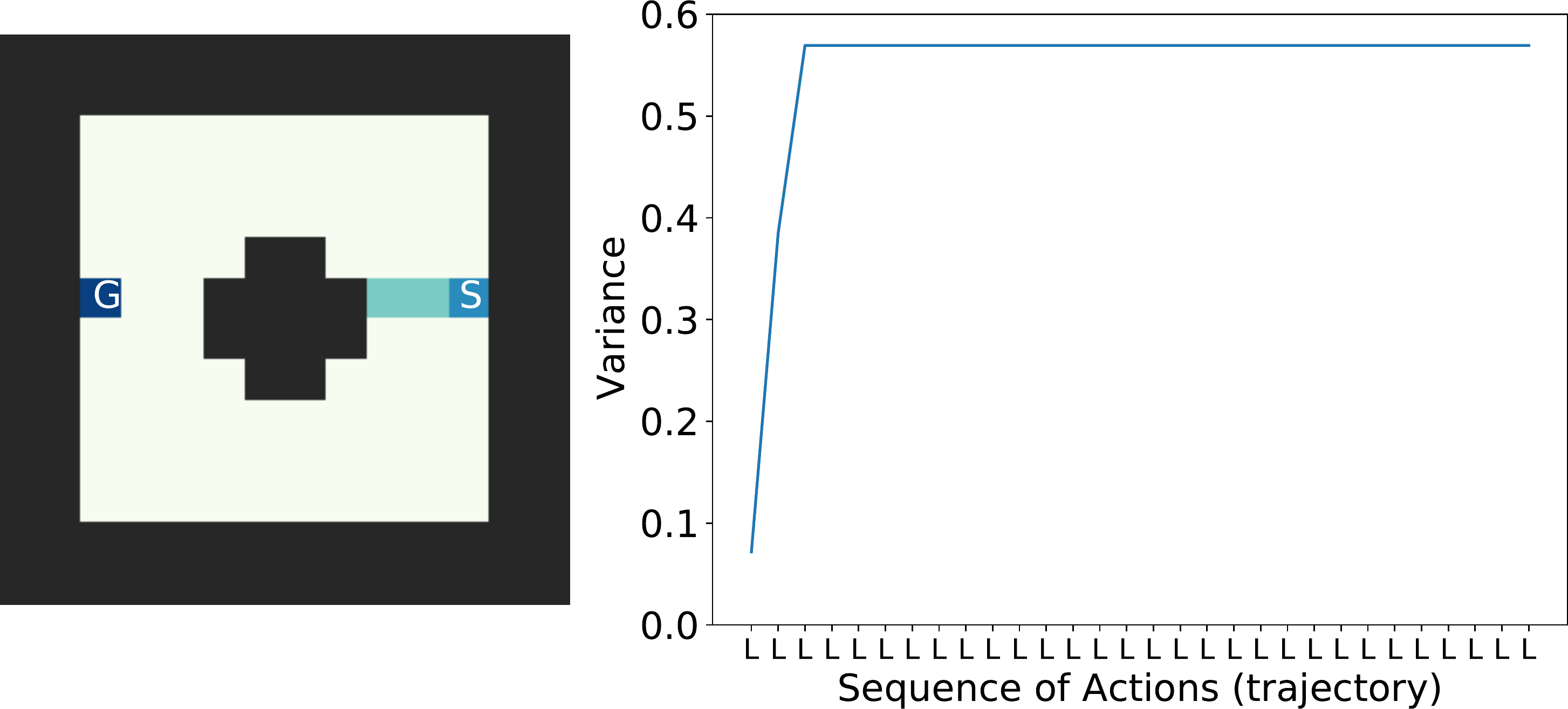}}
  \hspace{0.2cm}
  \subfloat[RE-MOVE with human feedback]{
  \includegraphics[width=7cm]{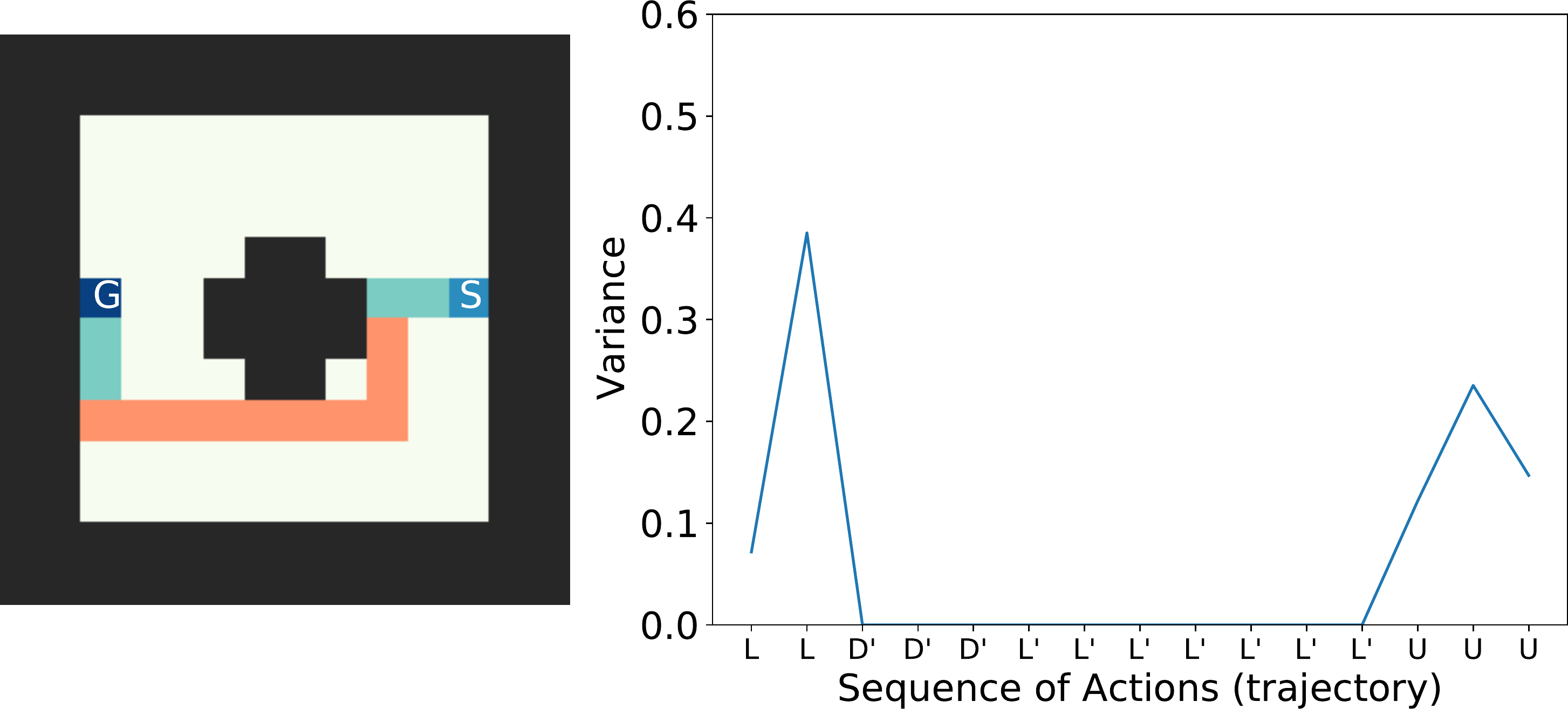}}
  
    \caption{Testing RE-MOVE with different obstacle configurations using the Partial Global Observation space}\label{fig:partial_global_2}
\end{figure}

As described in Section \ref{sec:observation_space} we test RE-MOVE with agents trained on the Partial Global Visibility Observation Space. Figure \ref{fig:partial_global_1}- \ref{fig:partial_global_2} shows the trajectories followed by the agent and the corresponding uncertainties in the action predictions. It can be noticed in Figure \ref{fig:partial_global_1} that there are intermittent spikes in the uncertainty even when the obstacle in not close to the agent. Although the increase in the uncertainty is much higher when the agent is close to the obstacle, these intermittent spikes make it tough to identify the correct time for asking for human feedback. This leads to RE-MOVE asking for human feedback even when it is not needed. A similar trend with the uncertainty predictions can be seen in Figure \ref{fig:sample2}. This makes the Partial Global Visibility Observation Space unsuitable for the RE-MOVE algorithm.
\section{Experiments with Partial Obstacles}
\begin{figure}[H]
    \centering  
  \subfloat[]{ 
  \includegraphics[width=0.45\columnwidth]{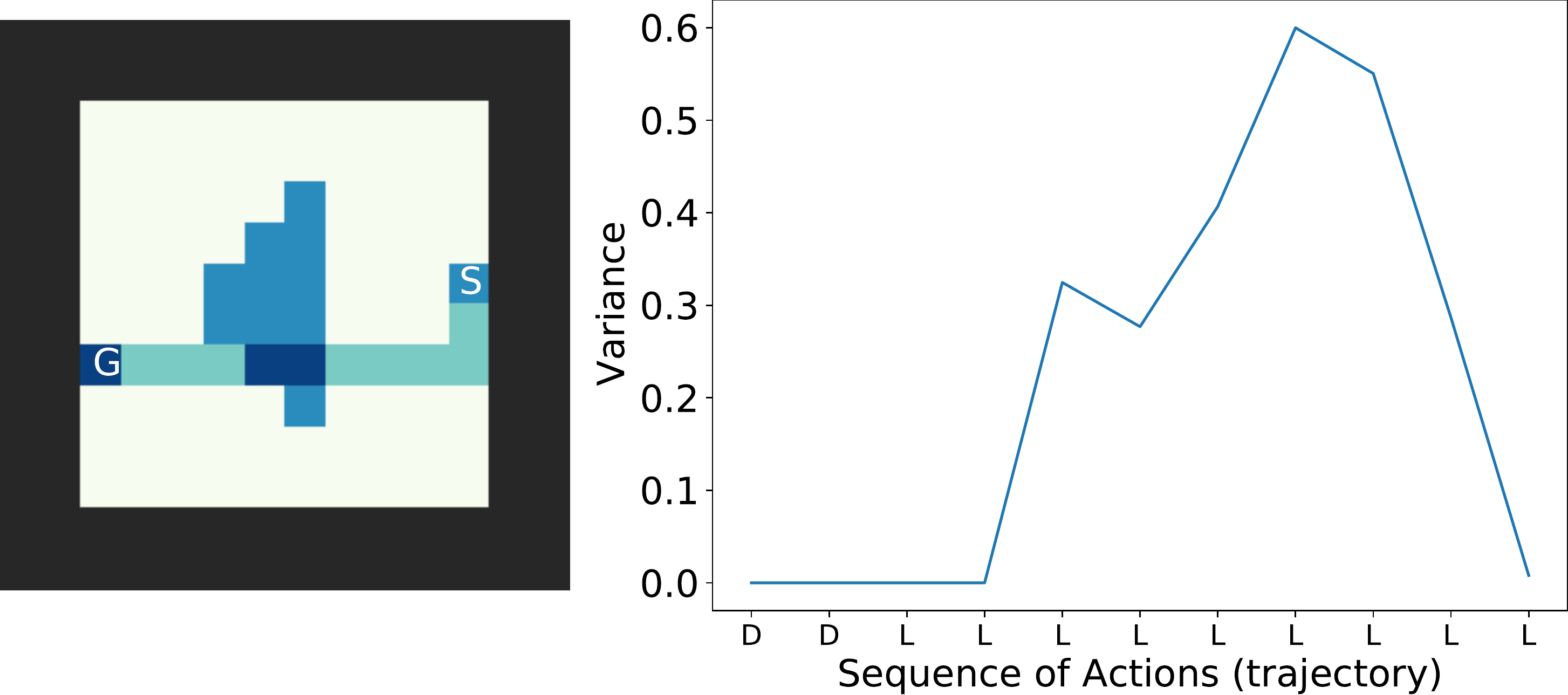}}
  \hfill
  \subfloat[]{ 
  \includegraphics[width=0.45\columnwidth]{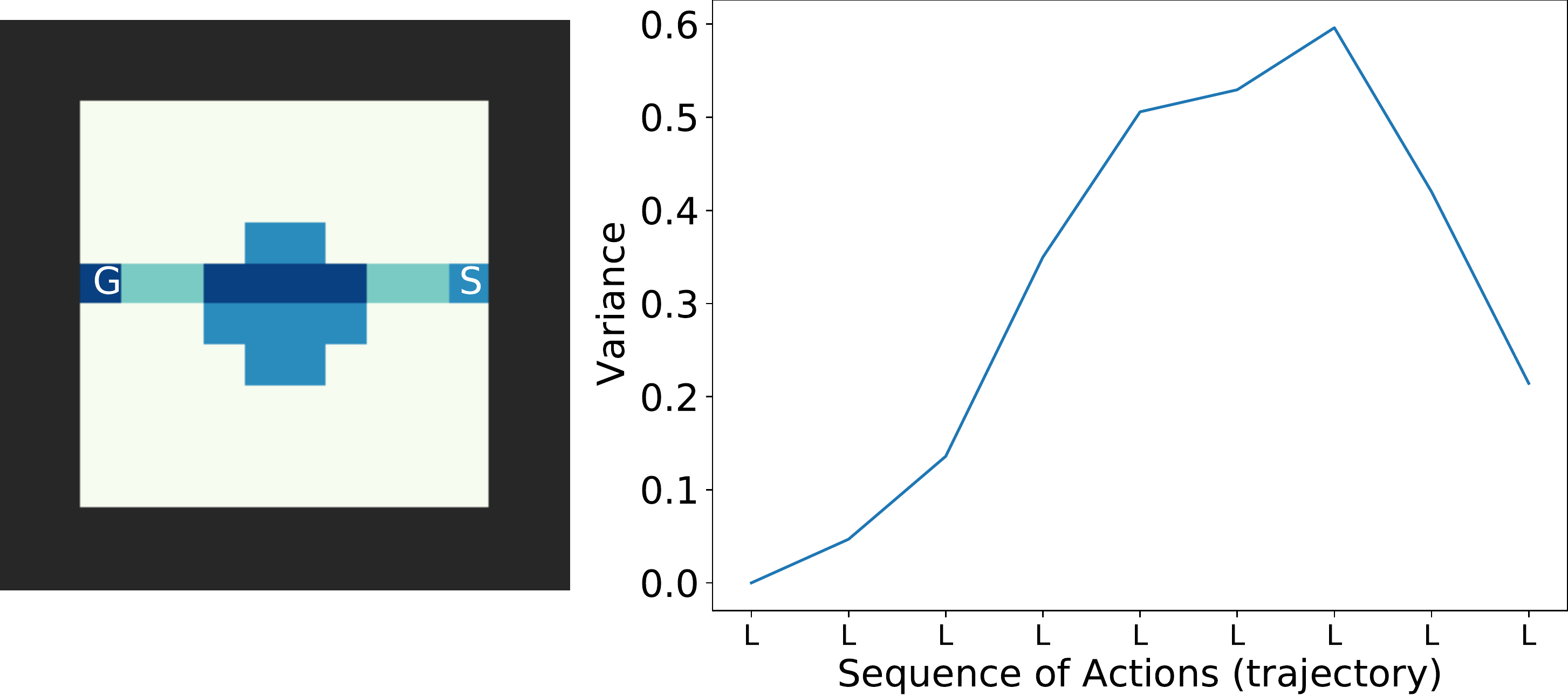}}
  
    \caption{Testing with \textit{partial obstacles} or obstacles through which the agent can pass. It can be noticed that the uncertainty in the action predictions increases when the agent passes through the obstacle, indicating that the uncertainty estimates work correctly.} \label{fig:partial_obs}
\end{figure}

In this section, we conduct experiments with obstacles that the agent can pass through, although it is not desirable to do so. For example, these obstacles may represent a slippery section of a road in an autonomous driving problem where the car can drive through the slippery patch but with a higher chance of losing control. Hence it is necessary that RE-MOVE identifies such scenarios and asks for human feedback. Figure \ref{fig:partial_obs} shows the uncertainty in the action predictions as the agent passes through the obstacle. It can be seen that the uncertainty increases as the agent reaches close to the obstacle and then passes through it. This shows that RE-MOVE can identify such obstacles based on increased uncertainty.
\clearpage
\section{ChatGPT Prompts}
\begin{figure}[H]
    \centering  
  \includegraphics[scale=0.3]{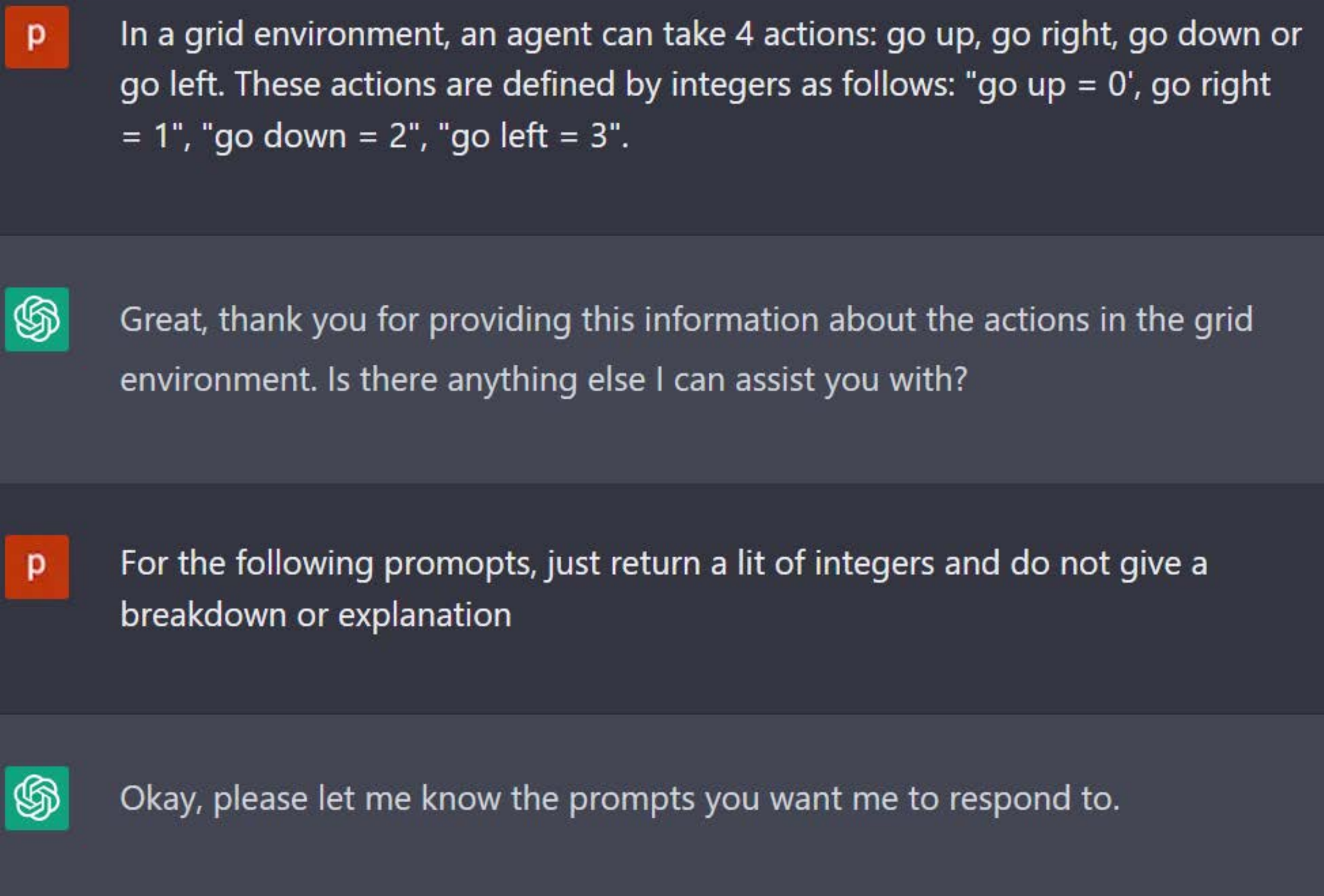}
    \caption{Initial instructions given to ChatGPT, indicating the rules of converting actions into corresponding integer values.}
\end{figure}
\begin{figure}[H]
    \centering  
    \subfloat[]{
  \includegraphics[width=0.4\columnwidth]{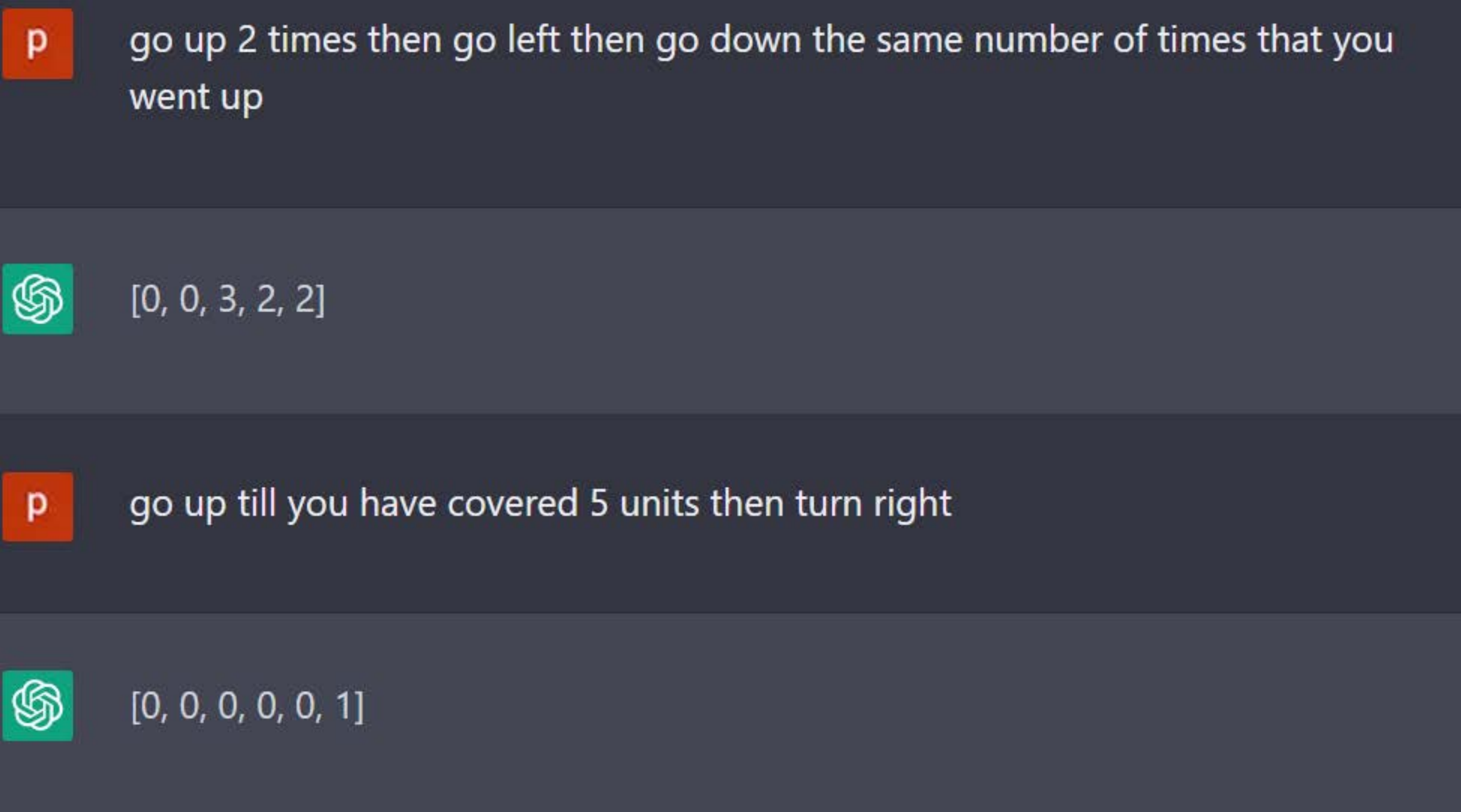}}
  \subfloat[]{
  \includegraphics[width=0.4\columnwidth]{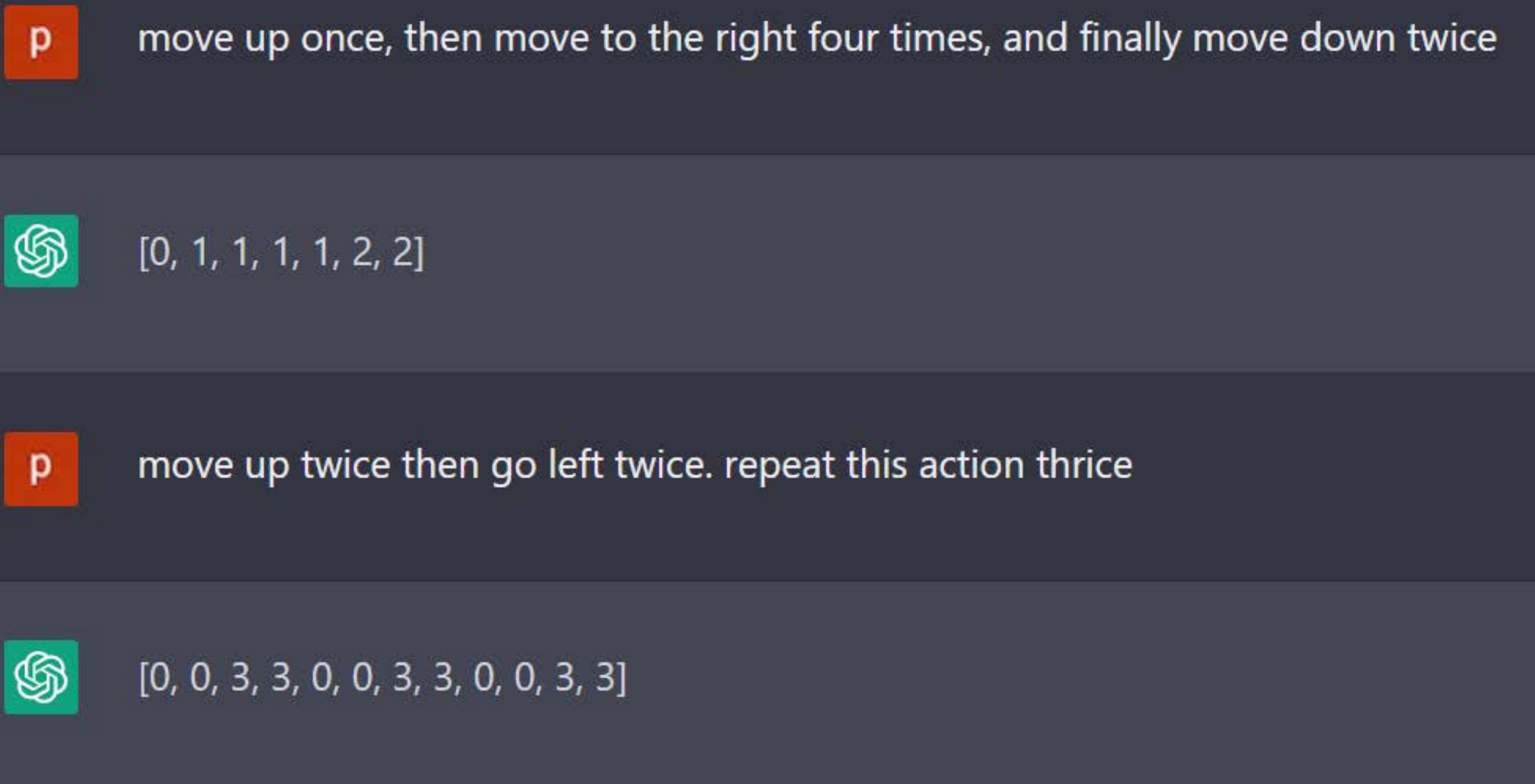}}
    \caption{Testing the first set of question prompts with ChatGPT. These prompts are straightforward in defining the sequence of actions to be taken by the agent.}
\end{figure}
\begin{figure}[H]
    \centering  
    \subfloat[]{
  \includegraphics[width=8cm]{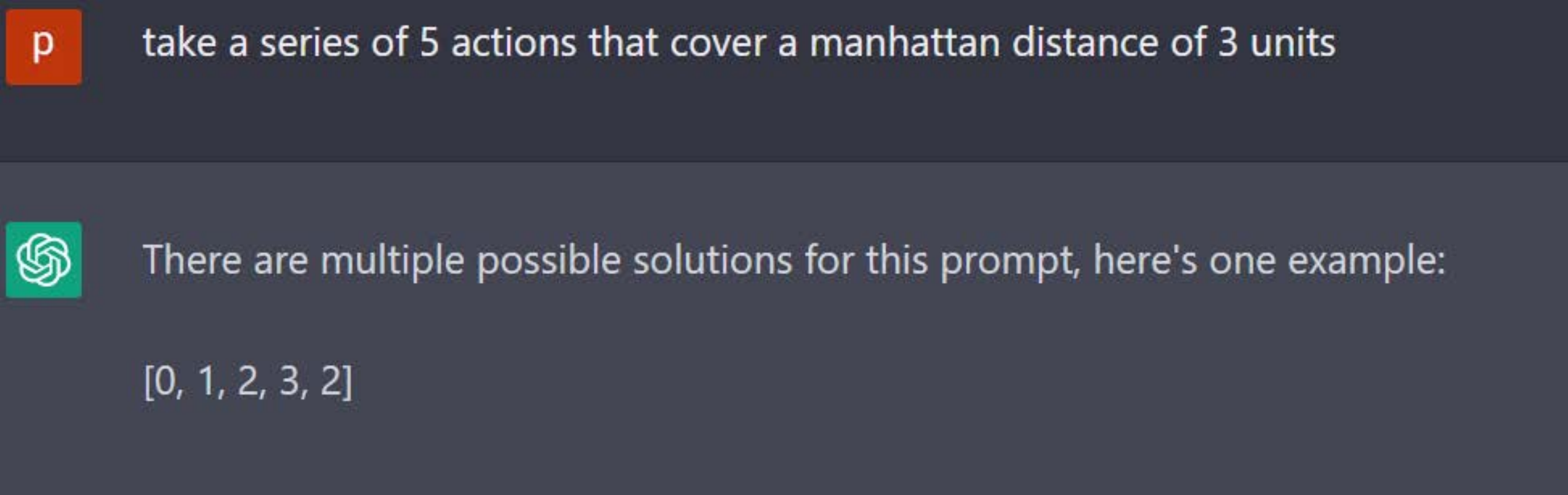}}
  \\
  \subfloat[]{
  \includegraphics[width=8cm]{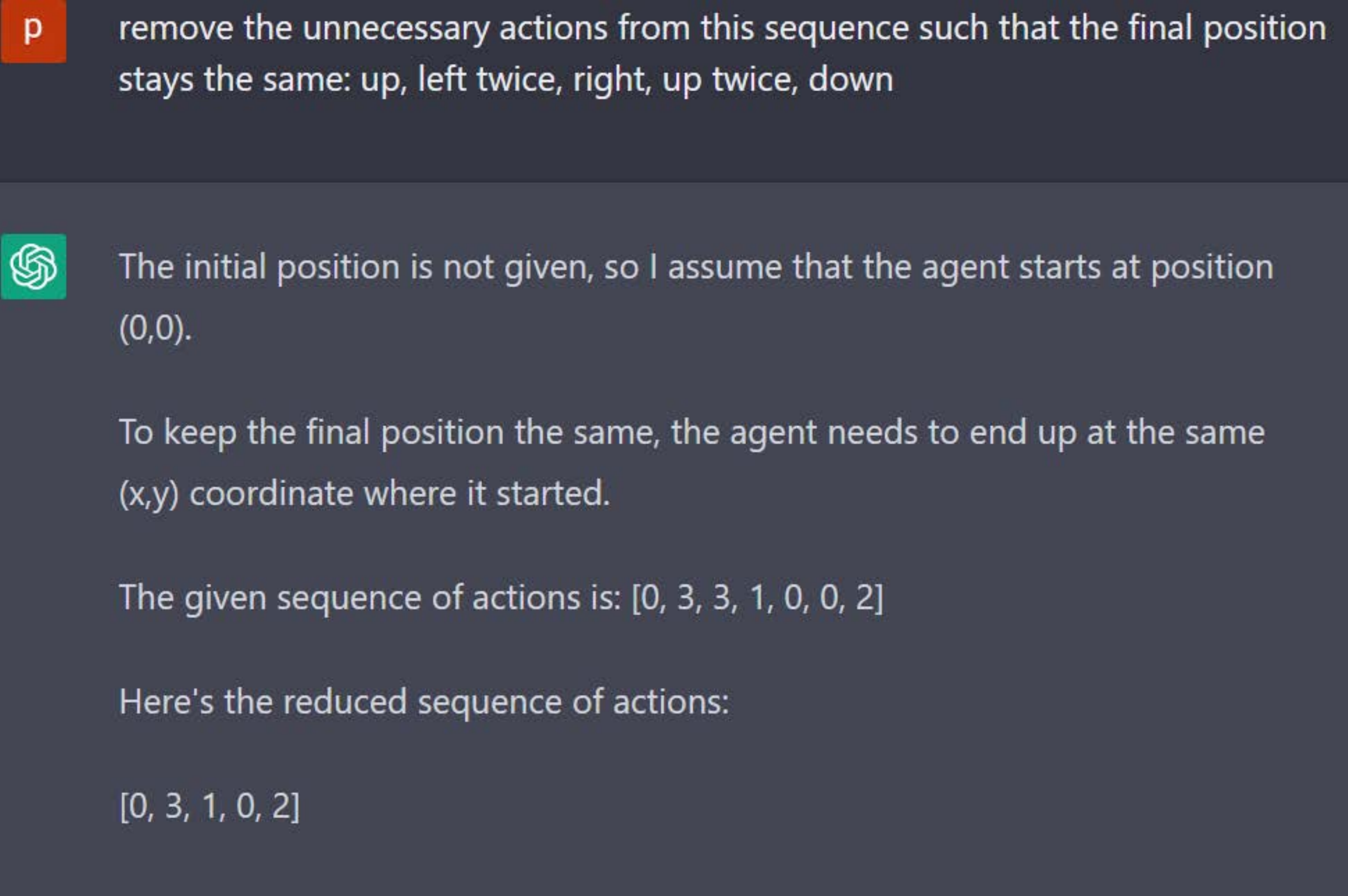}}
    \caption{Testing the second set of question prompts where the instructions are more abstract and it can be seen that ChatGPT can successfully convert these prompts in action sequences.}
\end{figure}
\begin{table}
\begin{center}
\begin{tabular}{l} 
 \textbf{Instruction:} In a grid environment, an agent can take 4 actions: go up, go right, go down or go left. \\ 
 These actions are defined by integers as follows: "go up = 0', go right = 1", "go down = 2", "go left = 3". \\ 
 \hline\hline
 \textbf{Question Prompt:} go up 2 times then go left then go down the same number of times that you went up \\
 \textbf{Output:} [0, 0, 3, 2, 2]\\ 
 \hline
 \textbf{Question Prompt:} go right three times, then step down once and then go left twice\\
 \textbf{Output:} [1, 1, 1, 2, 3, 3] \\
 \hline
 \textbf{Question Prompt:} step up once, then move left and right alternatively four times each\\
 \textbf{Output:} [0, 3, 1, 3, 1, 3, 1, 3, 1]\\
 \hline
 \textbf{Question Prompt:} go down once, move right four times, and then move up twice.\\
 \textbf{Output:} [2, 1, 1, 1, 1, 0, 0]\\
 \hline
 \textbf{Question Prompt:} move left once, then go up two steps, and finally move to the right three times\\
 \textbf{Output:} [3, 0, 0, 1, 1, 1]\\
 \hline
 \textbf{Question Prompt:} go right thrice, move down once, and then move to the left four times\\
 \textbf{Output:} [1, 1, 1, 2, 3, 3, 3, 3]\\
 \hline
 \textbf{Question Prompt:} move to the left twice, go up three steps, and then move to the right twice.\\
 \textbf{Output:} [3, 3, 0, 0, 0, 1, 1]\\
 \hline
 \textbf{Question Prompt:} go down twice, then move to the right twice, and finally go up thrice\\
 \textbf{Output:} [2, 2, 1, 1, 0, 0, 0]
 
\end{tabular}
\caption{Additional question prompts that were given to ChatGPT and their corresponding outputs.}
\end{center}
\end{table}

\section{Additional Real-world Experiments}\label{additional_real_world}
\begin{figure}[H]
    \centering
\includegraphics[width=\textwidth]{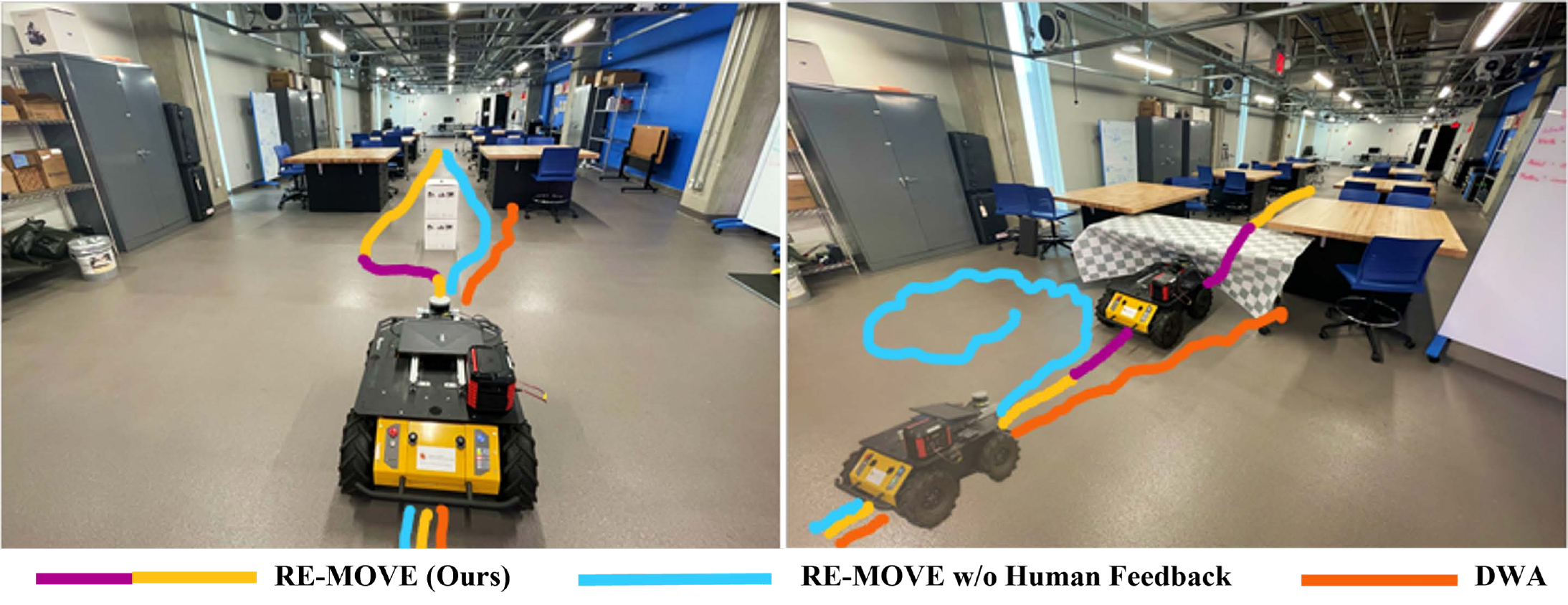}
    \caption{Navigation trajectories generated by RE-MOVE(Ours), RE-MOVE w/o human feedback and DWA\cite{DWA} in real-world environments. \textbf{[Left]:} Scenario 1;\textbf{[Right]:} Scenario 3. We observe that both RE-MOVE and DWA are able to reach the goal by avoiding obstacles and through the free spaces in scenario 1. However, DWA or similar LiDAR-based navigation methods fail in scenario 3 since there are no free spaces observable from the LiDAR scan (i.e., leads to freezing). In contrast, our RE-MOVE formulation seeks human assistance to navigate through the pliable region (i.e., through the tablecloth) instead of completely depending on the robot's LiDAR sensor-based observations. Hence, our methods can be generalized to address the perception challenges that occur in real-world environments where sensory observations are ambiguous due to perceptually deceptive objects or obstacles.}
    \label{fig:real_trajs_appendix}
    \vspace{-5pt}
 \end{figure}

\end{document}